\documentclass[10pt,twocolumn,letterpaper]{article}

\usepackage[pagenumbers]{cvpr} 

\usepackage{float}
\usepackage[dvipsnames]{xcolor}
\usepackage{array}
\usepackage{colortbl}
\usepackage{graphicx}
\usepackage{epigraph}
\usepackage{subcaption}
\usepackage[autolanguage]{numprint}
\usepackage[utf8]{inputenc}
\usepackage{csquotes}
\usepackage{soul}
\usepackage{amssymb}
\usepackage{booktabs}
\usepackage{amsmath}
\usepackage{multirow}
\usepackage{siunitx}
\usepackage{xcolor}
\usepackage{graphicx} 
\usepackage{algorithm}
\usepackage{algpseudocode}
\usepackage{amsmath}
\usepackage{listings}
\usepackage{overpic}
\usepackage{dashrule}
\usepackage{enumitem} 
\newcommand{\cmark}{\checkmark} 
\newcommand{\xmark}{\times} 

\usepackage{tikz}
\definecolor{cvprblue}{rgb}{0.21,0.49,0.74}
\definecolor{MyCustomBlue}{HTML}{150CA6}
\usepackage[
    pagebackref,
    breaklinks,
    colorlinks=true, 
    citecolor=MyCustomBlue, 
    pdftitle={Cupid},
]{hyperref}
\usepackage{xspace}
\newcommand{\Cupid}{\textsc{Cupid}\xspace}

\newcommand{\pose}{\boldsymbol{\theta}}
\newcommand{\object}{\mathcal{O}}
\newcommand{\Image}{\mathbf{I}}
\newcommand{\image}{\mathbf{I}^{\textrm{cond}}}
\newcommand{\mask}{\mathbf{M}^{\textrm{occ}}}
\newcommand{\render}{\mathbf{\mathcal{P}}}
\newcommand{\latent}{\mathbf{z}}
\newcommand{\timestep}{t}
\newcommand{\latentt}{\mathbf{z}_{\timestep}}
\newcommand{\latentzero}{\mathbf{z}_{0}}
\newcommand{\noise}{\boldsymbol{\epsilon}}
\newcommand{\velocity}{\mathbf{v}}
\newcommand{\parameter}{\phi}
\newcommand{\encoder}{\varphi}
\newcommand{\point}{\mathbf{x}}
\newcommand{\pixcoord}{\mathbf{u}}
\newcommand{\feat}{\mathbf{f}}
\newcommand{\sflow}{\mathbf{\mathcal{G}_{S}}}
\newcommand{\lflow}{\mathbf{\mathcal{G}_{L}}}

\newcommand{\numpts}{L}

\newcommand{\trellis}{{\textsc{TRELLIS}}\xspace}
\newcommand{\dino}{\textsc{DINO}\xspace}
\newcommand{\interp}{\operatorname{BilinearInterp}}
\newcommand{\slatenc}{\operatorname{SlatEncoder}}

\newcommand{\low}{\mathrm{l}}
\newcommand{\high}{\mathrm{h}}

\usepackage{amsmath,amsfonts,bm}

\def\Tabref#1{Tab.~\ref{#1}}
\def\Secref#1{Sec.~\ref{#1}}
\def\Appref#1{App.~\ref{#1}}
\def\Figref#1{Fig.~\ref{#1}}
\def\Eqref#1{Eq.~\ref{#1}}

\def\eqref#1{equation~\ref{#1}}

\def\1{\bm{1}}

\DeclareMathAlphabet{\mathsfit}{\encodingdefault}{\sfdefault}{m}{sl}
\SetMathAlphabet{\mathsfit}{bold}{\encodingdefault}{\sfdefault}{bx}{n}

\DeclareMathOperator*{\argmin}{arg\,min}

\def\paperID{12412} 
\def\confName{CVPR}
\def\confYear{2026}

\title{\Cupid: Generative 3D Reconstruction via Joint Object and Pose Modeling}

\author{
Binbin Huang$^{\dag,1,2}$\quad Haobin Duan$^{\dag,1,2}$ \quad Yiqun Zhao$^{1,2}$\quad Zibo Zhao$^{3}$\quad Yi Ma$^{1,2}$\quad Shenghua Gao$^{\ddag, 1,2}$
\vspace{8pt}\\
	$^{1}$The University of Hong Kong\qquad $^{2}$Transcengram\qquad $^{3}$Tencent\vspace{8pt}\\
	{\url{https://cupid3d.github.io}}
}

\begin{document}
\maketitle

\renewcommand{\thefootnote}{\fnsymbol{footnote}}
\footnotetext{$\dag$ Equal contribution.  $\ddag$ Corresponding author}

\begin{abstract}
    We introduce \Cupid, a generative 3D reconstruction framework that jointly models the full distribution over both canonical objects and camera poses. Our two-stage flow-based model first generates a coarse 3D structure and 2D-3D correspondences to estimate the camera pose robustly. Conditioned on this pose, a refinement stage injects pixel-aligned image features directly into the generative process, marrying the rich prior of a generative model with the geometric fidelity of reconstruction. This strategy achieves exceptional faithfulness, outperforming state-of-the-art reconstruction methods by over 3 dB PSNR and 10\% in Chamfer Distance. As a unified generative model that decouples the object and camera pose, \Cupid naturally extends to multi-view and scene-level reconstruction tasks without requiring post-hoc optimization or fine-tuning.
\end{abstract}
\vspace{-2mm}
\section{Introduction}
\label{sec:intro}

Inferring an object's 3D structure from a single 2D image is a fundamental challenge in computer vision. Consider the physics of image formation: $\Image = \render(\object, \pose)$, where $\Image$ is the rendered image, $\object$ is the 3D object in an \textit{object-centric}, \textit{view-agnostic} canonical frame, $\pose$ is the camera pose in that frame, and $\render(\cdot, \cdot)$ projects the 3D object onto the image. Since object and pose jointly determine $\Image$, given a single image, it is extremely challenging to disentangle the intrinsic properties of object $\object$ (\emph{e.g.}, shape and texture) from 
the extrinsic property of the partial observation, namely the camera viewpoint $\pose$. 

Current approaches often tackle this problem from only one aspect.
On one hand, \textbf{3D generative models}~\citep{xiang2025structured,zhang2024clay,zhao2025hunyuan3d,liu2024cage,lu2025orientation} excel at generating a high-quality canonical 3D object $\object$, while completely ignoring the camera pose $\pose$, a crucial component of image formation. As a result, contemporary image-conditioned 3D generators typically produce plausible \textit{variations}, but struggle to achieve a \textit{faithful reconstruction} consistent with the input view, leading to shape inconsistencies and color drift, especially when real-world inputs deviate from the training distribution. Furthermore, integrating generated 3D assets into complex scene reconstruction is highly challenging, requiring costly and fragile post-hoc optimization for spatial alignment~\citep{yao2025castcomponentaligned3dscene,dong2025hiscene,geng2025one}. The difficulty arises primarily from the missing camera pose prior, which is never modeled in these approaches.

On the other hand, common \textbf{3D reconstruction methods}~\citep{schoenberger2016sfm, depth_anything_v2,  wang2025vggt, wang2024dust3r, wang2025moge} produce 3D geometry through 2D-to-3D lifting in a \textit{pixel-aligned}, \textit{view-centric} frame, which essentially fixes the camera pose $\pose$ as an identity constant. As a result, they lack the generative ability to plausibly ``imagine" occluded regions, yielding incomplete models or collapsing the scene into a single, undifferentiated mesh even given sufficient views~\citep{wang2025vggt, wang2024dust3r, wang2025moge}. Although some methods~\citep{hong2023lrm, wang2023pf} attempt to deterministically complete unseen regions, learning such an entangled data distribution limits their ability to truly ``understand" the underlying 3D spatial relationship that is crucial to embodied AI.

\begin{figure}[t]
    \centering
    \includegraphics[width=\linewidth]{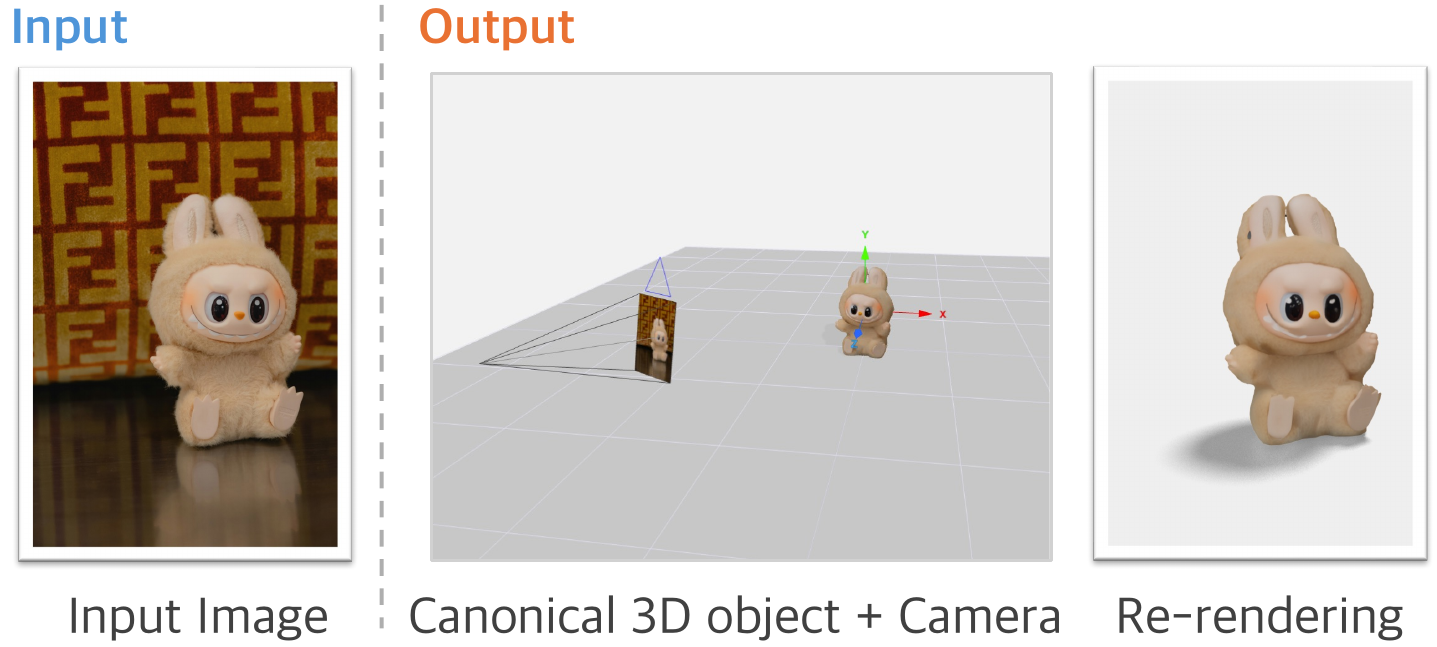}
    \caption{\textbf{Generative 3D reconstruction from a single image.} 
    Left: Input image. 
    Middle: Our method recovers a canonical 3D model (intrinsic properties such as shape and texture) along with the exact camera pose (extrinsic property) that generated the input.
    Right: Re-rendering the 3D model from the estimated pose faithfully reproduces the input.}
    \vspace{-4pt}
    \label{fig:teaser_small}
\end{figure}
\newlength{\teaserwidthb}
\setlength{\teaserwidthb}{0.248\textwidth}

\newcommand{\betweencols}{\hspace{0.15em}}
\begin{figure*}[!ht]
\captionsetup[subfigure]{labelformat=empty}
\centering
\makebox[\textwidth][c]{
\renewcommand{\arraystretch}{0.4}
\begin{tabular}{@{}c@{\betweencols}c@{\betweencols}c@{\betweencols}c@{}}

\begin{subfigure}[t]{\teaserwidthb}
\includegraphics[width=\columnwidth]{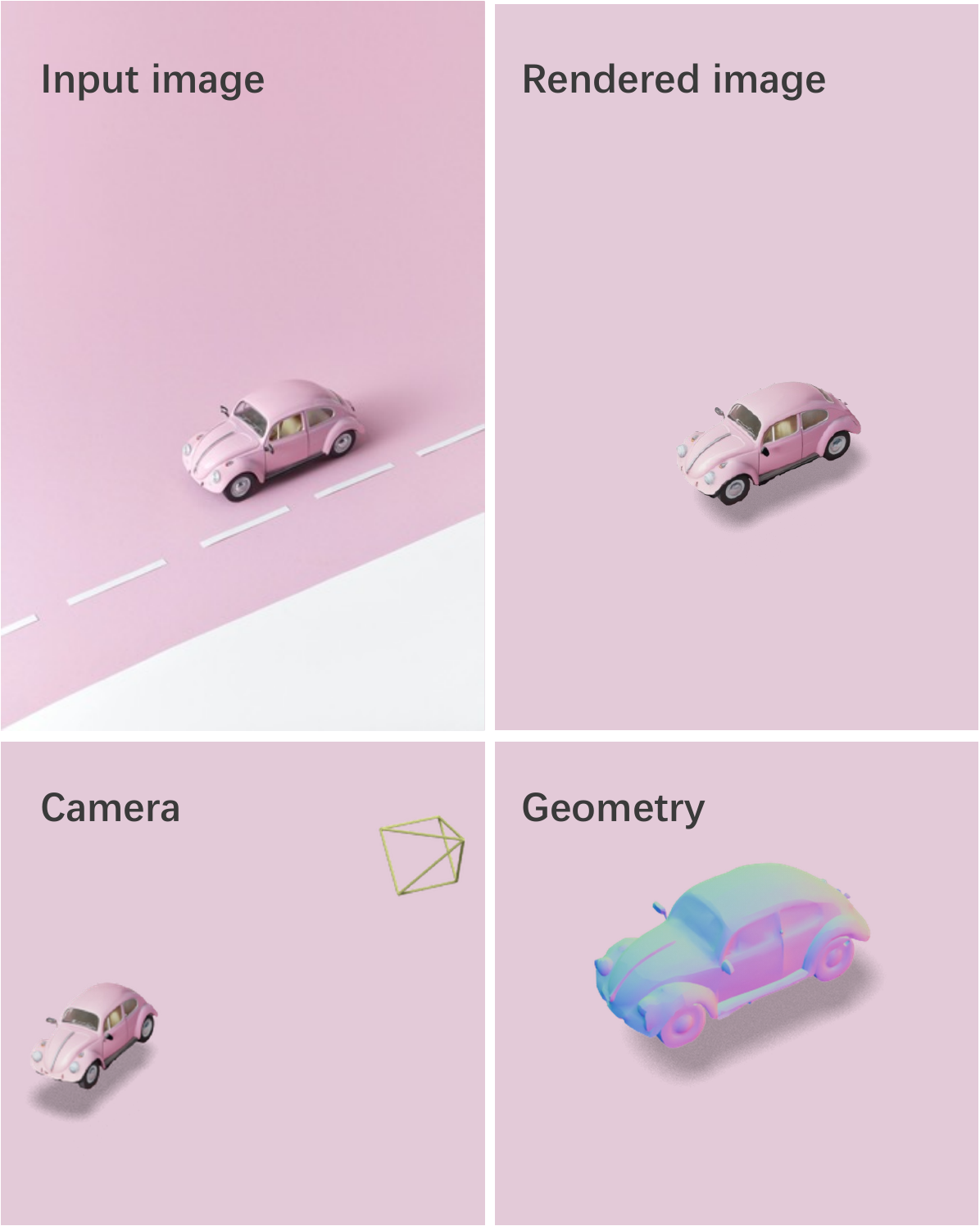}
\end{subfigure} &
\begin{subfigure}[t]{\teaserwidthb}
\includegraphics[width=\columnwidth]{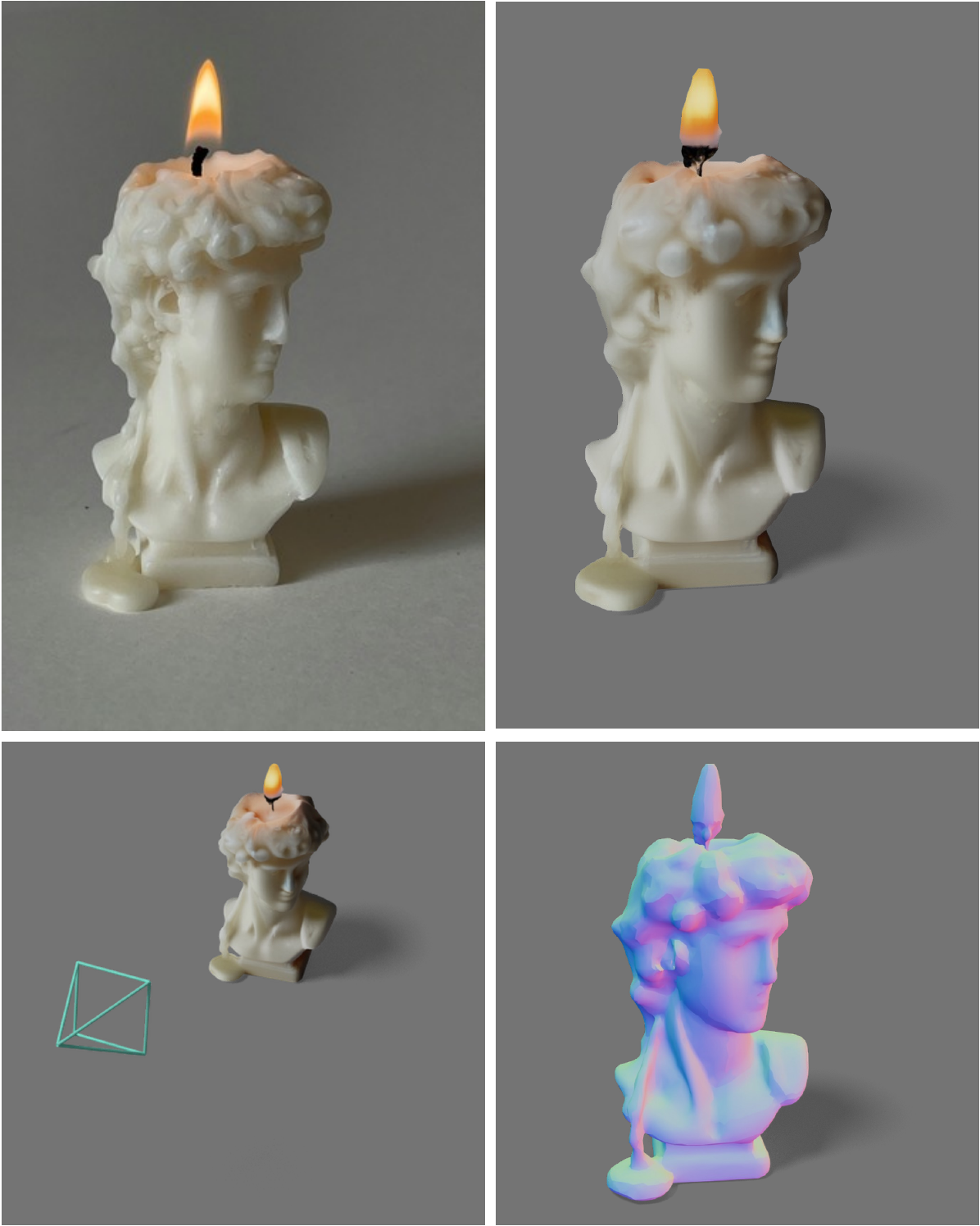}
\end{subfigure} &
\begin{subfigure}[t]{\teaserwidthb}
\includegraphics[width=\columnwidth]{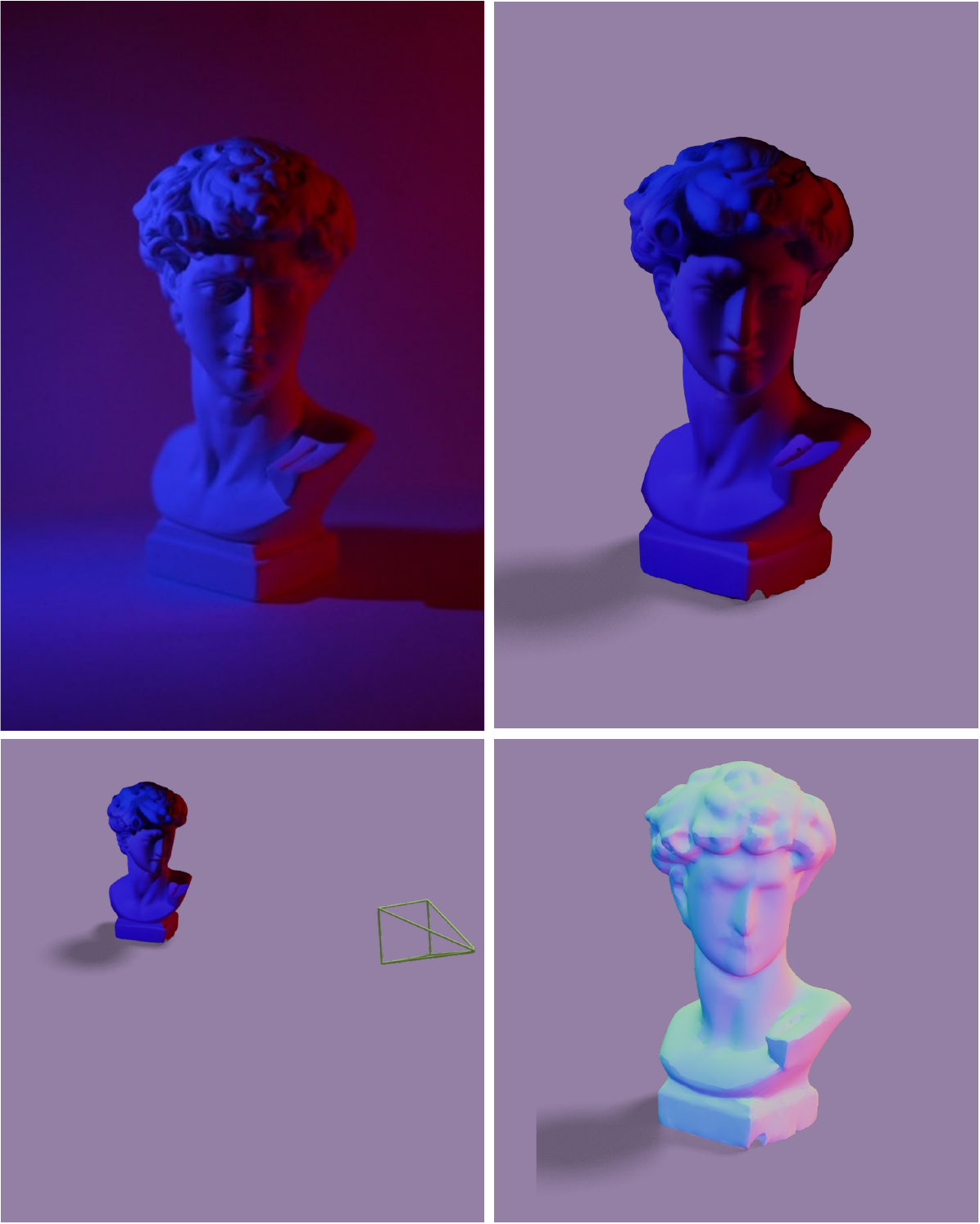}
\end{subfigure} &
\begin{subfigure}[t]{\teaserwidthb}
\includegraphics[width=\columnwidth]{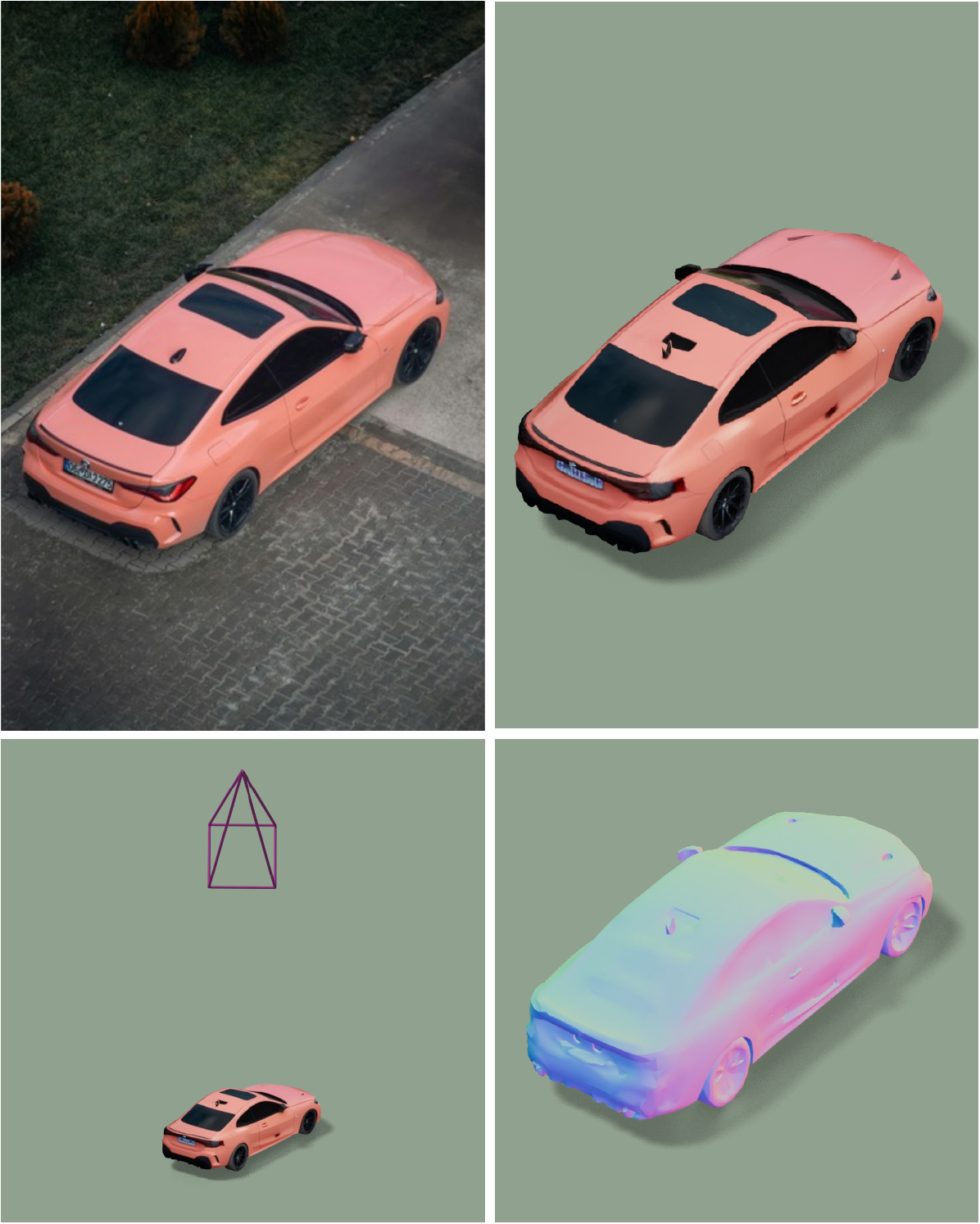}
\end{subfigure} \\
\begin{subfigure}[t]{\teaserwidthb}
\includegraphics[width=\columnwidth]{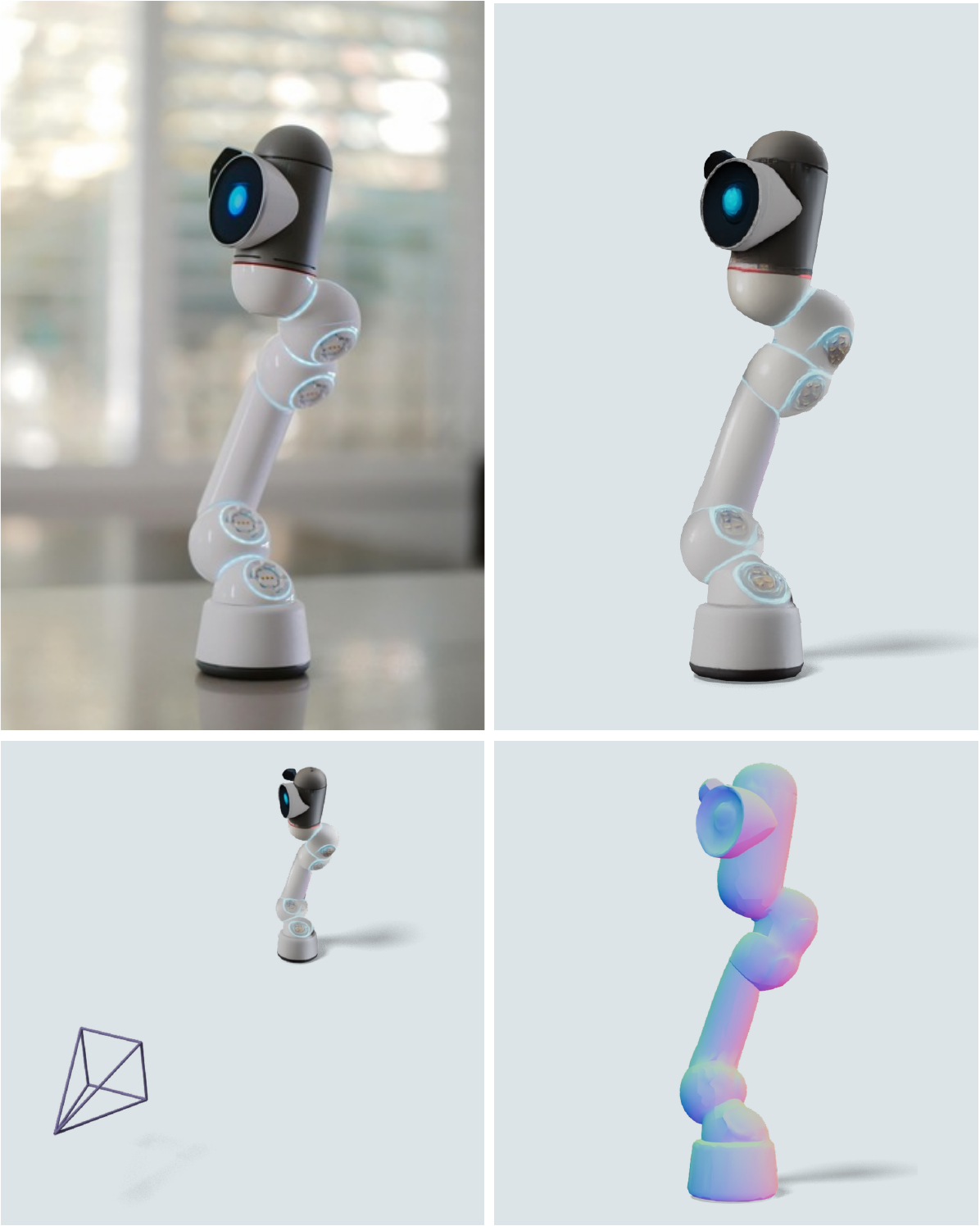}
\end{subfigure} &
\begin{subfigure}[t]{\teaserwidthb}
\includegraphics[width=\columnwidth]{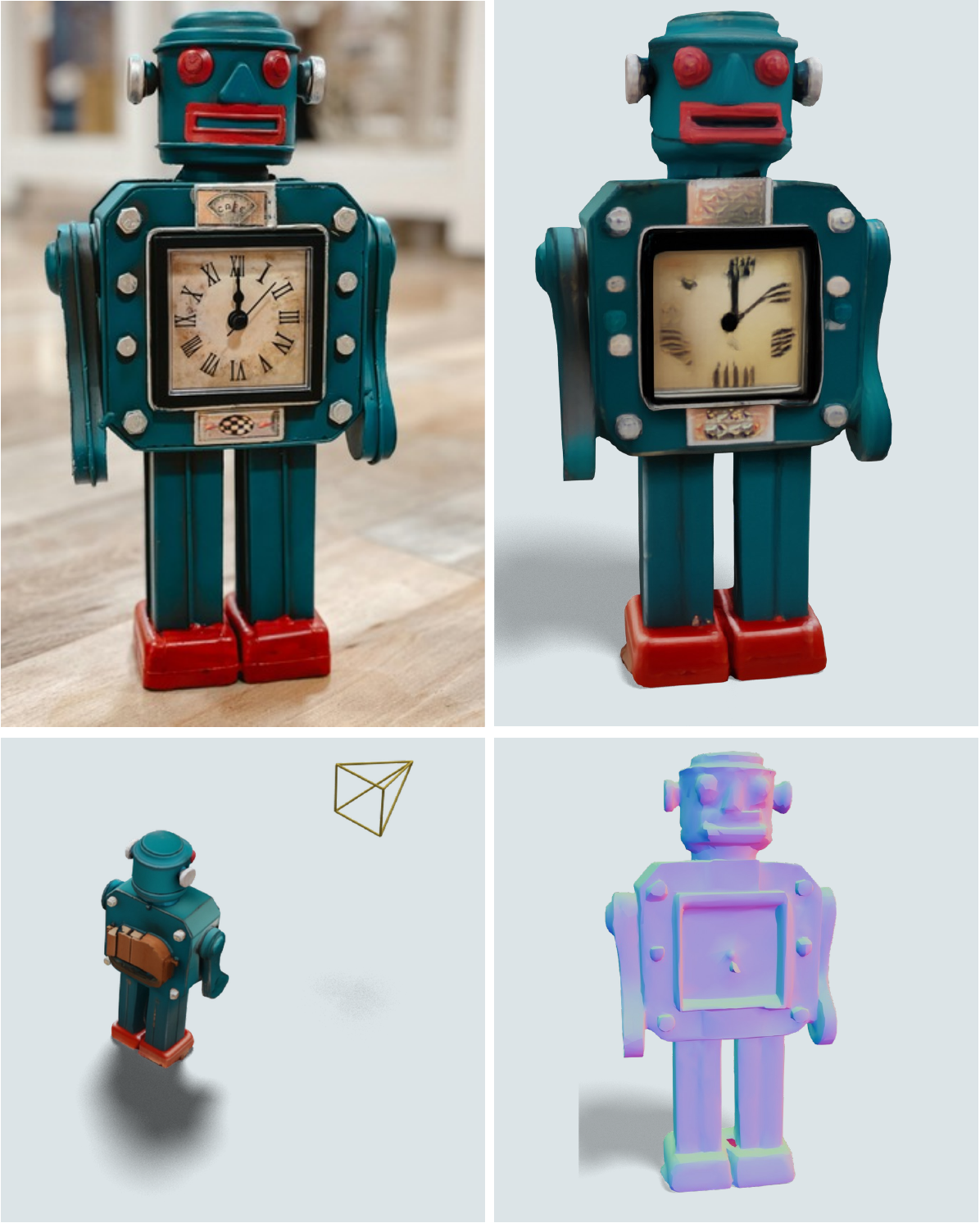}
\end{subfigure} &
\begin{subfigure}[t]{\teaserwidthb}
\includegraphics[width=\columnwidth]{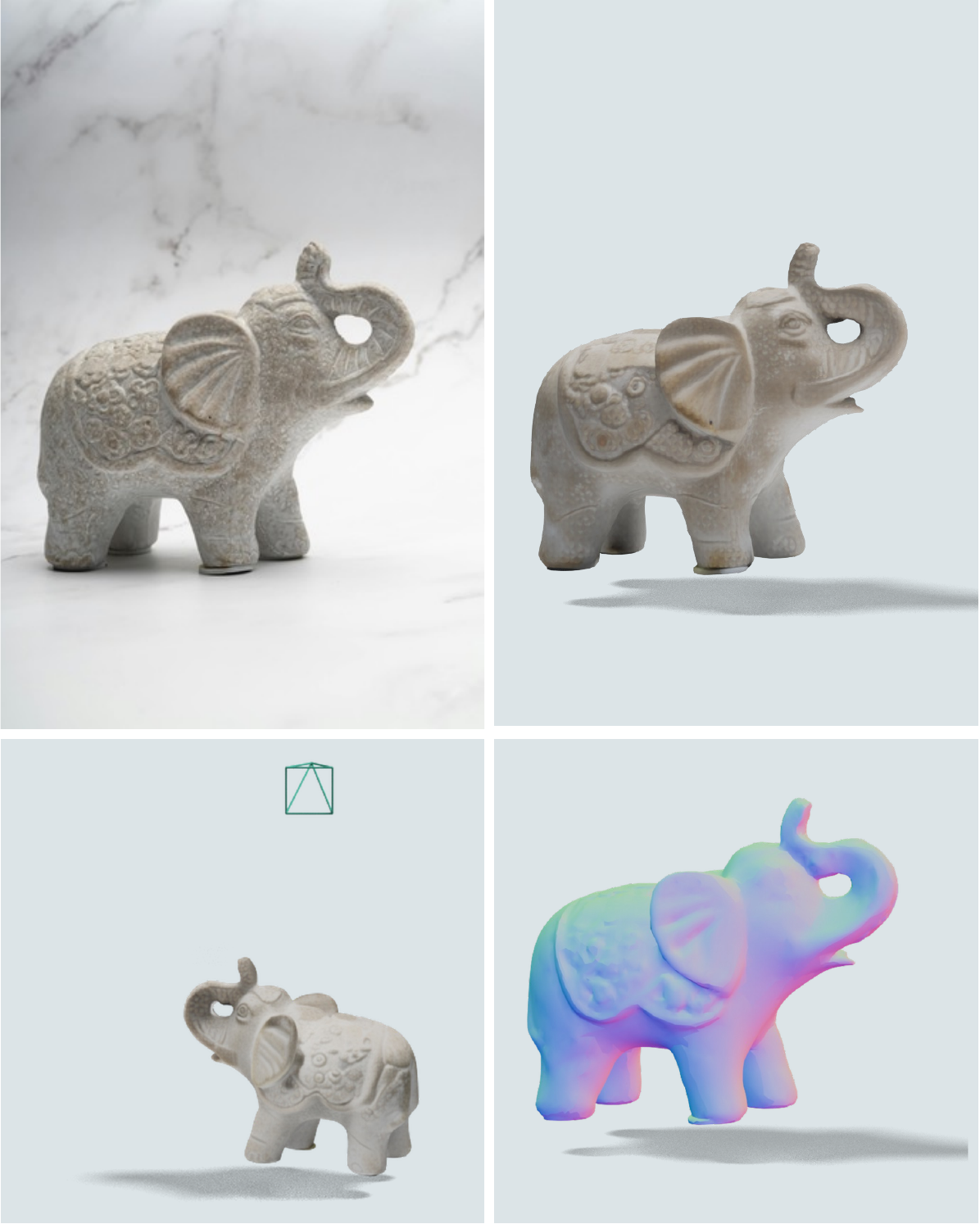}
\end{subfigure} &
\begin{subfigure}[t]{\teaserwidthb}
\includegraphics[width=\columnwidth]{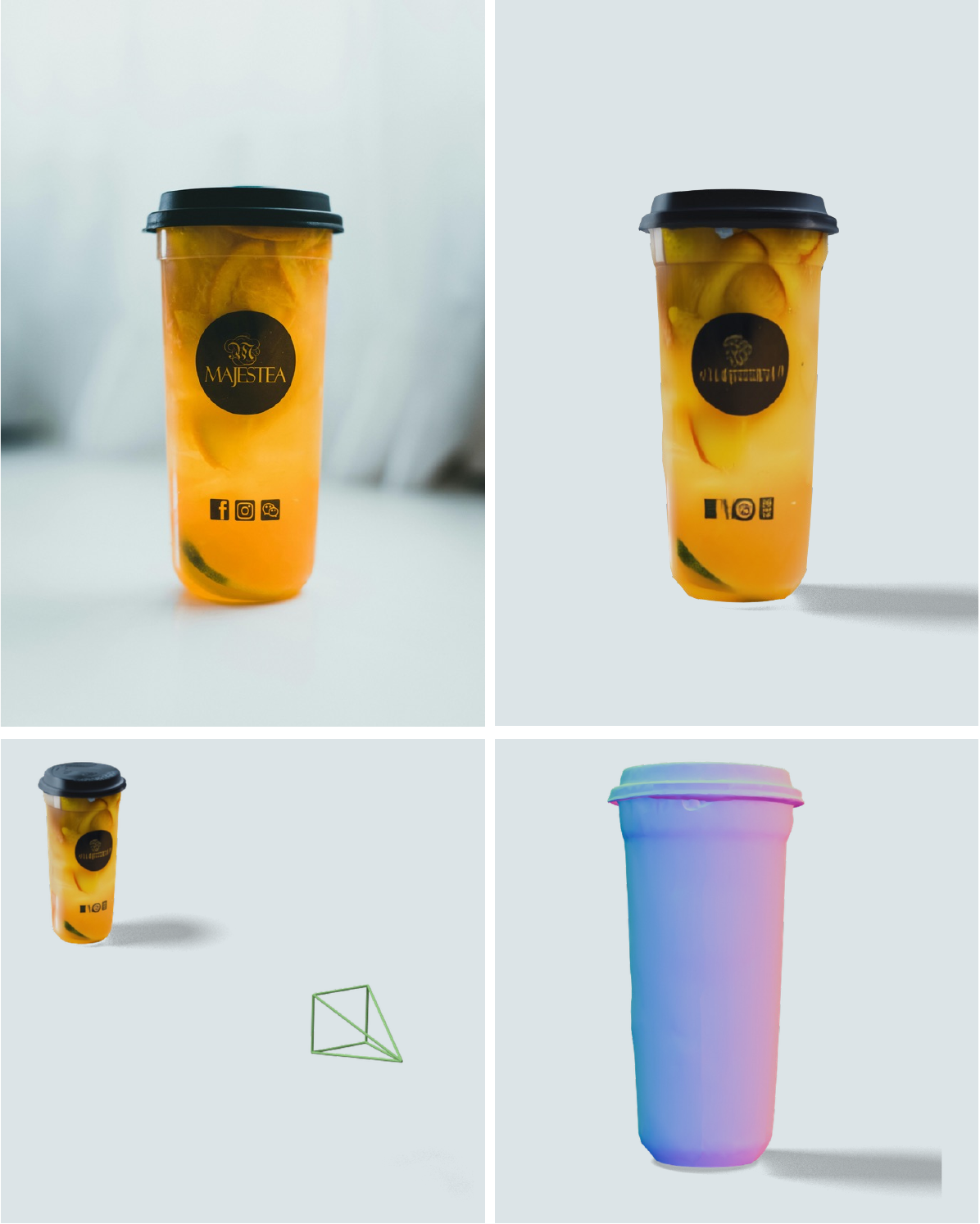}
\end{subfigure} \\ 
\begin{subfigure}[t]{\teaserwidthb}
\includegraphics[width=\columnwidth]{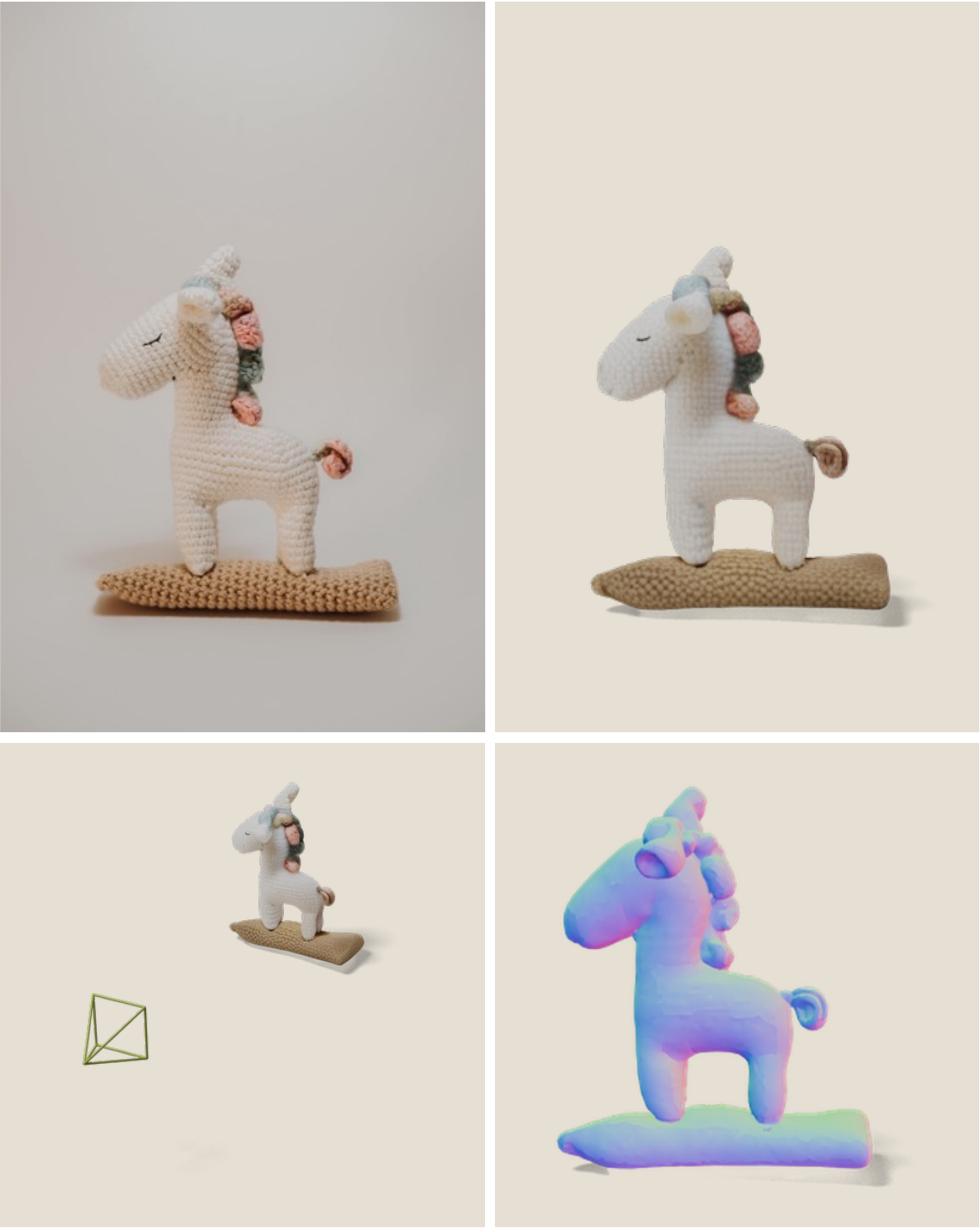}
\end{subfigure} &
\begin{subfigure}[t]{\teaserwidthb}
\includegraphics[width=\columnwidth]{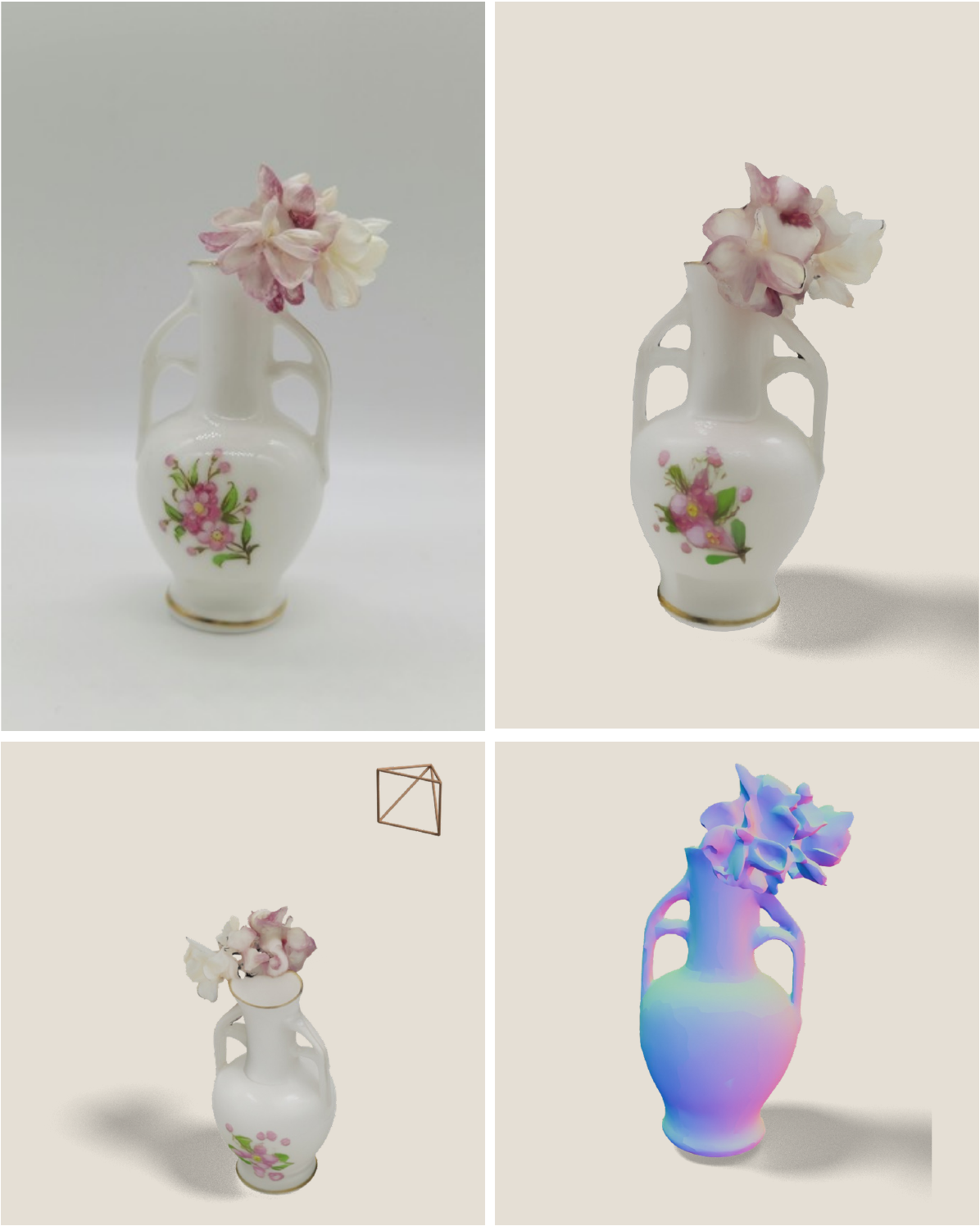}
\end{subfigure} &
\begin{subfigure}[t]{\teaserwidthb}
\includegraphics[width=\columnwidth]{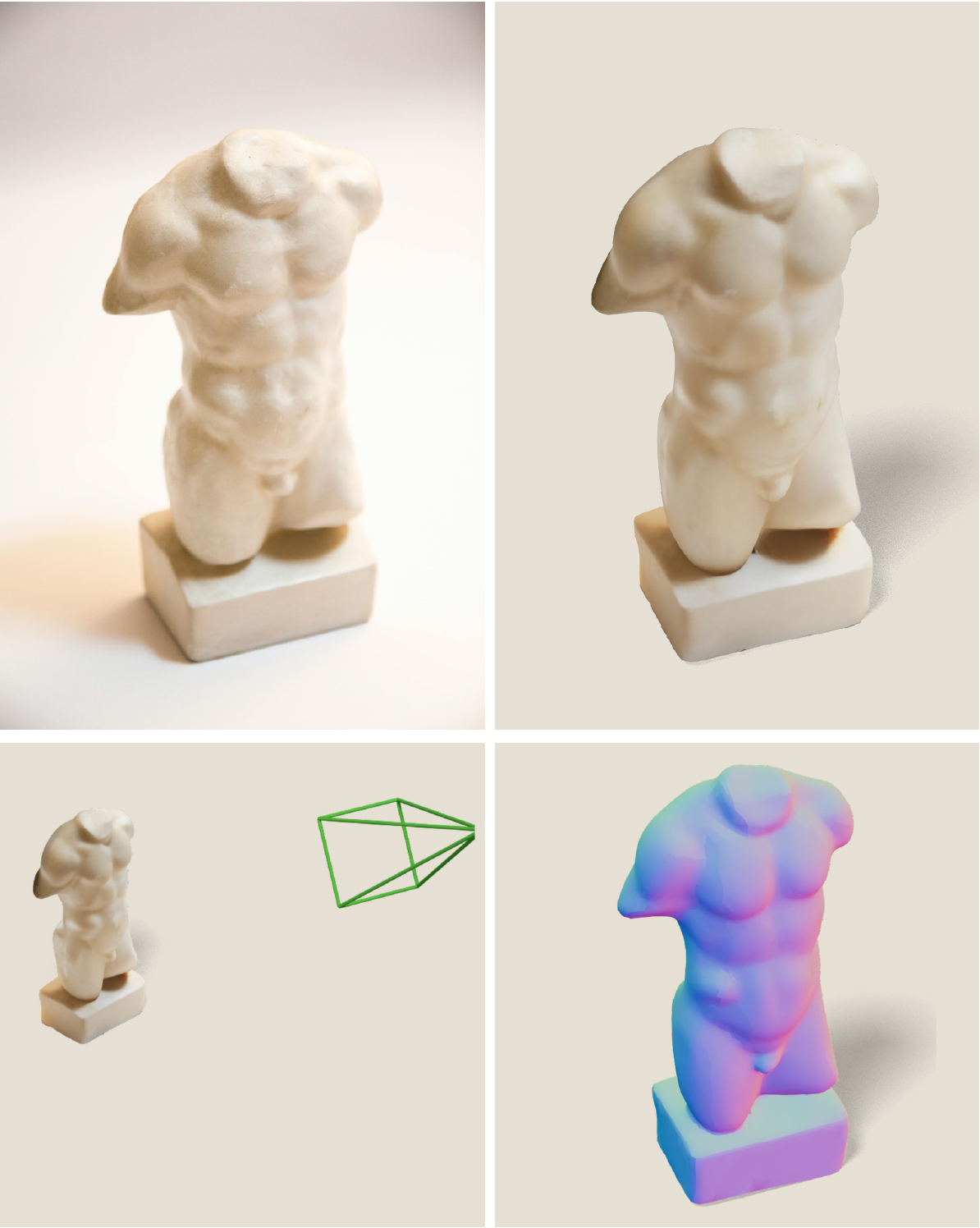}
\end{subfigure} &
\begin{subfigure}[t]{\teaserwidthb}
\includegraphics[width=\columnwidth]{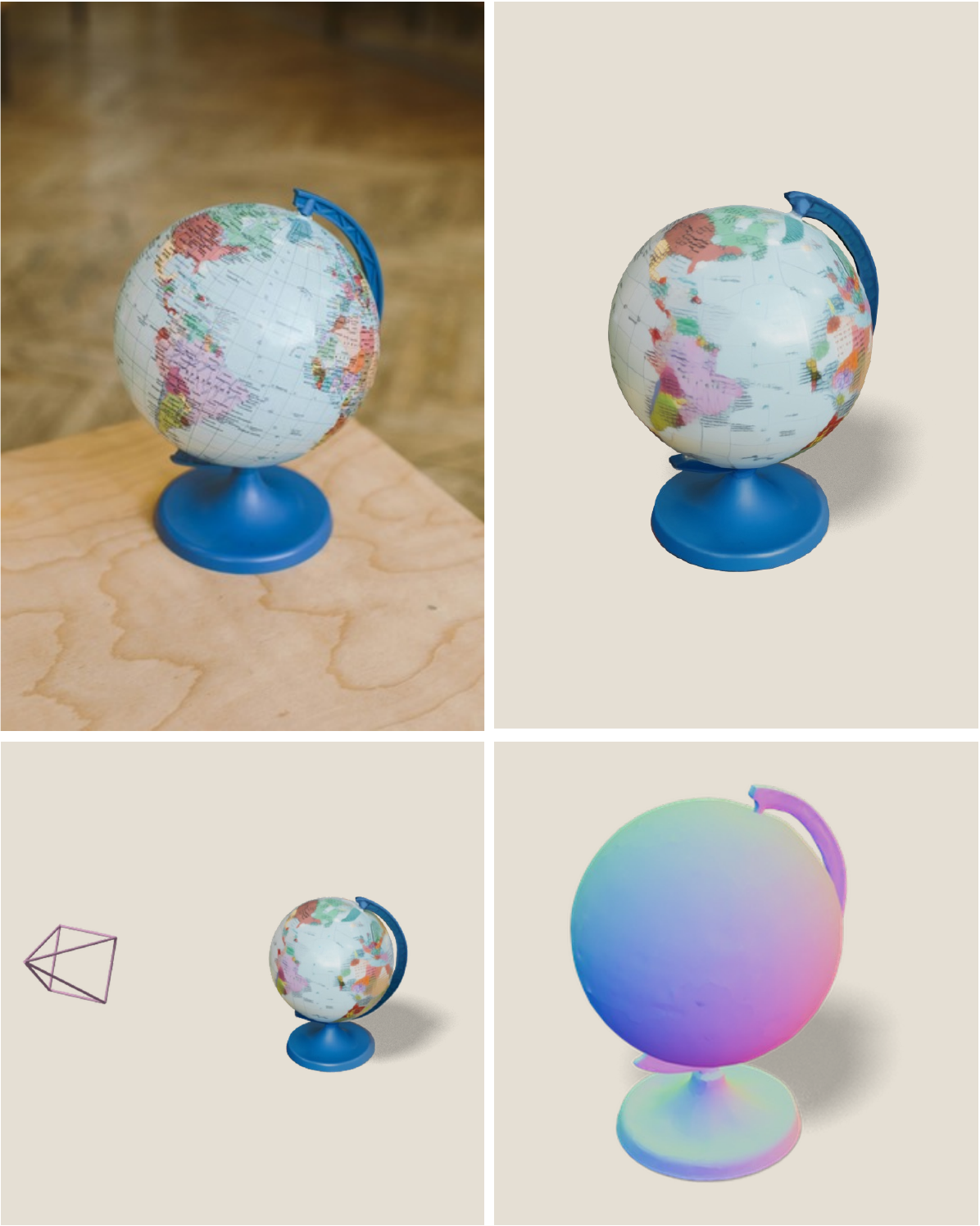}
\end{subfigure} \\ 

\multicolumn{4}{c}{%
  \makebox[0pt]{\color{gray}\hdashrule{4\teaserwidthb}{0.5pt}{4pt 2pt}}%
} \\ [0.1em]

\begin{subfigure}[t]{\teaserwidthb}
\includegraphics[width=\columnwidth]{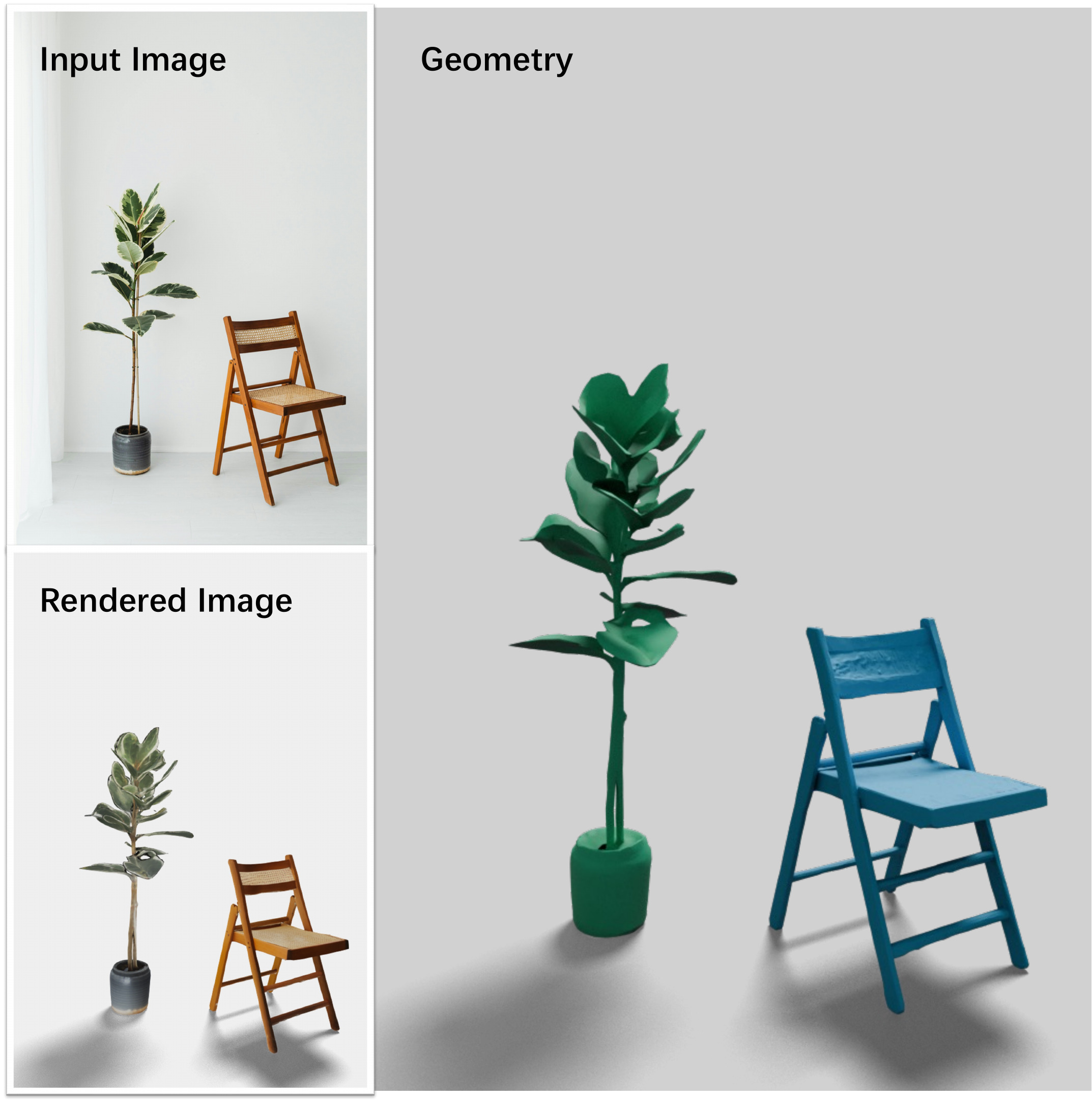}
\end{subfigure} & 
\begin{subfigure}[t]{\teaserwidthb}
\includegraphics[width=\columnwidth]{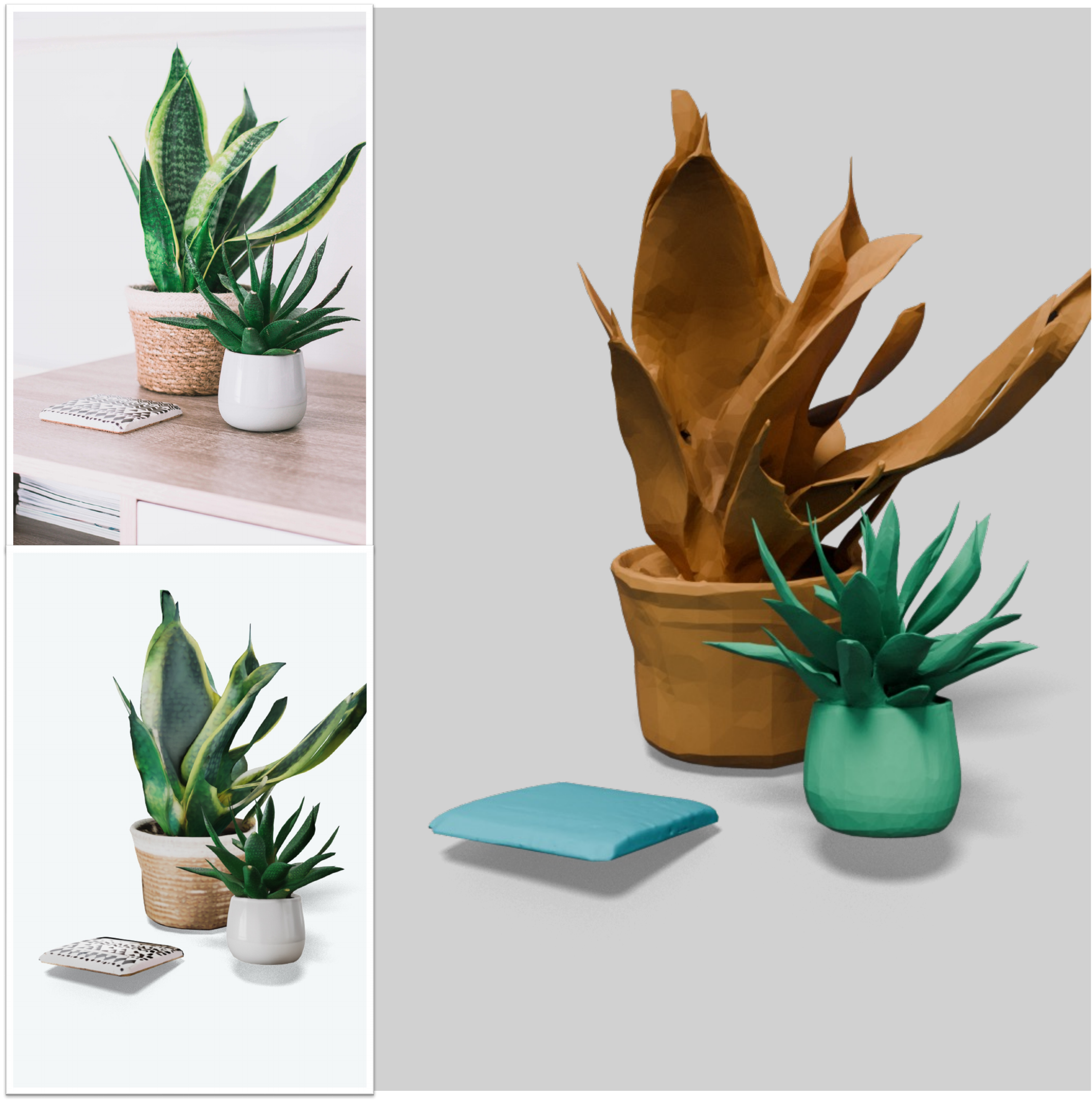}
\end{subfigure} &
\begin{subfigure}[t]{\teaserwidthb}
\includegraphics[width=\columnwidth]{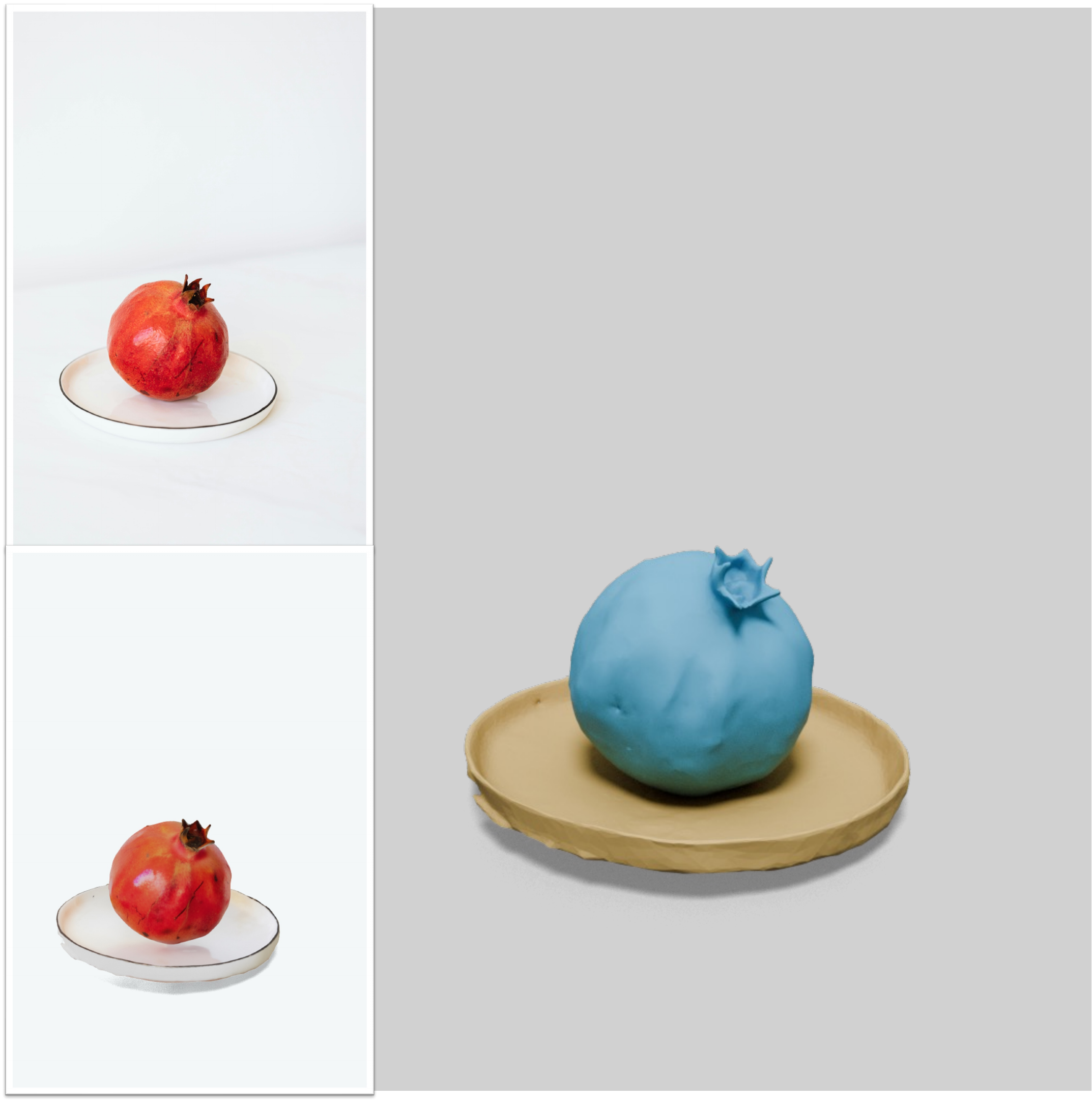}
\end{subfigure} &
\begin{subfigure}[t]{\teaserwidthb}
\includegraphics[width=\columnwidth]{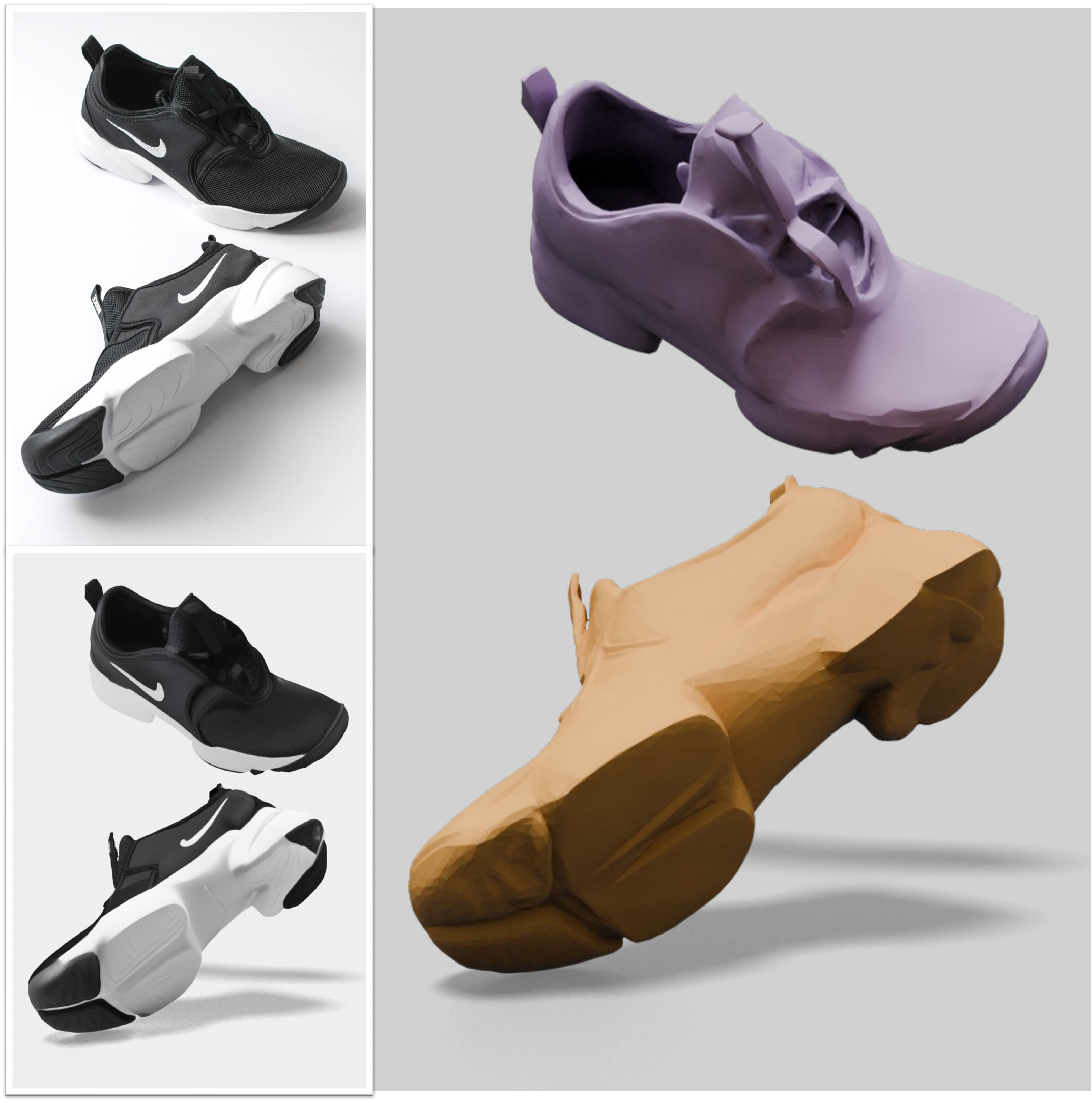}
\end{subfigure}  \\
\end{tabular}
}
\caption{
\textbf{Results for \textit{generative 3D reconstruction} from a single test image.} Given an input image (top left), \Cupid estimates camera pose (bottom left) and reconstructs 3D model (bottom right), re-rendering the input (top right). It is robust to changes in scale, placement, and lighting while preserving fine details, and supports component-aligned scene reconstruction (bottom row). All results are produced in seconds via feed-forward sampling of the learned model. 
See \href{https://cupid3d.github.io/}{\texttt{cupid3d.github.io}} for an immersive view of the interactive 3D results.
}
\label{fig:teaser}
\end{figure*}

Observing the gap between these two lines of work, we argue an ideal system should \textbf{jointly model both components within a unified framework},
deriving this joint distribution from a large set of 2D observations without fixing camera pose as identity or specifying how the canonical object is oriented\footnote{
3D artist-curated training data is typically placed in a display-ready frame that naturally encodes human priors on canonicalness. Our learned representation captures this dominant mode without explicit labeling.
}.
This mirrors how humans perceive the world: as illustrated in \Figref{fig:teaser_small}, we readily infer that the image captures the plush toy from its left frontal side. This implies that, through exposure to abundant visual data,  we maintain a view-agnostic mental impression of 3D objects while simultaneously recalling viewpoint that would most reproduce a given 2D observation. This formulation essentially bridges 3D reconstruction and 3D generation into a unified task, which we call \textbf{\textit{Generative 3D Reconstruction}}.

To this end, we introduce \Cupid (for ``in-CUbe PIxel Distribution''), a probabilistic reconstruction framework designed to achieve this synthesis by simultaneously generating a full 3D object $\object$ and reasoning the exact camera pose $\pose$ that can \textit{faithfully} reproduce the 2D observation. \Cupid formulates reconstruction as a conditional sampling process within a unified flow-based model that operates in two stages. First, it jointly generates a coarse 3D structure and a dense field of 3D-to-2D correspondences, enabling camera pose recovery through a simple PnP solver~\citep{abdel2015direct}. Second, conditioned on this recovered pose, a refinement stage injects \textit{pixel-aligned features} from the input image directly into the generative process. This novel conditioning mechanism enforces consistency with the visual evidence, marrying the rich prior of a generative model with the geometric fidelity of reconstruction. As \Figref{fig:teaser} shows, this simple yet effective approach produces geometrically complete 3D reconstructions that are texturally faithful to the input image.

The primary contributions of this work can be summarized as follows:
\begin{itemize}[leftmargin=*]
    \item \textbf{A unified generative reconstruction framework.} \Cupid bridges 3D generation and reconstruction by jointly modeling a canonical 3D object and camera pose. This unified design excels at single-view reconstruction, and can be applied to multi-view and component-aligned scene reconstruction without task-specific tuning or optimization.
    \item \textbf{A novel pose-conditioned refinement strategy.} Our two-stage approach uses the recovered pose to inject pixel-local features, a mechanism critical for achieving high-fidelity reconstructions free from the shape and color inconsistencies typical of generative models.
    \item \textbf{Strong object-level reconstruction performance.} 
     \Cupid leverages its generative capabilities to set a new standard in reconstruction fidelity, outperforming top holistic reconstruction models~\citep{hong2023lrm} by over 3 dB PSNR while matching the geometric accuracy of state-of-the-art monocular estimators~\citep{wang2025vggt,wang2025moge}.
\end{itemize}
\section{Related Work}

\paragraph{3D reconstruction from many images.}
Complete 3D reconstruction traditionally requires multiple views and Structure-from-Motion~\citep{schoenberger2016sfm,schoenberger2016mvs}. DUSt3R~\citep{wang2024dust3r} predicts pixel-aligned point maps for image pairs, enabling efficient recovery of view-centric camera poses and partial geometry; subsequent works improve efficiency and flexibility~\citep{wang2025vggt,wang2025moge2,Qianqian_2025_CVPR,zhang2025flarefeedforwardgeometryappearance,Yang_2025_Fast3R,karaev24cotracker3,depth_anything_v2,slam3r,must3r_cvpr25}. CUT3R processes images recurrently~\citep{Qianqian_2025_CVPR}; VGGT jointly predicts depth, point maps, and poses~\citep{wang2025vggt}; and MoGe introduces affine-invariant point maps for monocular geometry~\citep{wang2025moge}. 
These approaches lack generative capabilities and cannot plausibly infer occluded regions, resulting in partial geometry from limited input views. Even with abundant views, they produce a single fused scene without object separation. In contrast, our method models a joint distribution over 3D objects and camera pose, enabling completion of partial observations into fully compositional, object-level scenes.

\vspace{-2mm}
\paragraph{3D reconstruction from one or a few images.} 
High-quality multiview data are hard to obtain in practice, motivating single- or sparse-view reconstruction with large models~\citep{xu2024sparp,wei2024meshlrm,liu2024meshformer,TripoSR2024,tang2024lgm,xu2024grm,xu2024freesplatter}. LRM~\citep{hong2023lrm} shows that complete 3D can be regressed from a single image under a fixed camera. PF-LRM~\citep{wang2023pf} enables pose-free sparse-view reconstruction via differentiable PnP~\citep{chen2022epro}. Subsequent work improves single-view quality by augmenting novel views from 2D generative priors and using more efficient representations~\citep{xu2024instantmesh,xu2024grm,liu2023one2345} or architectures~\citep{LaRa}. 
Yet their entangled \textit{pixel-aligned} representations never decouple 3D content and viewpoint, typically assuming a fixed camera (e.g., unit focal length), making them poorly suited to real-world images with unknown camera parameters. Their deterministic formulations cannot model uncertainty in unseen regions, failing to produce plausible 3D variations. Our generative framework instead enables synthesizing diverse, high-fidelity 3D objects under flexible conditioning.

\begin{figure*}[!t]
\centering
\includegraphics[width=\textwidth]{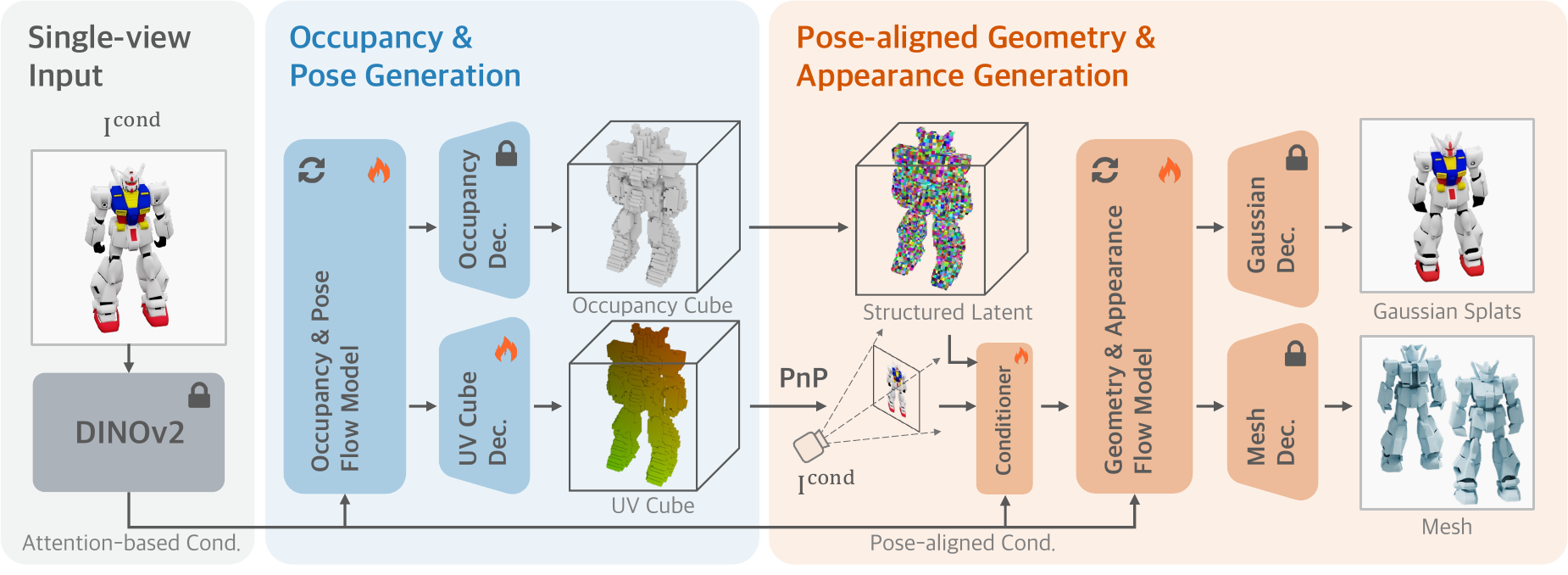}
\caption{\textbf{Overview of \Cupid's pipeline}. From a single input image, \Cupid first generates an occupancy and a UV cube in canonical space. Then, a \textit{Perspective-n-Point (PnP)} solver (\emph{i.e.}, \Eqref{eq:pnp}) recovers the camera pose. Using this recovered camera pose, we extract pose-aligned conditioning latents and visual features (\emph{i.e.}, \Eqref{eq:latent}), along with noisy structured latents, to generate the geometry and appearance features, which will be decoded to the 3D Gaussian splats and mesh.}
\label{fig:method}
\end{figure*}

\vspace{-2mm}
\paragraph{3D reconstruction from generative priors.} 
Dreamfusion~\citep{poole2022dreamfusion} pioneers 3D generation from 2D diffusion models~\citep{rombach2021highresolution,saharia2022photorealistic}. Zero-1-to-3~\citep{liu2023zero1to3} and subsequent work~\citep{shi2023zero123++,watson2022novel,li2024instant3d,liu2024syncdreamer,gao2024cat3d,hoellein2024viewdiff,wu2024reconfusion} fine-tune image diffusion models for single image-to-3D generation. 
Recent methods~\citep{zhou2025stable, ren2025gen3c, song2025historyguidedvideodiffusion} leverage video diffusion priors to achieve greater view consistency. Nevertheless, they rely on 2D generative priors that require dense supervision to reconstruct 3D objects and inherit the limitations of many-image-to-3D, most notably their difficulty in producing compositional scenes. Recent work~\cite{huang2025midi,yao2025castcomponentaligned3dscene,dong2025hiscene} incorporate native 3D generative priors~\cite{xiang2025structured,zhang2024clay} for scene reconstruction. Yet, without explicitly modeling camera pose, these methods resort to costly and fragile post-hoc optimization or additional tuning on spatial layouts~\cite{huang2025midi}. Notably, CAST~\cite{yao2025castcomponentaligned3dscene} designs a multi-stage pipeline for spatial alignment: it first estimates a point cloud~\cite{wang2025moge} then aligns each generated object via 3D–3D correspondences. In contrast, our approach achieves image-to-3D alignment in a single forward pass while delivering monocular geometry quality on par with MoGe~\cite{wang2025moge}, enabling streamlined compositional scene reconstruction without complex pipelines.

\section{Method}

\subsection{Problem definition and method overview}
We formulate our generative reconstruction as estimating the joint posterior $p(\object, \pose \mid \image)$ under the observation constraint $\image = \render(\object, \pose)$, where $\image$ is the input image, $\object$ is the 3D object, $\pose$ is the object-centric camera pose, and $\render(\cdot, \cdot)$ projects the 3D object to the image.

We first use an encoder $\encoder$ to map the 3D object and camera pose to a volumetric latent feature $\latent = \encoder(\object, \pose)$, which can be decoded into 3D representations (e.g., Gaussian splats or meshes) by specific decoders~\citep{shen2023flexible,kerbl20233d}. 
We use Rectified Flow~\citep{lipman2022flow} for the latent $\latent$ generation. The model interpolates between data $\latentzero$ and noise $\noise$ over time $\timestep$ via $\latentt = (1 - \timestep) \latentzero + \timestep \noise$. The reverse process follows a time-dependent velocity field $\velocity(\latentt, \image, \timestep) = \nabla_{\timestep} \latentt$ that drives noisy samples toward the data distribution. We parameterize $\velocity$ with a neural network $\velocity_{\parameter}$ and train it using the Conditional Flow Matching (CFM) objective:
\begin{equation}
    \mathcal{L}_{\mathrm{CFM}}(\parameter) = \mathbb{E}_{\timestep, \latentzero, \noise} \left\| \velocity_{\parameter}(\latentt, \image, \timestep) - (\noise - \latentzero) \right\|_2^2.
\end{equation}

We first describe the representation of objects and camera poses in \Secref{sec:representation}. To enable efficient 3D generation, we utilize a cascaded flow modeling approach for $\velocity_\parameter$~\citep{xiang2025structured}. In the first stage, we generate an occupancy cube along with the camera pose. The second stage predicts the 3D shape and texture features in the occupied regions based on the outputs from the first stage. This process is detailed in \Secref{sec:flow_model}. The pipeline is illustrated in \Figref{fig:method}.

\begin{figure*}[t]
    \centering
    \includegraphics[width=\textwidth]{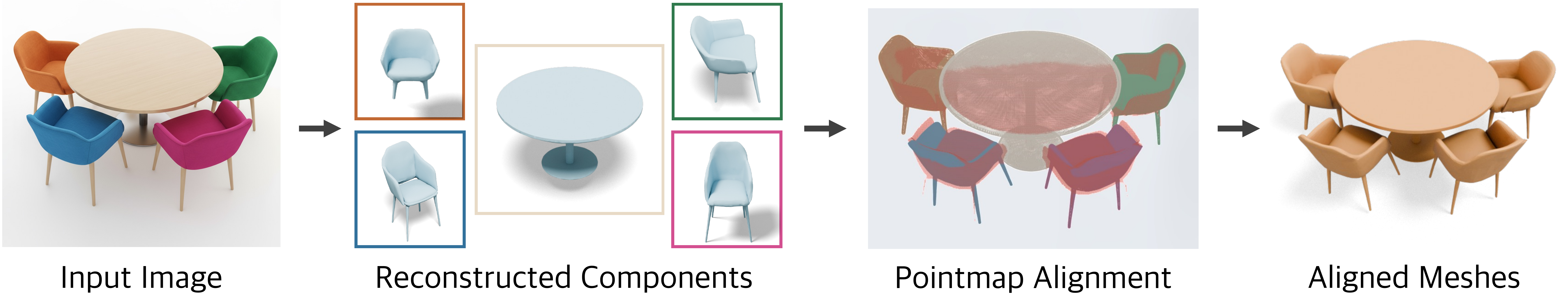}
    \caption{
    \textbf{Component-aligned scene reconstruction.}
    For a scene with multiple objects, our method can rebuild each object using the occlusion-aware 3D generator and then solve 3D–3D similarity transformation to accurately recompose the scene.}
    \label{fig:placeholder}
\end{figure*}

\subsection{Representation of 3D Object and camera pose}
\label{sec:representation}

\paragraph{3D object representation.}
To enable network training, we first tensorize 3D objects with a voxel-based representation: $\object \triangleq \{\point_i, \feat_i\}_{i=1}^{\numpts}$, where $\point_i \in \mathbb{R}^3$ represents the coordinates of the $i$-th active voxel in a cube, and $\feat_i$ encodes associated features derived from multi-view DINOv2~\citep{oquab2023dinov2} feature aggregation, subsequently compressed by a 3D variational autoencoder (VAE) encoder. Here, $\numpts$ denotes the number of voxels. This voxel set can be decoded to 3D formats such as mesh or Gaussian splat~\citep{xiang2025structured}.

\vspace{-2mm}
\paragraph{Camera pose parameterization.}
Camera pose $\pose$ is represented by the projection matrix $\mathbf{P} = \mathbf{K}[\mathbf{R}|\mathbf{t}] \in \mathbb{R}^{3\times4}$, where $\mathbf{K}$ is intrinsic, and $(\mathbf{R}, \mathbf{t})$ are extrinsic parameters mapping object to camera space.
Inspired by recent pose estimation work~\citep{wang2024dust3r,zhang2024cameras}, we propose an over-parameterized pose representation\footnote{While camera pose can be encoded as a compact 1D token (e.g., a 12D vector), it can be challenging for 3D generators with 3D tokens~\cite{zhang2024cameras}.} via an \textit{in-cube pixel distribution}. 
Specifically, we reparameterize $\pose$ as dense 3D-2D correspondence: $\pose \triangleq \{\point_i, \pixcoord_i\}_{i=1}^{\numpts}$ where $\pixcoord_i: (u_i,v_i) \in [0,1]^2$ is the normalized 2D pixel coordinates anchored at the 3D voxel $\point_i \in \object$. Each anchor’s pixel coordinate is obtained by 
$\pixcoord_i = \pi\big(\mathbf{P},\point_i\big)$, where $\pi$ denotes perspective projection. Such a reparameterized pose, \emph{i.e.}, 3D-2D correspondence, can be presented as a 3D UV cube.
Given these correspondences, we recover the global camera matrix $\mathbf{P}^{*}$ using a least-squares solver~\citep{abdel2015direct}:
\begin{equation}
\mathbf{P}^{*}
= \argmin_{\mathbf{P}} \sum_{i=1}^{\numpts} \big\Vert\pi(\mathbf{P},\point_i) - \pixcoord_i\big\Vert^2.
\label{eq:pnp}
\end{equation}
We then decompose $\mathbf{P}^{*}$ into $(\mathbf{K}, \mathbf{R}, \mathbf{t})$ via RQ decomposition. Unlike image-based pose representations (e.g., 2D ray or point maps~\citep{zhang2024cameras,wang2024dust3r}), we reparameterize pose within a 3D cube. Intuitively, the $(u,v)$ coordinates act like view-dependent colors defined on a 3D occupancy grid (see the UV cube in the \Figref{fig:method}). Under this view, joint object–pose generation transforms to producing a 3D object with view-dependent colors.

\subsection{Cascaded Flow Modeling}

\label{sec:flow_model}
We employ a two-stage cascaded flow model to jointly sample a 3D object and its corresponding camera pose, \emph{i.e.}, $\latent = \{ (\point_i, \feat_i), (\point_i,\pixcoord_i) \}_{i=1}^{\numpts}$. In the first stage, the flow model for occupancy and pose generation ($\sflow$) generates (i) an occupancy cube that indicates active voxels and (ii) a UV cube that encodes object-centric camera pose (\Secref{sec:representation}).
In the second stage, conditioned on the predicted occupancy and camera pose, the pose-aligned geometry and appearance generation model $\lflow$ synthesizes DINO features $\feat_i$ for each active voxel, yielding the final $\latent$.

\paragraph{Occupancy and pose generation.}
In this stage, given a conditioning image \(\image\), we generate an occupancy cube and a UV cube with resolution $r$. The occupancy cube \(\mathbf{G}_o \in \{0,1\}^{r \times r \times r \times 1}\) encodes binary values indicating active/inactive voxels. The UV cube \(\mathbf{G}_{uv} \in [0,1]^{r \times r \times r \times 2}\) stores normalized pixel coordinates $(u,v)$ for each voxel. Voxels sharing the same pixel coordinates lie on the same ray\footnote{We do not model occlusion as recent studies suggest that transformers can effectively model light transport~\citep{jin2024lvsm, zeng2025renderformer}.}. Next, we train a 3D VAE to compress the UV cube into a low-resolution feature grid \(\mathbf{S}_{uv} \in \mathbb{R}^{s\times s\times s \times C}\), improving efficiency with near-lossless pose accuracy (mean RRE / RTE $<$ 0.5°). For fine-tuning \trellis, we concatenate the original feature grid \(\mathbf{S}_o\) with \(\mathbf{S}_{uv}\) and incorporate linear layers at both the input and output of the flow network \(\sflow\). Once the occupancy and UV cubes are generated, we obtain $\{ (\point_i, \pixcoord_i(\pose)) \}_{i=1}^{\numpts}$ by collecting active voxels and solving the camera pose with \Eqref{eq:pnp}.

\paragraph{Pose-aligned geometry and appearance generation.}
In this stage, we generate the detailed 3D latent $\{ \feat_i \}_{i=1}^{\numpts}$ only at active voxels. Our experiments show that the original $\lflow$ from \trellis, which uses globally-attended image information, often suffers from color drift and loss of fine details (see \Figref{fig:color_drifting} in \Appref{sec:comparison_3dgen}). To address this, we leverage the calculated camera pose from the first stage to inject locally attended pixel information in each voxel.

Specifically, we leverage the estimated camera pose to compute the features located at the $ i$-th voxel as follows:
\begin{equation}
\begin{aligned}
\feat^\dino_{i} &= \interp(\pixcoord_i, \dino(\image)) \in \mathbb{R}^{1024}, \\
\{\feat^{\high}_{i}\}_{i=1}^{\numpts} &= \slatenc\big(\{\point_i, \feat^\dino_i\}_{i=1}^{\numpts}\big), \quad \feat^{\high}_{i} \in \mathbb{R}^{8},
\end{aligned}
\label{eq:latent}
\end{equation}
where each voxel's pixel coordinate $\mathbf{u}_i$ is obtained by projecting the coordinates of the $i$-th 3D voxel center onto the image plane with the calculated camera pose, $\interp$ denotes bilinear interpolation and $\slatenc$ is the 3D VAE encoder. While \dino captures high-level semantics, it loses low-level cues needed for precise 3D geometry and appearance reconstruction. To compensate, we extract complementary low-level features $\feat^{\low}$ from $\image$ with a lightweight convolutional head and sample them at $\pixcoord_i$ via $\interp$.
Finally, for each voxel, we concatenate the current noisy voxel feature $\feat_i^{t}$ at time step $t$ with the pixel-aligned features and fuse them into the flow transformer with a linear layer:
\(l_t = \mathrm{Linear}\big([\feat_i^{t}\,\oplus\, \feat^{\high}_{i} \,\oplus\, \feat^{\low}_{i}]\big)\).
This pose-aligned fusion significantly improves 3D geometry accuracy and appearance fidelity with respect to the input image.

\begin{table*}[t]
\renewcommand{\arraystretch}{1.0}
\centering
\caption{
\textbf{Monocular geometry accuracy.} \Cupid outperforms all 3D reconstruction and generation baselines and matches point-map regression methods that predict only partial geometry. Note that VGGT uses a ground-truth object mask, which may overestimate accuracy.
}
\label{tab:geometry}
\resizebox{0.8\textwidth}{!}{
\begin{tabular}{l c c c c c c | c c c c c}
\toprule
\multirow{2}{*}{\textbf{Method}} & \multirow{2}{*}{\textbf{3D}} &
\multicolumn{5}{c|}{\textbf{Toys4k}} &
\multicolumn{5}{c}{\textbf{GSO}} \\
& &
\textbf{mIOU} & \textbf{CD} & \textbf{CD} & \textbf{F-score} & \textbf{F-score} &
\textbf{mIOU} & \textbf{CD} & \textbf{CD} & \textbf{F-score} & \textbf{F-score} \\
& &
(avg)$\uparrow$ & (avg)$\downarrow$ & (med)$\downarrow$ & (0.01)$\uparrow$ & (0.05)$\uparrow$ &
(avg)$\uparrow$ & (avg)$\downarrow$ & (med)$\downarrow$ & (0.01)$\uparrow$ & (0.05)$\uparrow$ \\
\midrule
\textcolor{gray}{VGGT} & $\xmark$ &
\textcolor{gray}{--} & \textcolor{gray}{1.144} & \textcolor{gray}{0.498} & \textcolor{gray}{61.85} & \textcolor{gray}{95.90} &
\textcolor{gray}{--} & \textcolor{gray}{1.396} & \textcolor{gray}{0.388} & \textcolor{gray}{65.98} & \textcolor{gray}{95.95} \\
\textcolor{gray}{MoGe} & $\xmark$ &
\textcolor{gray}{92.80} & \textcolor{gray}{1.284} & \textcolor{gray}{0.581} & \textcolor{gray}{58.54} & \textcolor{gray}{95.31} &
\textcolor{gray}{96.18} & \textcolor{gray}{1.743} & \textcolor{gray}{0.575} & \textcolor{gray}{58.99} & \textcolor{gray}{94.68} \\ 
OnePoseGen & $\cmark$ &
9.34 & 153.2 & 59.92 & 6.11 & 24.10 &
12.16 & 116.2 & 60.56 & 7.28 & 25.77 \\
LaRa & $\cmark$ &
68.11 & 32.15 & 16.59 & 18.57 & 57.67 &
70.63 & 34.23 & 19.36 & 13.48 & 49.95 \\
OpenLRM & $\cmark$ &
86.26 & 2.726 & 1.291 & 40.42 & 90.60 &
91.35 & 3.741 & 1.858 & 34.14 & 87.20 \\
Ours & $\cmark$ &
\textbf{92.43} & \textbf{2.534} & \textbf{0.236} & \textbf{69.82} & \textbf{97.76} &
\textbf{95.27} & \textbf{1.823} & \textbf{0.434} & \textbf{61.01} & \textbf{95.59} \\
\bottomrule
\end{tabular}
}
\end{table*}

\newlength{\imgsize}  
\setlength{\imgsize}{1.8cm} 
\newlength{\labelcolw} 
\setlength{\labelcolw}{0.5cm}

\begin{figure*}[!t]
\centering
\setlength{\tabcolsep}{0pt}
\renewcommand{\arraystretch}{0.0}

\begin{tabular}{>{\centering\arraybackslash}m{\labelcolw} *{8}{>{\centering\arraybackslash}m{\imgsize}}}
\rotatebox{90}{\small Input}
& \includegraphics[width=\imgsize,height=\imgsize]{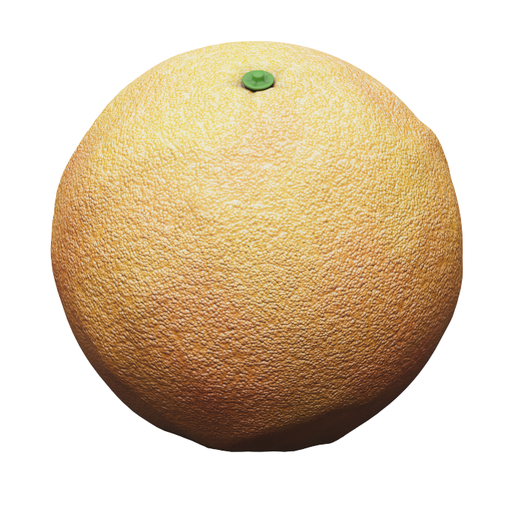}
& \includegraphics[width=\imgsize,height=\imgsize]{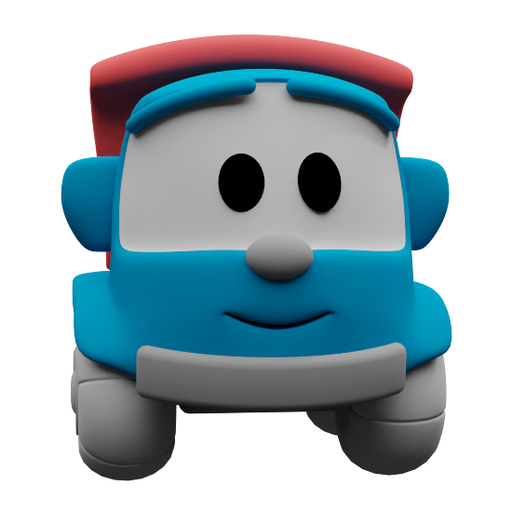}
& \includegraphics[width=\imgsize,height=\imgsize]{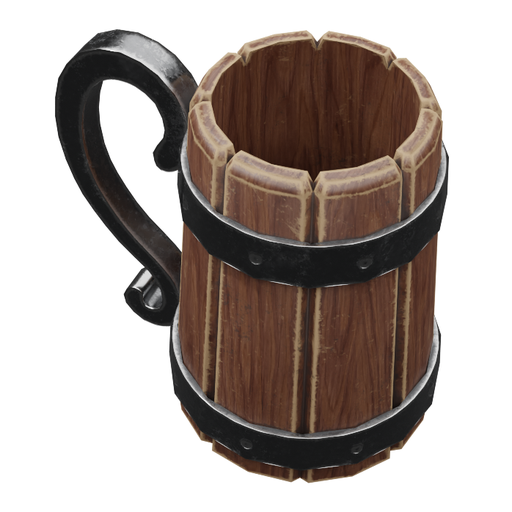}
& \includegraphics[width=\imgsize,height=\imgsize]{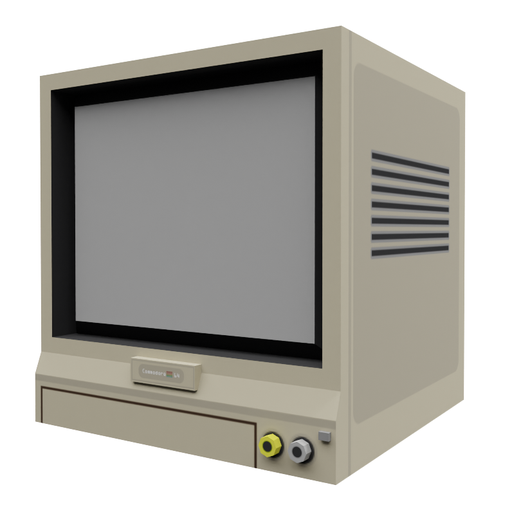}
& \includegraphics[width=\imgsize,height=\imgsize]{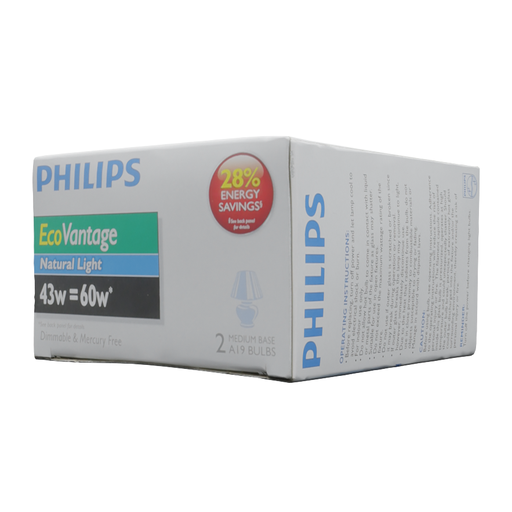}
& \includegraphics[width=\imgsize,height=\imgsize]{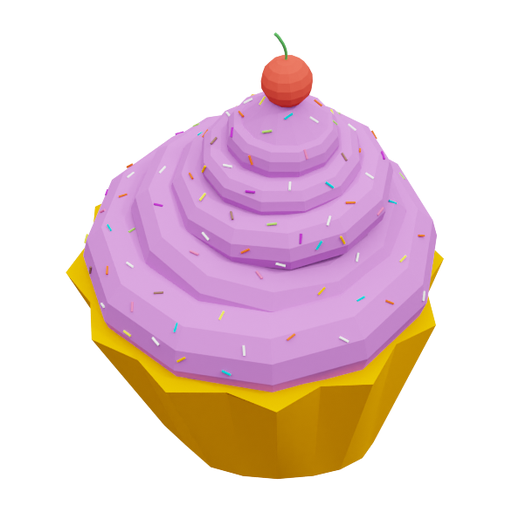}
& \includegraphics[width=\imgsize,height=\imgsize]{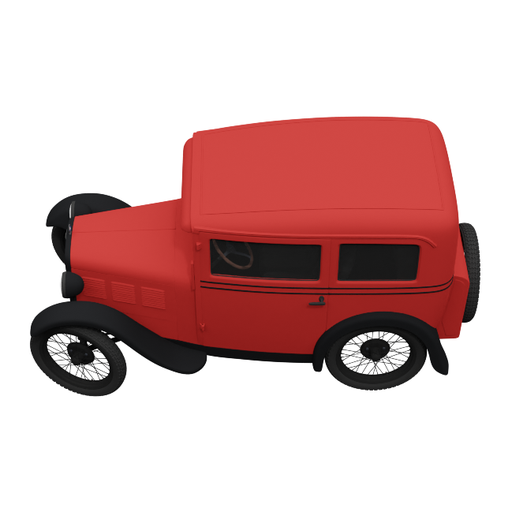}
& \includegraphics[width=\imgsize,height=\imgsize]{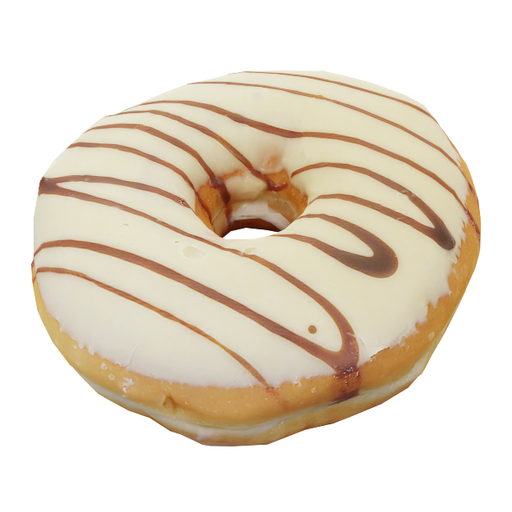}
\\
\rotatebox{90}{\small Ours}
& \includegraphics[width=\imgsize,height=\imgsize]{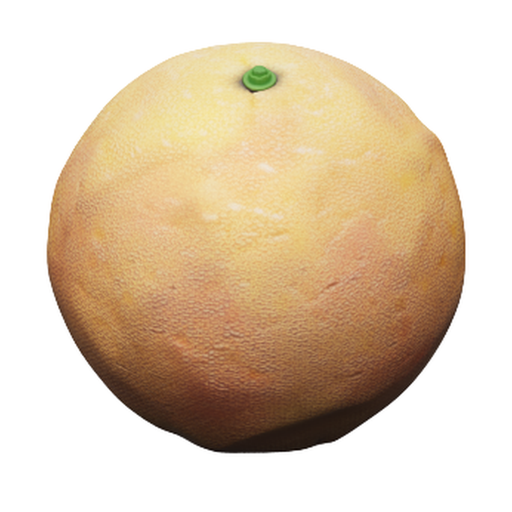}
& \includegraphics[width=\imgsize,height=\imgsize]{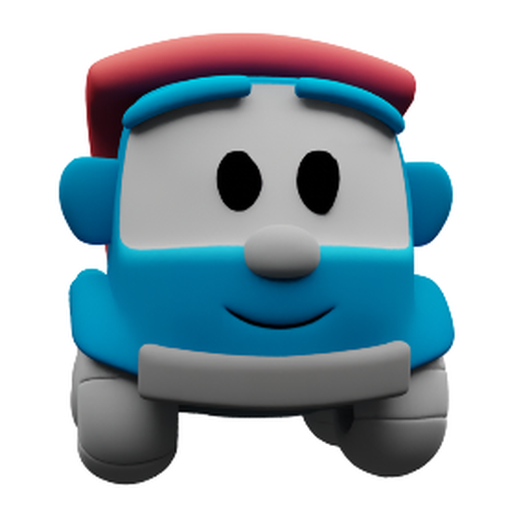}
& \includegraphics[width=\imgsize,height=\imgsize]{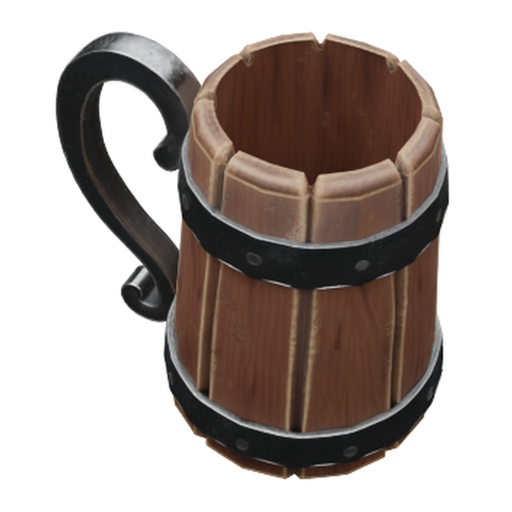}
& \includegraphics[width=\imgsize,height=\imgsize]{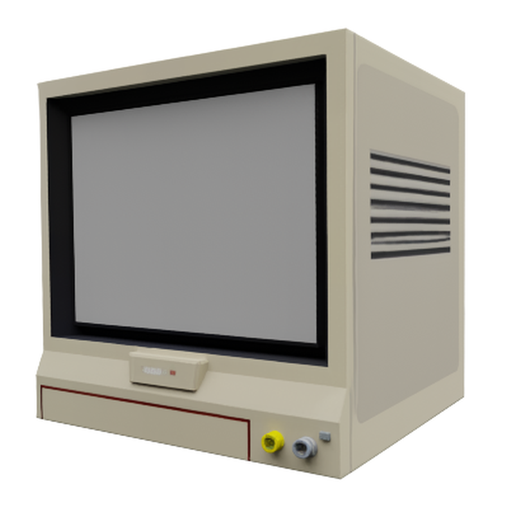}
& \includegraphics[width=\imgsize,height=\imgsize]{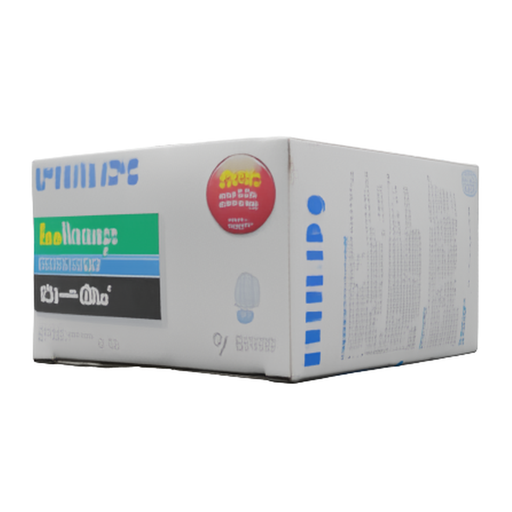}
& \includegraphics[width=\imgsize,height=\imgsize]{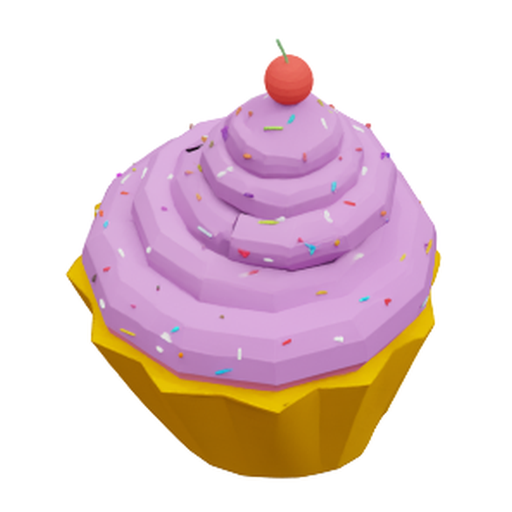}
& \includegraphics[width=\imgsize,height=\imgsize]{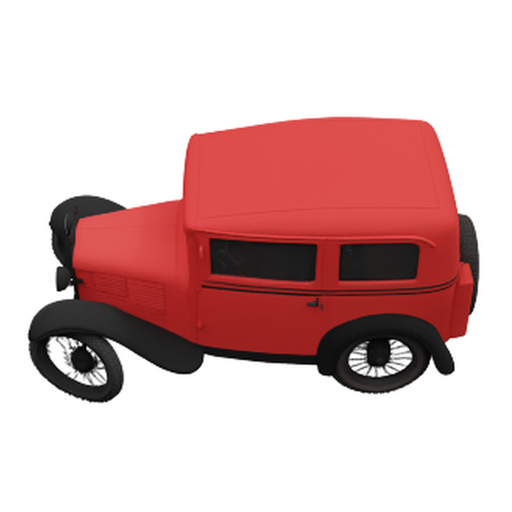}
& \includegraphics[width=\imgsize,height=\imgsize]{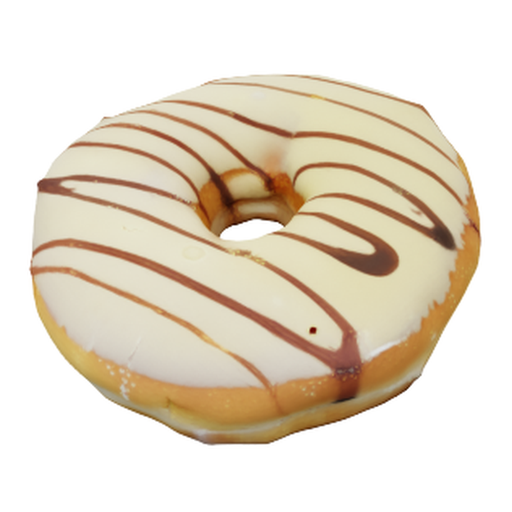}
\\
\rotatebox{90}{\small LRM}
& \includegraphics[width=\imgsize,height=\imgsize]{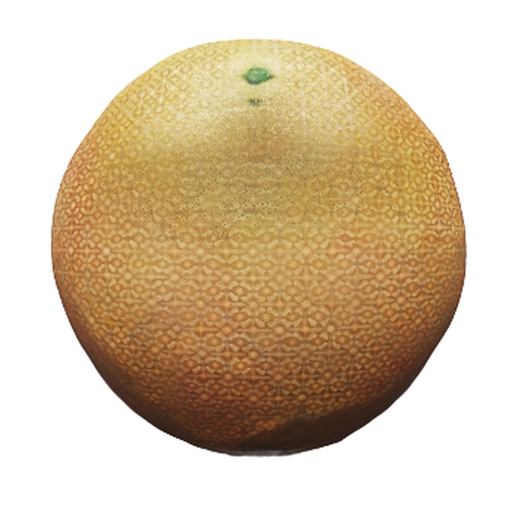}
& \includegraphics[width=\imgsize,height=\imgsize]{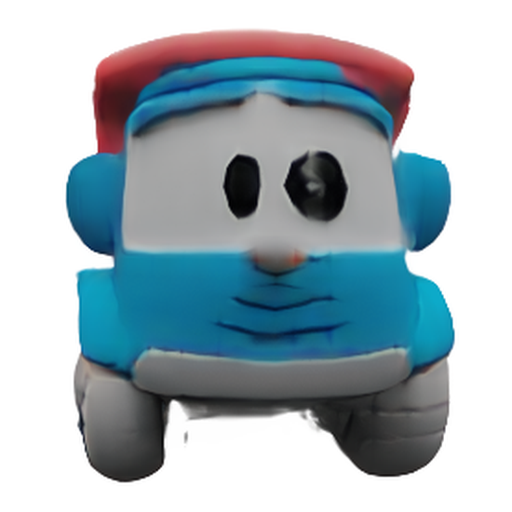}
& \includegraphics[width=\imgsize,height=\imgsize]{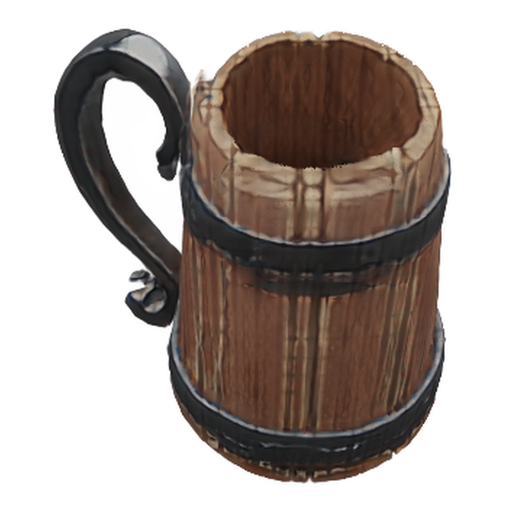}
& \includegraphics[width=\imgsize,height=\imgsize]{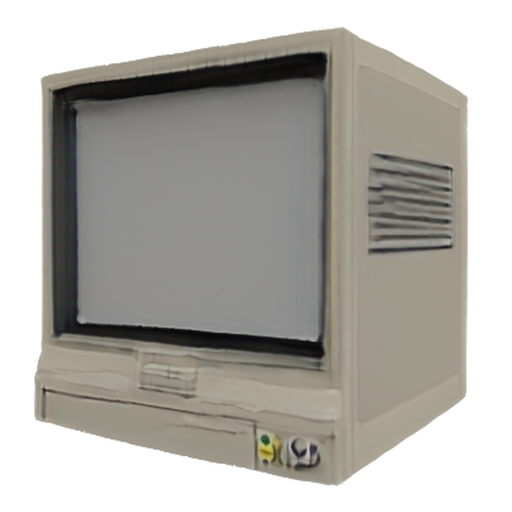}
& \includegraphics[width=\imgsize,height=\imgsize]{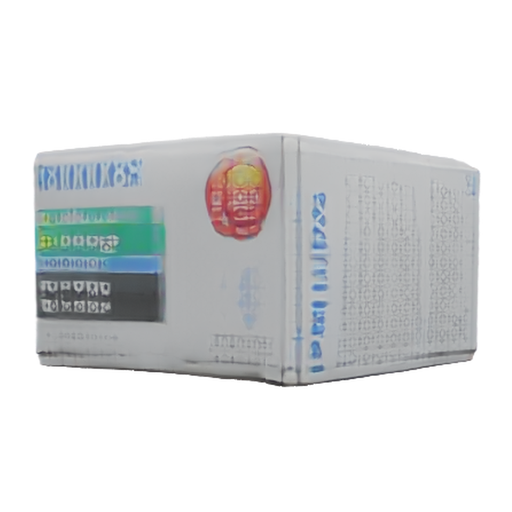}
& \includegraphics[width=\imgsize,height=\imgsize]{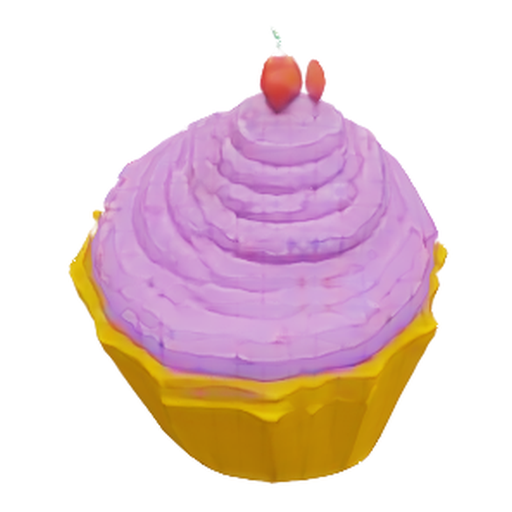}
& \includegraphics[width=\imgsize,height=\imgsize]{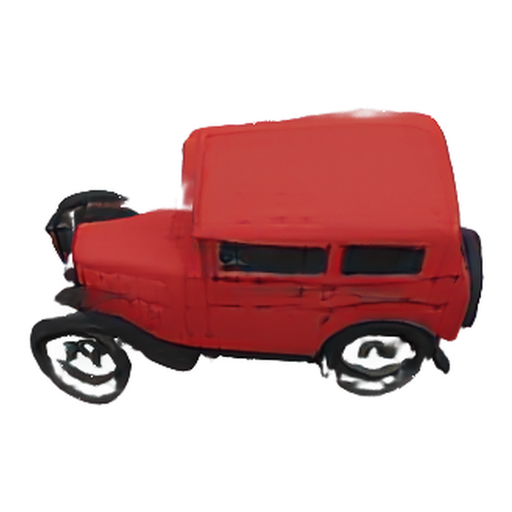}
& \includegraphics[width=\imgsize,height=\imgsize]{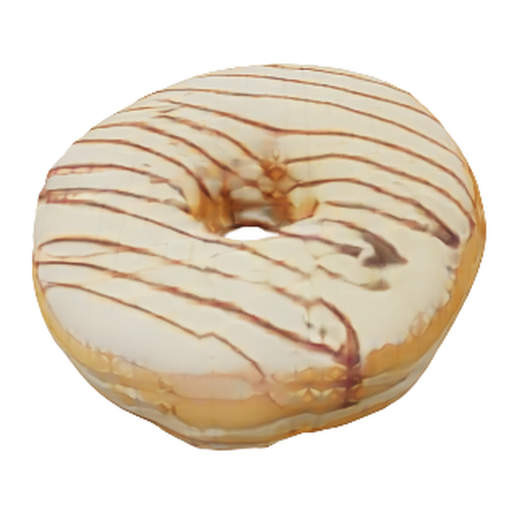}
\\
\rotatebox{90}{\small LaRa}
& \includegraphics[width=\imgsize,height=\imgsize]{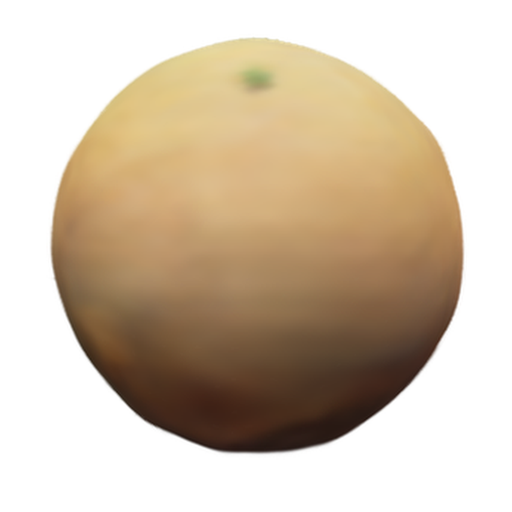}
& \includegraphics[width=\imgsize,height=\imgsize]{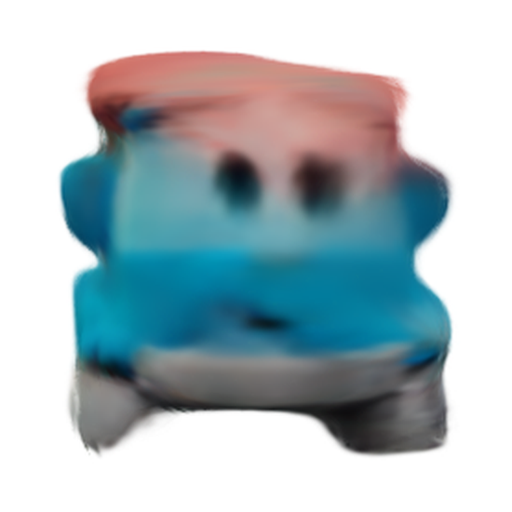}
& \includegraphics[width=\imgsize,height=\imgsize]{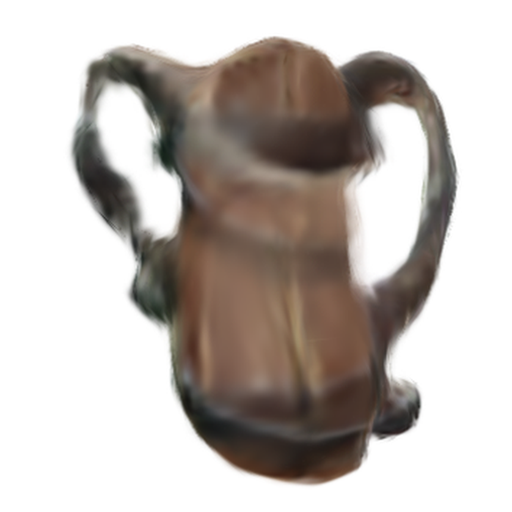}
& \includegraphics[width=\imgsize,height=\imgsize]{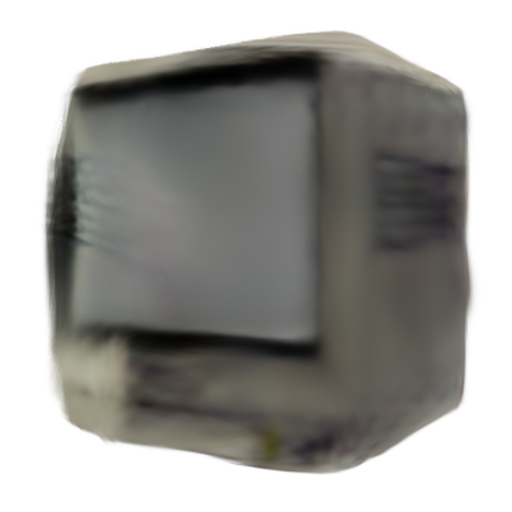}
& \includegraphics[width=\imgsize,height=\imgsize]{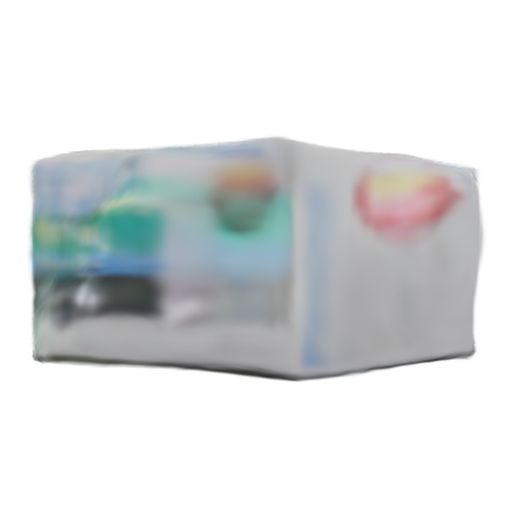}
& \includegraphics[width=\imgsize,height=\imgsize]{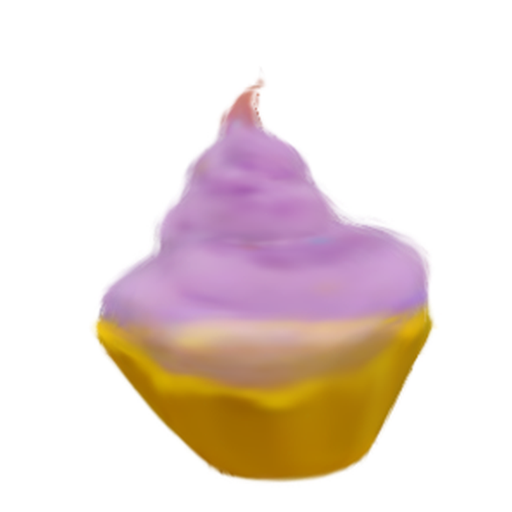}
& \includegraphics[width=\imgsize,height=\imgsize]{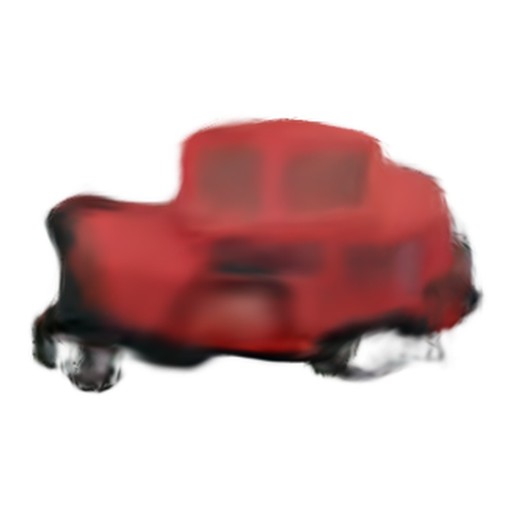}
& \includegraphics[width=\imgsize,height=\imgsize]{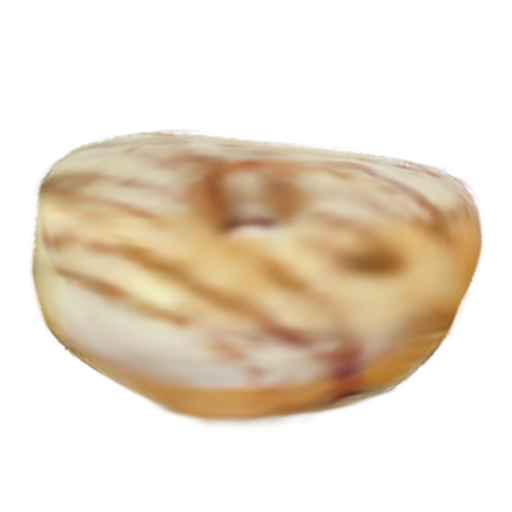}
\\
\rotatebox{90}{\small OnePoseGen}
& \includegraphics[width=\imgsize,height=\imgsize]{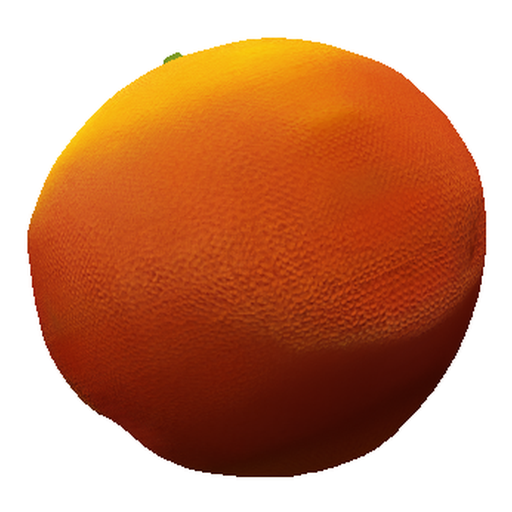}
& \includegraphics[width=\imgsize,height=\imgsize]{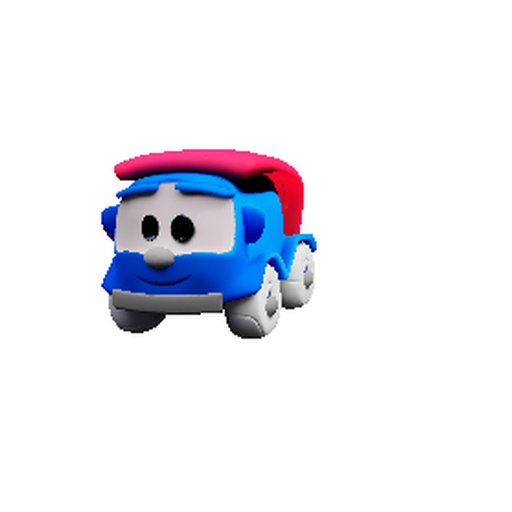}
& \includegraphics[width=\imgsize,height=\imgsize]{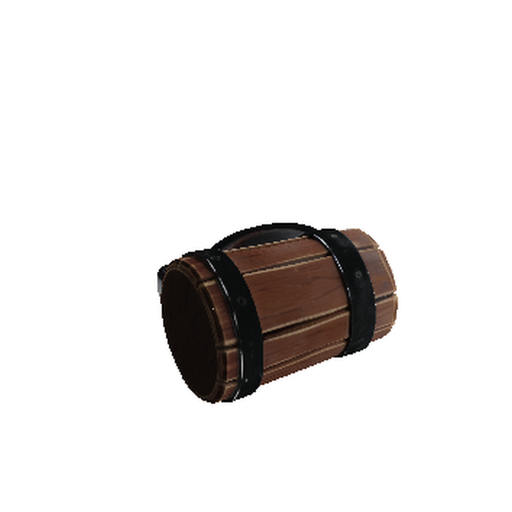}
& \includegraphics[width=\imgsize,height=\imgsize]{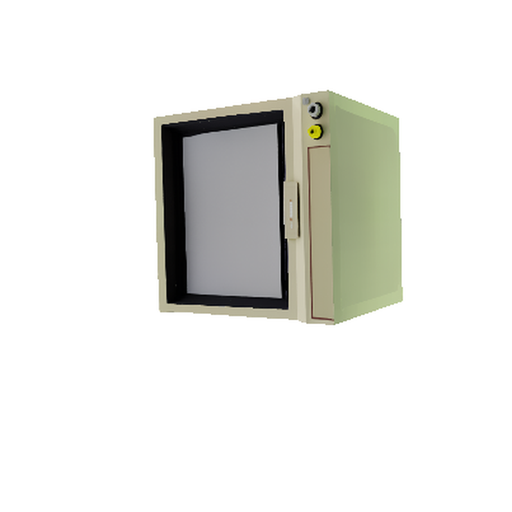}
& \includegraphics[width=\imgsize,height=\imgsize]{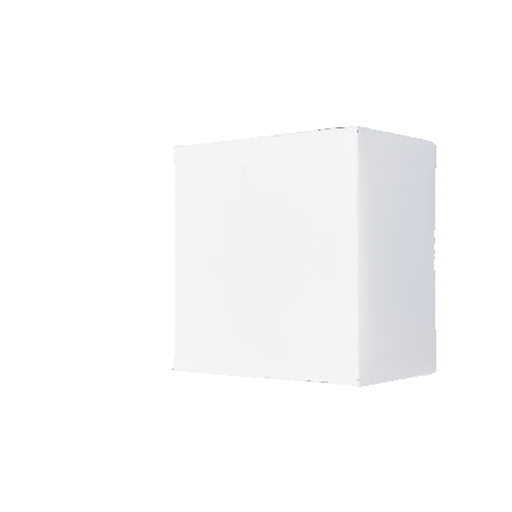}
& \includegraphics[width=\imgsize,height=\imgsize]{figures/exp_sv_app/Toys4k_3891/oneposeviagen_crop.png}
& \includegraphics[width=\imgsize,height=\imgsize]{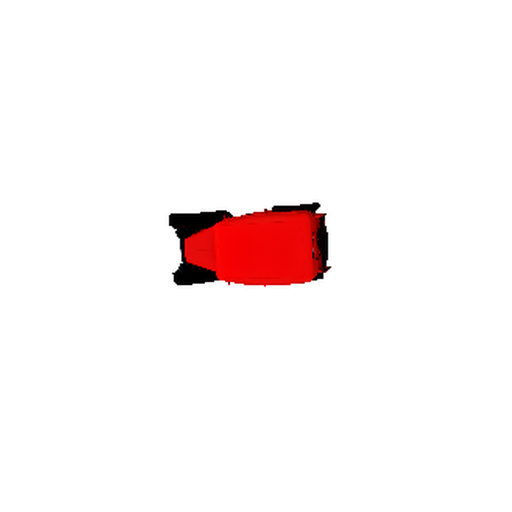}
& \includegraphics[width=\imgsize,height=\imgsize]{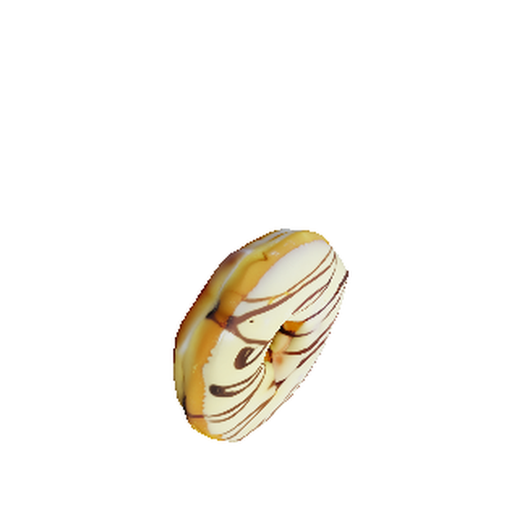}
\\
\end{tabular}

\caption{
\textbf{
Qualitative comparison on input view consistency.
}
We render the input view using its generated camera pose.
For view centric methods (LRM, LaRa), we use ground-truth intrinsic for rendering as they do not model intrinsic. 
Our method produces the highest-fidelity geometry and appearance; LRM hallucinates incorrect details, LaRa is overly blurry due to 2D diffusion inconsistencies, and 3D generation method OnePoseGen frequently fails to register pose reliably.
}
\label{fig:view_consistency}
\end{figure*}

\section{Component-aligned Scene Reconstruction}

By explicitly modeling the spatial relationship between the generated object and the input camera pose, our approach seamlessly integrates powerful generative priors into 3D reconstruction. \Cupid naturally extends to scene-level compositional reconstruction by using explicit 3D-to-2D correspondences for spatial alignment. We first harness foundation models for object segmentation~\citep{ravi2024sam}, generate each 3D object, and compose them into a scene with a global depth prior~\citep{wang2025moge2}. \Figref{fig:placeholder} shows one such result with our method. More results are given in \Figref{fig:scene_supp} of \Appref{ap:scene-diversity}.

\paragraph{Occlusion-aware 3D generation.} Images with multiple objects often contain mutual occlusions, which our synthetic training data do not model. To address this, we introduce random masks on the conditioning image during fine-tuning, inspired by Amodal3R~\citep{wu2025amodal3r}. We adapt their masking strategy to our architecture; see \Appref{sec:occlusion} for implementation. Then, given an input image with multiple objects and their corresponding object masks, we reconstruct each object independently with our occlusion-aware generator, yielding dense 3D--2D correspondences per object.

\paragraph{Multi-component composition.} Since absolute depth is not preserved across objects, direct 3D-2D alignment to the image is ill-posed. We address this using MoGe~\citep{wang2025moge} to predict a global pointmap, thereby reducing 3D--2D alignment to 3D--3D alignment. For the $k$-th object, we collect the matched pair ${(\mathbf{x}_i^{k},\, \mathbf{u}_i)}$, where $\mathbf{x}_i^{k}$ is the coordinates of a visible 3D point on the generated 3D object and $\mathbf{u}_i$ is the corresponding pixel coordinates within the mask $\mathbf{M}^{k}$. We query the MoGe pointmap to obtain $\mathbf{p}_i = P(\mathbf{u}_i)$ in the camera frame, then estimate a per-object similarity transformation $\mathcal{S}_{k} = (s_{k}, \mathbf{R}_{k}, \mathbf{t}_{k})$ via the Umeyama method~\citep{umeyama2002least} on pairs $(\mathbf{x}_i^{k},\, \mathbf{p}_i)$. Applying $\mathcal{S}_{k}$ places each object in the shared camera frame, yielding a component-aligned scene reconstruction. This composition can also extend to multi-view input using VGGT~\citep{wang2025vggt}, allowing for more flexible input and transforming partial geometry into holistic geometry.

\vspace{-3mm}
\section{Experiments}

We evaluate key baselines on three tasks: monocular geometry prediction, input-view consistency, and single-image-to-3D reconstruction in \Secref{sec:Evaluation}. We exclude per-scene optimization methods and focus on learning-based systems. We ablate pose-conditioning in \Secref{sec:ablation}, with implementation details and pose encoding accuracy in \Appref{sec:implementation}. We do not extensively compare with state-of-the-art 3D generators, as improving their aesthetic or geometric quality is orthogonal to our goal. Qualitative comparisons with our baseline~\citep{xiang2025structured} for input-view consistency appear in \Secref{sec:comparison_3dgen}.

\vspace{-1mm}
\subsection{Evaluation}
\label{sec:Evaluation}
\paragraph{Baselines.}
We compare against three complementary families for single image 3D reconstruction: point-map regression, view-centric 3D reconstruction, and 3D generation with post-hoc pose alignment. 

\renewcommand{\arraystretch}{1.0}
\begin{table}[t]
\centering
\caption{
\textbf{Input-view consistency.} \Cupid achieves superior input view consistency, producing accurate appearance alignment.
}
\label{tab:appearance}
\centering
\resizebox{\linewidth}{!}{
\begin{tabular}{lc|ccc|ccc}
\toprule
\textbf{Dataset} & &
\multicolumn{3}{c|}{\textbf{Toys4K}} &
\multicolumn{3}{c}{\textbf{GSO}} \\
\textbf{Method} & \textbf{Pose} & \textbf{PSNR}$\uparrow$ & \textbf{SSIM}$\uparrow$ & \textbf{LPIPS}$\downarrow$ & \textbf{PSNR}$\uparrow$ & \textbf{SSIM}$\uparrow$ & \textbf{LPIPS}$\downarrow$ \\
\midrule
LaRa & $\xmark$ & 22.00 & 93.42 & 0.0884 & 19.81 & 91.61 & 0.1119 \\
OpenLRM & $\xmark$ & 26.41 & 80.17 & 0.1156 & 25.79 & 78.80 & 0.1268 \\
OnePoseGen & $\cmark$ & 17.43 & 89.37 & 0.1174 & 14.87 & 86.46 & 0.1386 \\
Ours & $\cmark$ & \textbf{30.05} & \textbf{96.81} & \textbf{0.0251} & \textbf{28.68} & \textbf{95.49} & \textbf{0.0354} \\
\bottomrule
\end{tabular}
}
\end{table}
\begin{table}[t]
\vspace{6pt}
\centering
\caption{\textbf{Comparison on full 3D quality. }We report CLIP image scores of novel views following \citep{gao2024cat3d}.}
\label{tab:clip}

\renewcommand{\arraystretch}{1.0}
\resizebox{1.0\linewidth}{!}{
\begin{tabular}{l|ccccc}
\toprule
 & OnePoseGen & LaRa & OpenLRM & TRELLIS & Ours \\
\midrule
\textbf{ViT-B/16} & 0.7933 & 0.8334 & 0.8939 & 0.9465 & \textbf{0.9501} \\
\textbf{ViT-L/14} & 0.7193 & 0.7682 & 0.8410 & 0.9210 & \textbf{0.9291} \\
\bottomrule
\end{tabular}
}
\end{table}

\textit{Point-map regression.}
VGGT~\citep{wang2025vggt} and MoGe~\citep{wang2025moge} simultaneously predict per-pixel 3D points and view-centric camera pose for 2D-to-3D reconstruction. These methods reconstruct only the geometry visible in the input view, often lacking robust priors for occluded regions.
\textit{View-centric 3D reconstruction.}
LRM~\citep{hong2023lrm} and LaRa~\citep{LaRa} generate view-centric 3D models, eliminating the need to estimate camera parameters at test time. LRM directly regresses 3D objects in view space, while LaRa enhances the input with novel views from Zero123++~\citep{shi2023zero123++} before reconstruction. We adopt an open-source implementation of LRM~\citep{openlrm} for comparisons.
\textit{3D generation with post-hoc alignment.}
OnePoseGen~\citep{geng2025one} integrates Higen3D~\citep{ye2025hi3dgen} and TRELLIS~\citep{xiang2025structured}, which produces canonical 3D models, with FoundationPose~\citep{wen2024foundationpose} to align the 3D model with the input view. This approach evaluates whether decoupling 3D generation from pose estimation and applying post-hoc alignment can achieve reprojection consistency. 

\vspace{-4mm}
\paragraph{Evaluation.}
We evaluate three aspects: (i) monocular geometry accuracy, (ii) input-view consistency to assess reprojection alignment and appearance fidelity, and (iii) novel-view consistency to assess full 3D fidelity. 
Experiments are conducted on Toys4K~\citep{stojanov2021using} and GSO~\citep{downs2022google}, containing approximately 3K synthetic and 1K real-world objects, respectively. For novel views, we render four uniform views rotating around the vertical axis. Tasks (i) and (ii) are particularly challenging for generative approaches, as they require producing accurate an object-centric camera pose that aligns with the input view.

\vspace{-2mm}
\paragraph{Monocular geometry accuracy.} 
Monocular geometry metrics evaluate pose alignment and visible geometry fidelity. Unlike multi-view approaches that report pose accuracy relative to the first frame, single-view generation cannot report relative pose metrics. Following MoGe~\citep{wang2025moge}, we use monocular geometry alignment as a proxy for pose accuracy. We construct ground-truth (GT) point clouds by unprojecting depth maps with camera parameters and report Mask IoU, Chamfer Distance (CD), and F-score, providing both mean and median statistics to mitigate outlier sensitivity. Since VGGT outputs point maps without masks, we use GT masks to avoid background contamination. For full 3D methods, we render and unproject the depth map to obtain the visible point cloud, using rendered alphas as the mask. Predicted point clouds are aligned to GT using scale–shift alignment from MoGe~\citep{wang2025moge}. Results are in \Tabref{tab:geometry}.

\begin{table}[t]
\centering
\caption{\textbf{Ablation studies of pose-aligned conditioning.}}
\label{tab:abl_appearance}
\renewcommand{\arraystretch}{1.065}
\resizebox{1.0\linewidth}{!}{
\begin{tabular}{l|ccc|ccc}
\toprule
\multirow{2}{*}{\textbf{Method}} & \multicolumn{3}{c|}{\textbf{GT Geo \& Pose}} & \multicolumn{3}{c}{\textbf{Sampled Geo \& Pose}} \\
 & \textbf{PSNR} & \textbf{SSIM} & \textbf{LPIPS} & \textbf{PSNR} & \textbf{SSIM} & \textbf{LPIPS} \\
\midrule
(a) Baseline (w/o PAC)   & 31.84 & 97.50 & 0.0219 & 27.47 & 95.64 & 0.0327 \\
(b) Position Embedding   & 32.07 & 97.58 & 0.0211 & 27.56 & 95.67 & 0.0323 \\
(c) Latent (w/o Occ.)    & 32.37 & 97.72 & 0.0201 & 27.85 & 95.87 & 0.0309 \\
(d) Latent (Occ.)        & 32.39 & 97.77 & 0.0199 & 27.74 & 95.80 & 0.0313 \\
(e) Latent (Visual Feat.)& \textbf{34.86} & \textbf{98.24} & \textbf{0.0168} & \textbf{30.05} & \textbf{96.81} & \textbf{0.0251} \\
\bottomrule
\end{tabular}
}
\end{table}
\setlength{\imgsize}{1.29cm} 

\newcommand{\cornerlabel}[1]{%
  \put(66,92){\makebox[0pt][r]{\scriptsize\textcolor{black}{#1}}}%
}

\newcommand{\AutoHeightCell}[1]{%
  \begin{minipage}[t]{\linewidth}%
    \centering
    #1%
  \end{minipage}%
}

\begin{figure}[!t]
\centering
\captionsetup[subfigure]{justification=centering}

\setlength{\arrayrulewidth}{0.6pt}

\begin{tabular}{@{} p{0.493\linewidth} @{\hspace{1.5pt}} | @{\hspace{1.5pt}} p{0.493\linewidth} @{}}

\AutoHeightCell{%
\setlength{\tabcolsep}{1pt}%
\renewcommand{\arraystretch}{1.0}%
\begin{tabular}{*{3}{>{\centering\arraybackslash}m{\imgsize}}}
    \begin{overpic}[width=\imgsize]{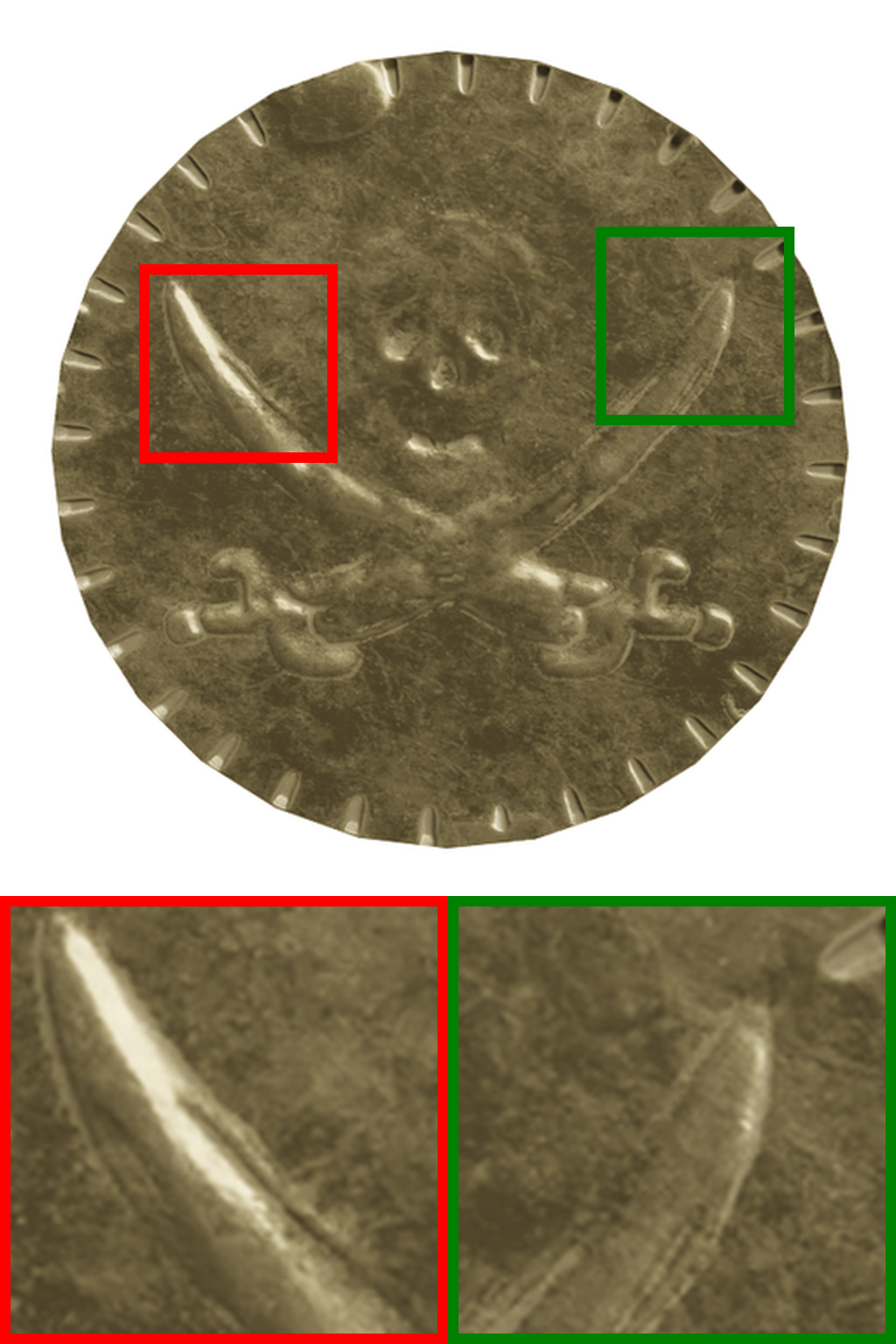}\cornerlabel{GT}\end{overpic} &
    \begin{overpic}[width=\imgsize]{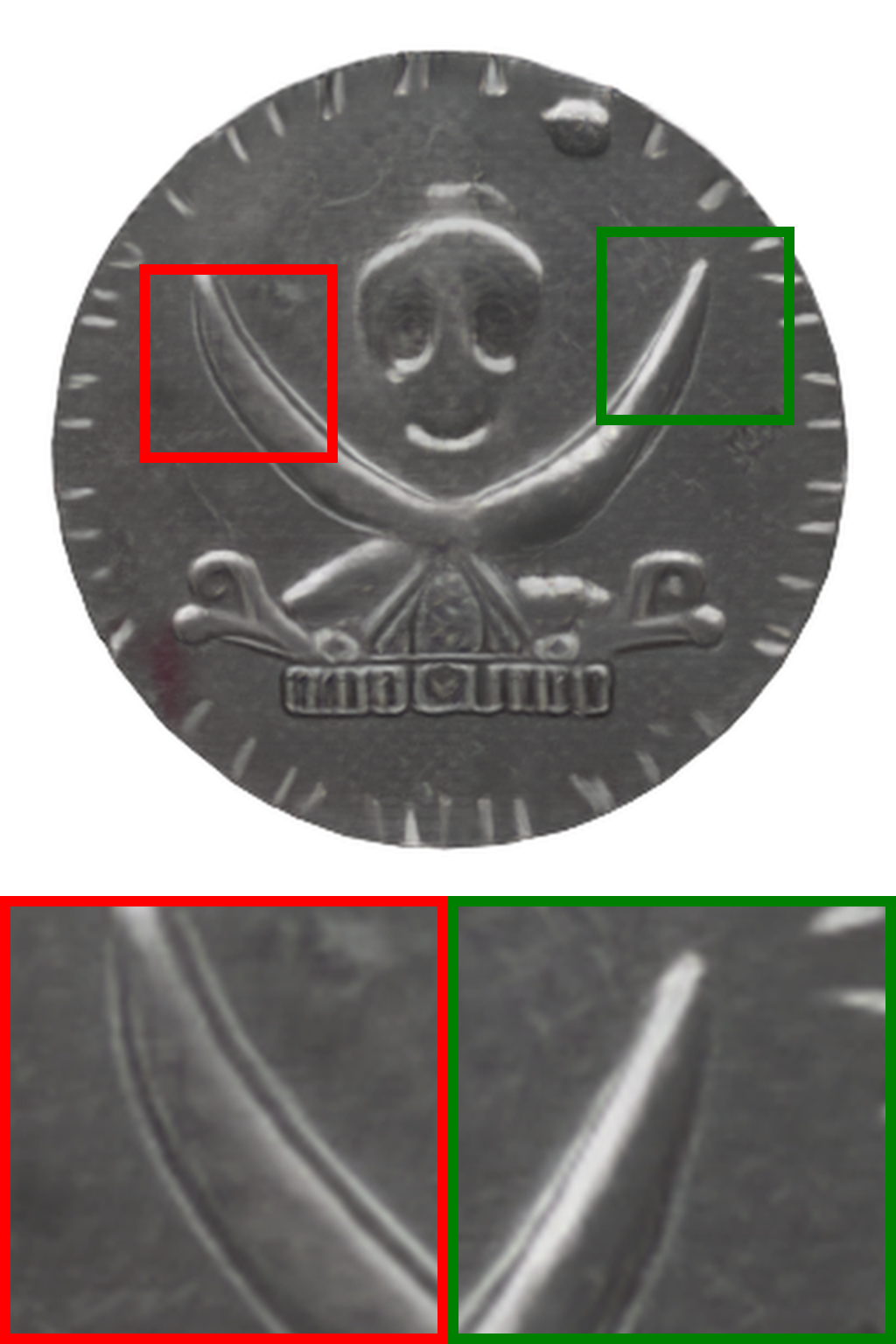}\cornerlabel{(a)}\end{overpic} &
    \begin{overpic}[width=\imgsize]{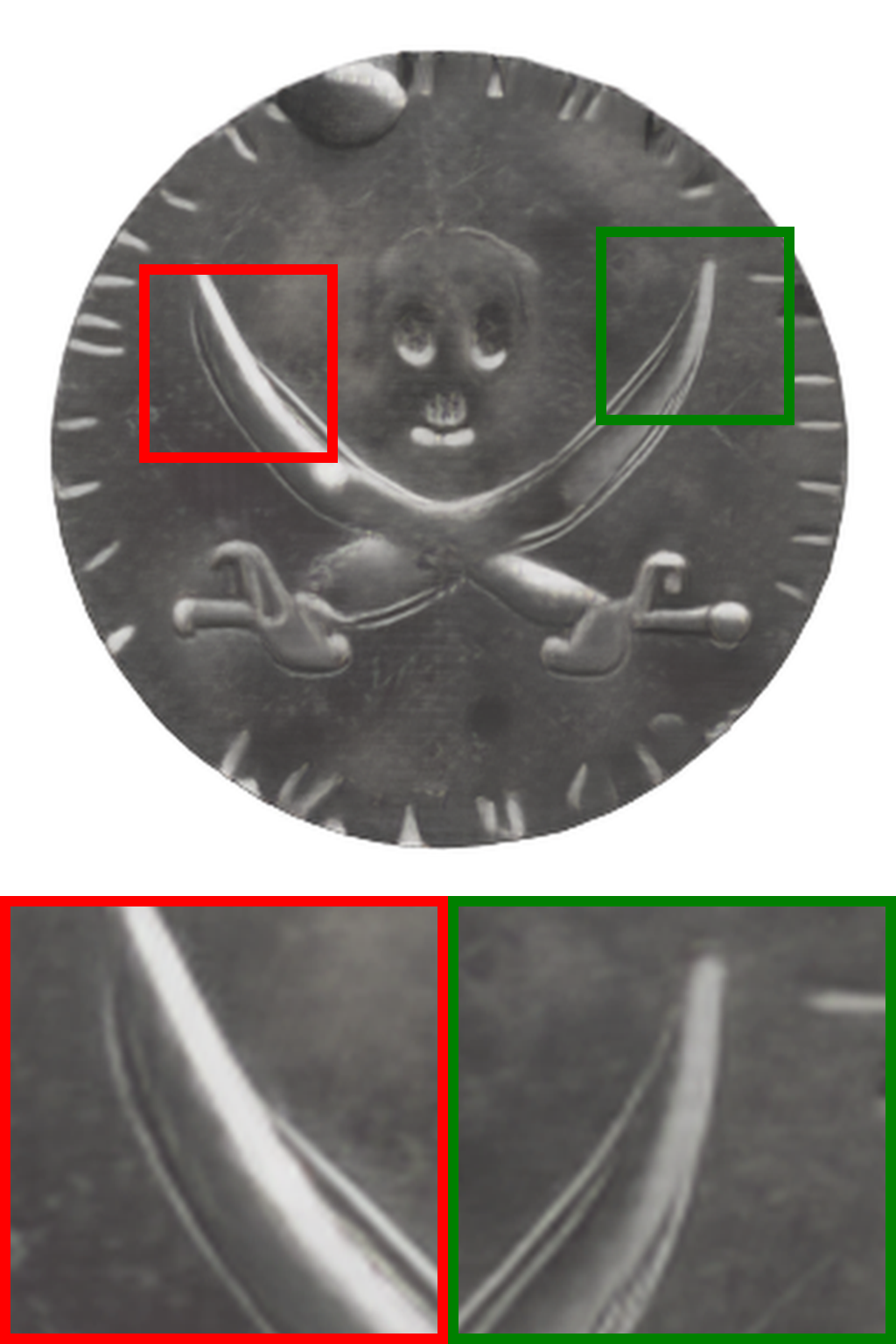}\cornerlabel{(b)}\end{overpic} \\
    \begin{overpic}[width=\imgsize]{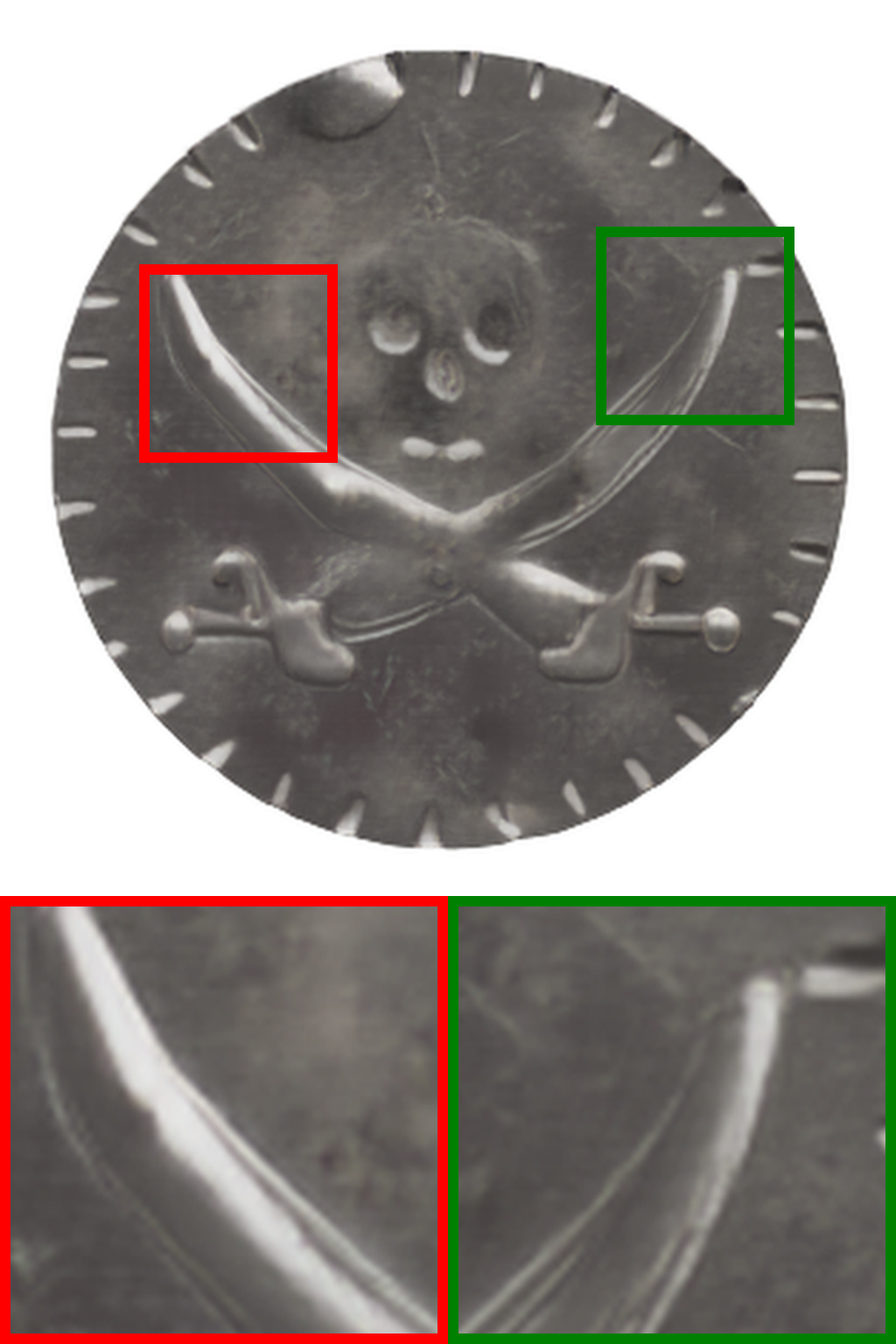}\cornerlabel{(c)}\end{overpic} &
    \begin{overpic}[width=\imgsize]{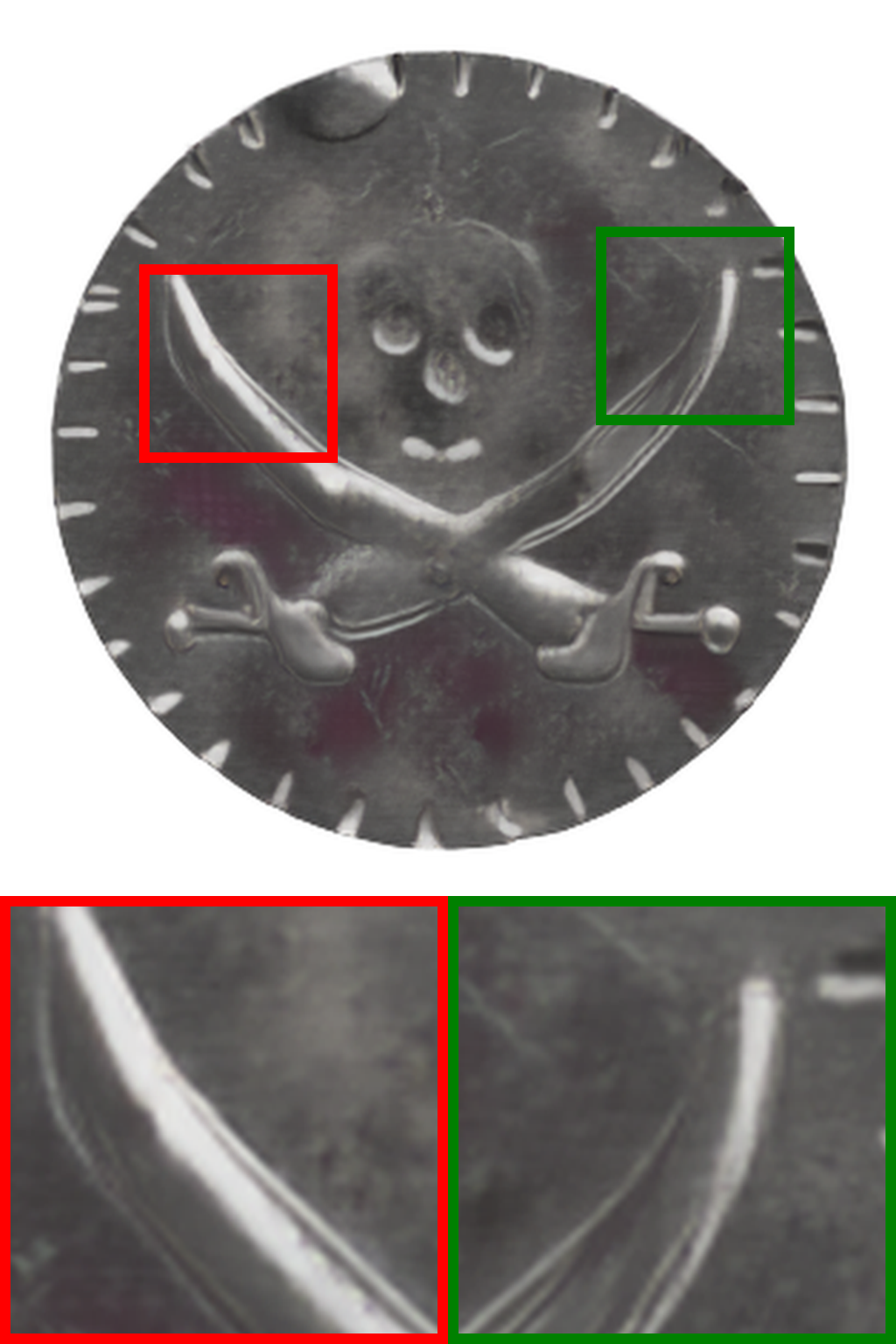}\cornerlabel{(d)}\end{overpic} &
    \begin{overpic}[width=\imgsize]{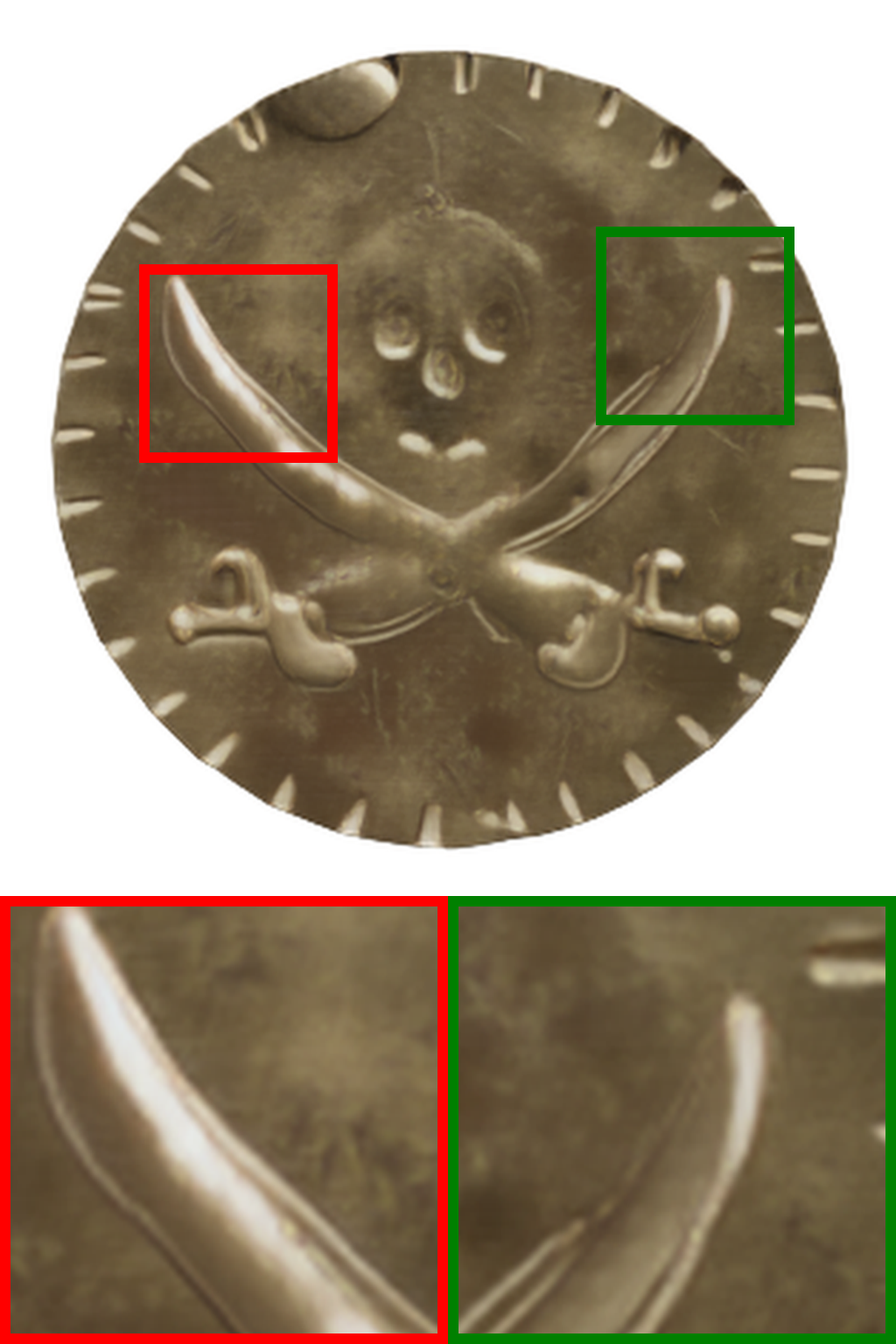}\cornerlabel{(e)}\end{overpic} \\
\end{tabular}
}
&
\AutoHeightCell{%
\setlength{\tabcolsep}{1pt}%
\renewcommand{\arraystretch}{1.0}%
\begin{tabular}{*{3}{>{\centering\arraybackslash}m{\imgsize}}}
    \begin{overpic}[width=\imgsize]{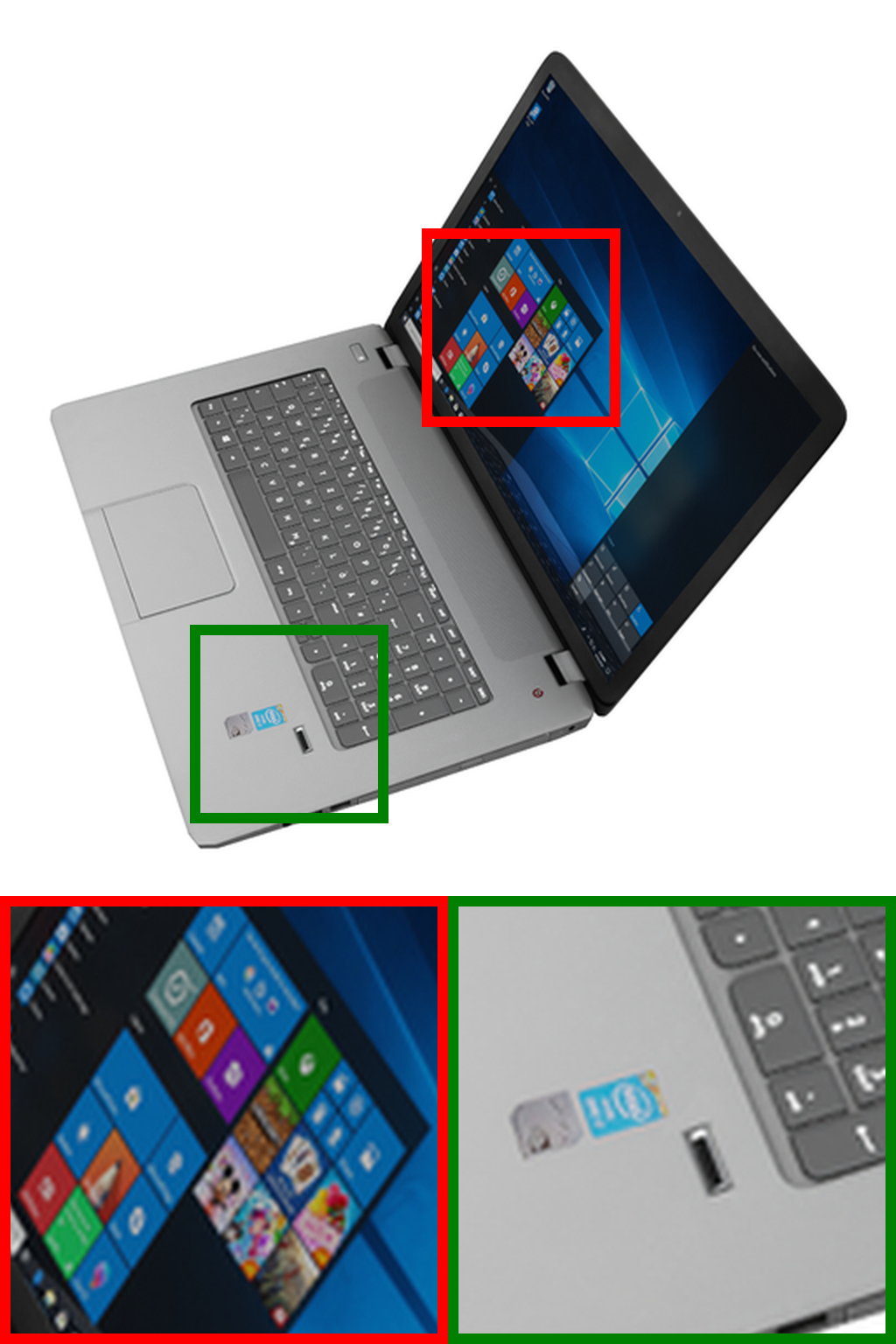}\cornerlabel{GT}\end{overpic} &
    \begin{overpic}[width=\imgsize]{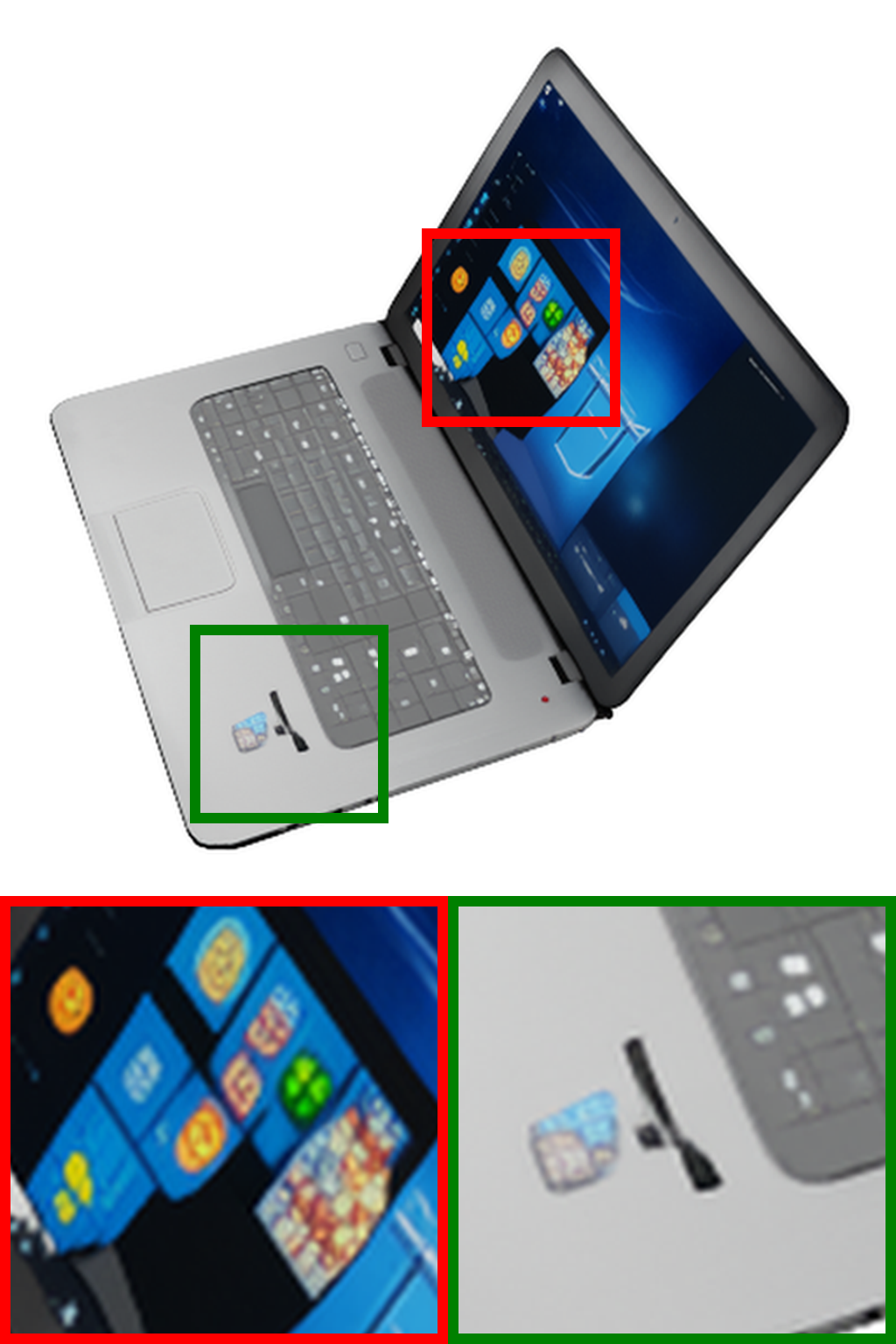}\cornerlabel{(a)}\end{overpic} &
    \begin{overpic}[width=\imgsize]{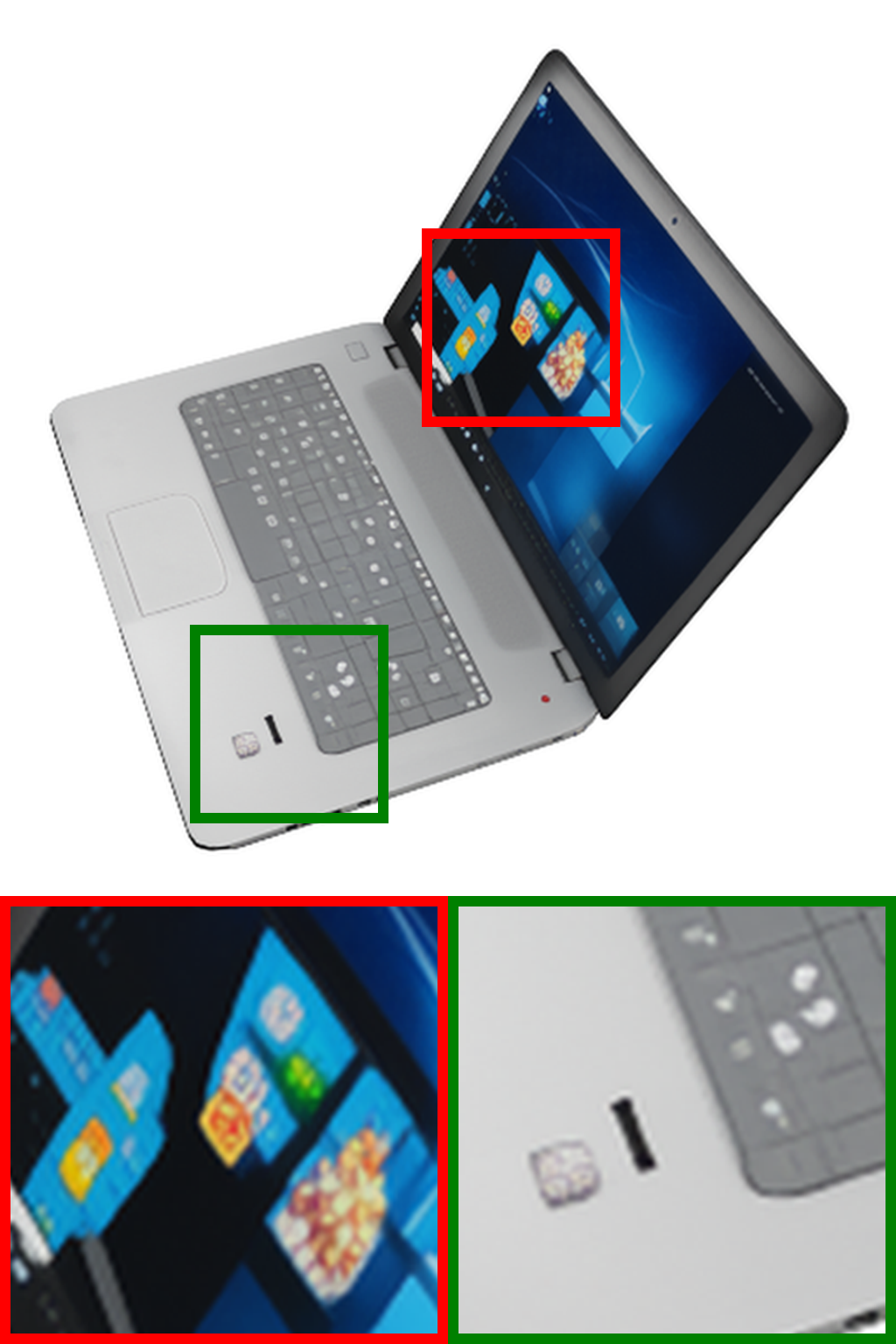}\cornerlabel{(b)}\end{overpic} \\
    \begin{overpic}[width=\imgsize]{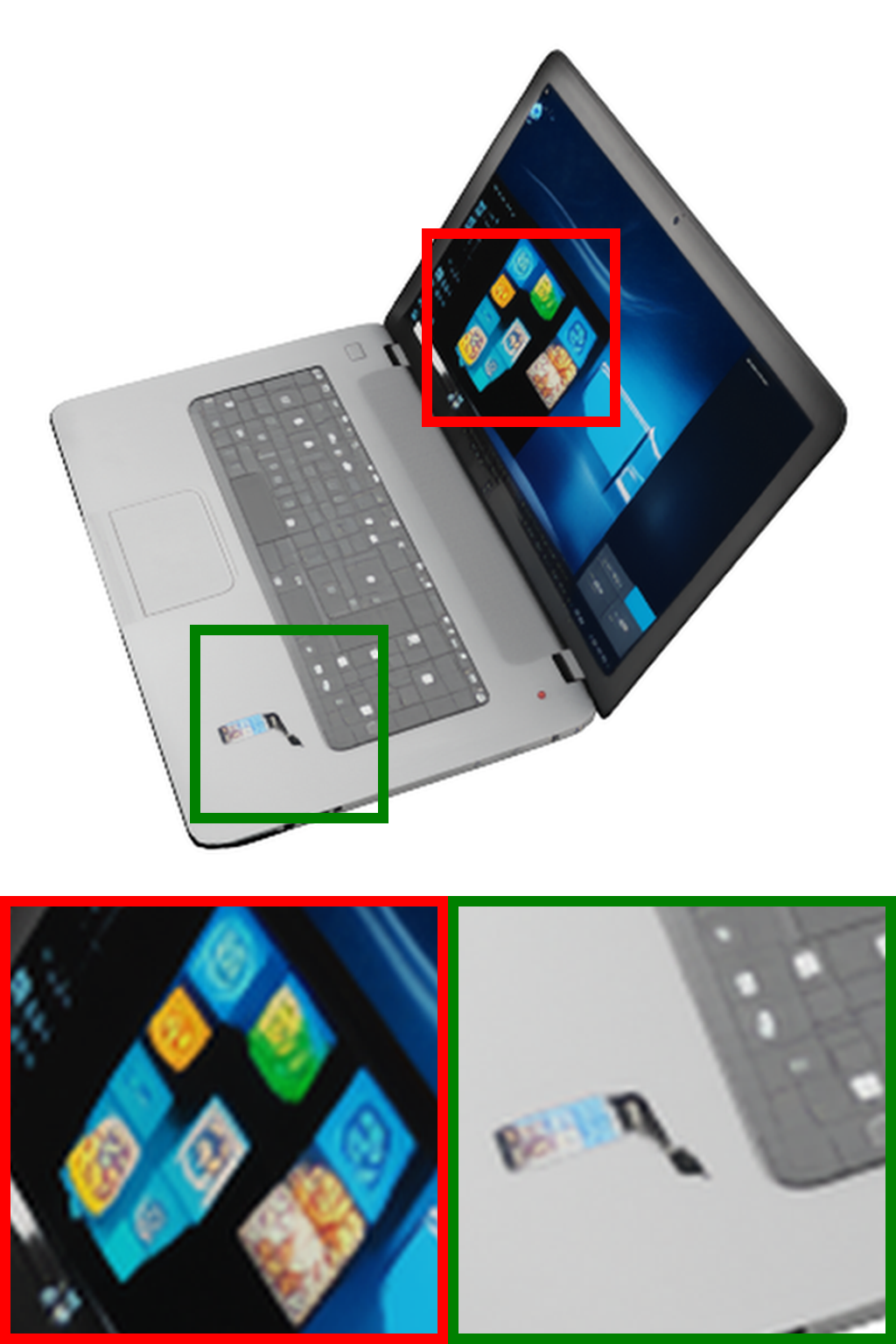}\cornerlabel{(c)}\end{overpic} &
    \begin{overpic}[width=\imgsize]{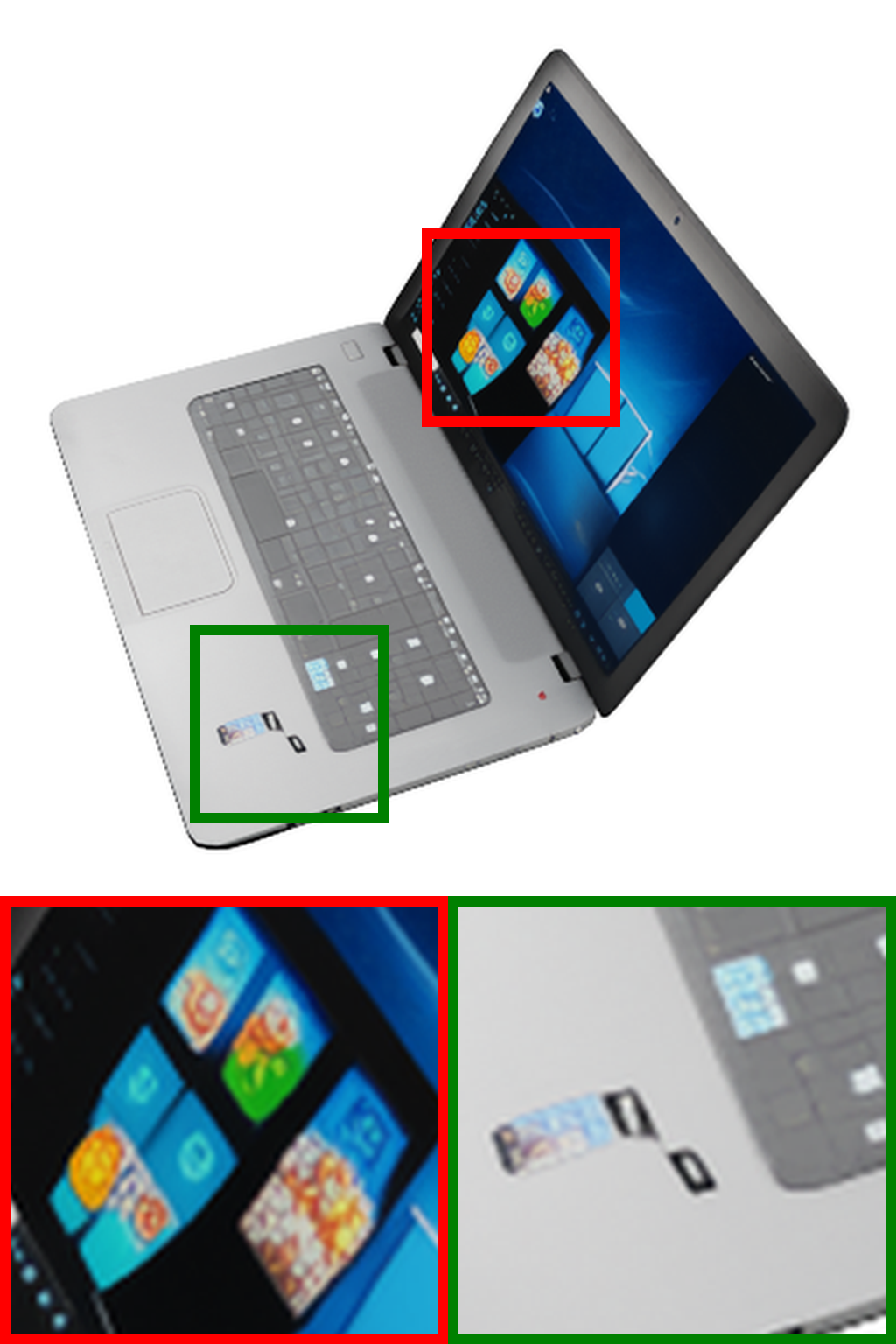}\cornerlabel{(d)}\end{overpic} &
    \begin{overpic}[width=\imgsize]{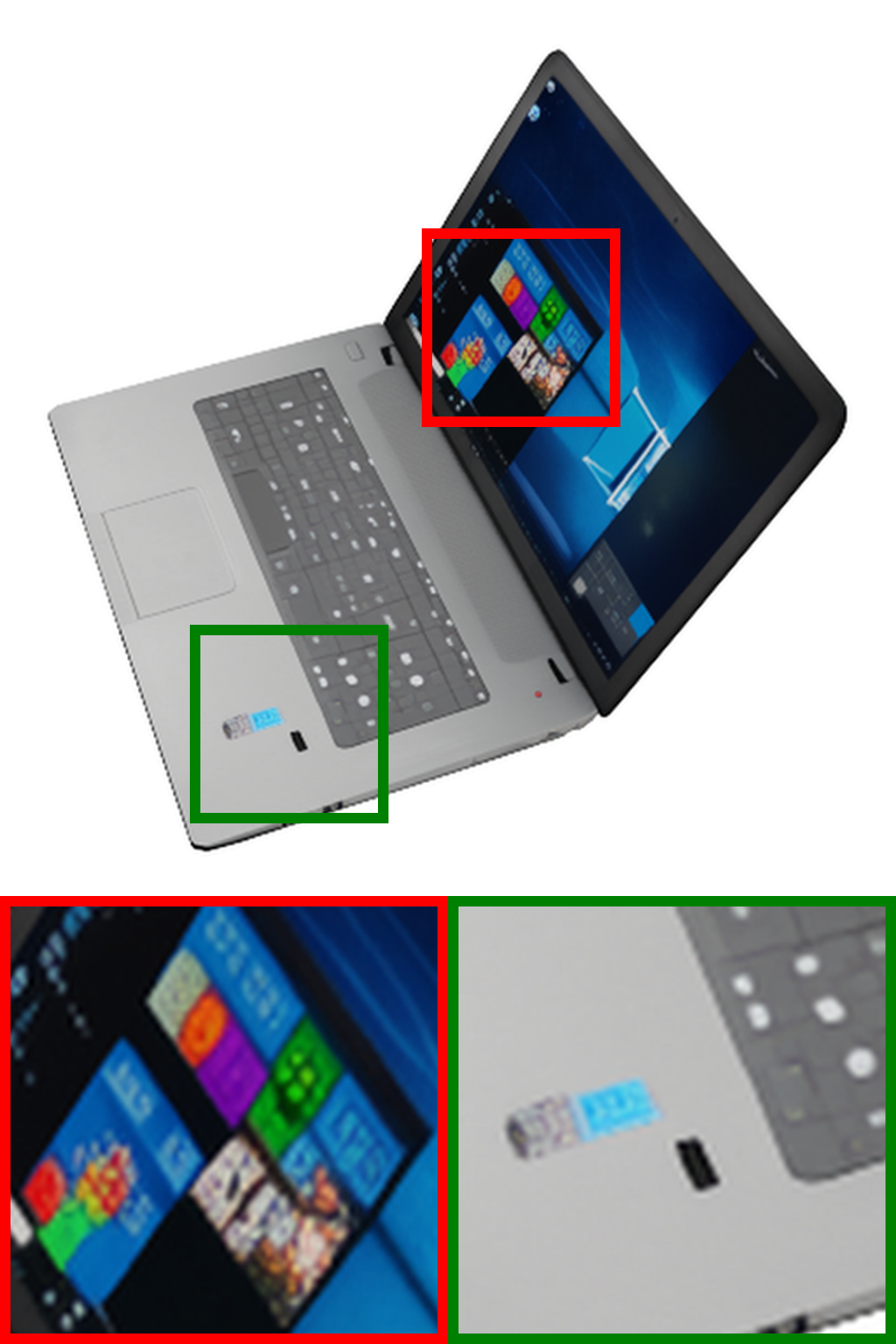}\cornerlabel{(e)}\end{overpic} \\
\end{tabular}
}
\end{tabular}

\caption{
\textbf{Qualitative comparison of various pose-aligned conditioning.} Our method (e) achieves the best visual quality in terms of color fidelity and detail.
}\label{fig:abl_appearance}
\end{figure}
\newcommand{\mvteaserwidthb}{.333\textwidth}
\begin{figure*}[t]
\captionsetup[subfigure]{labelformat=empty}
\centering
\makebox[\textwidth][c]{%
\renewcommand{\arraystretch}{0.4}%
\begin{tabular}{@{}c@{\betweencols}c@{\betweencols}c@{}} 
\begin{subfigure}[t]{\mvteaserwidthb}
  \includegraphics[width=\columnwidth]{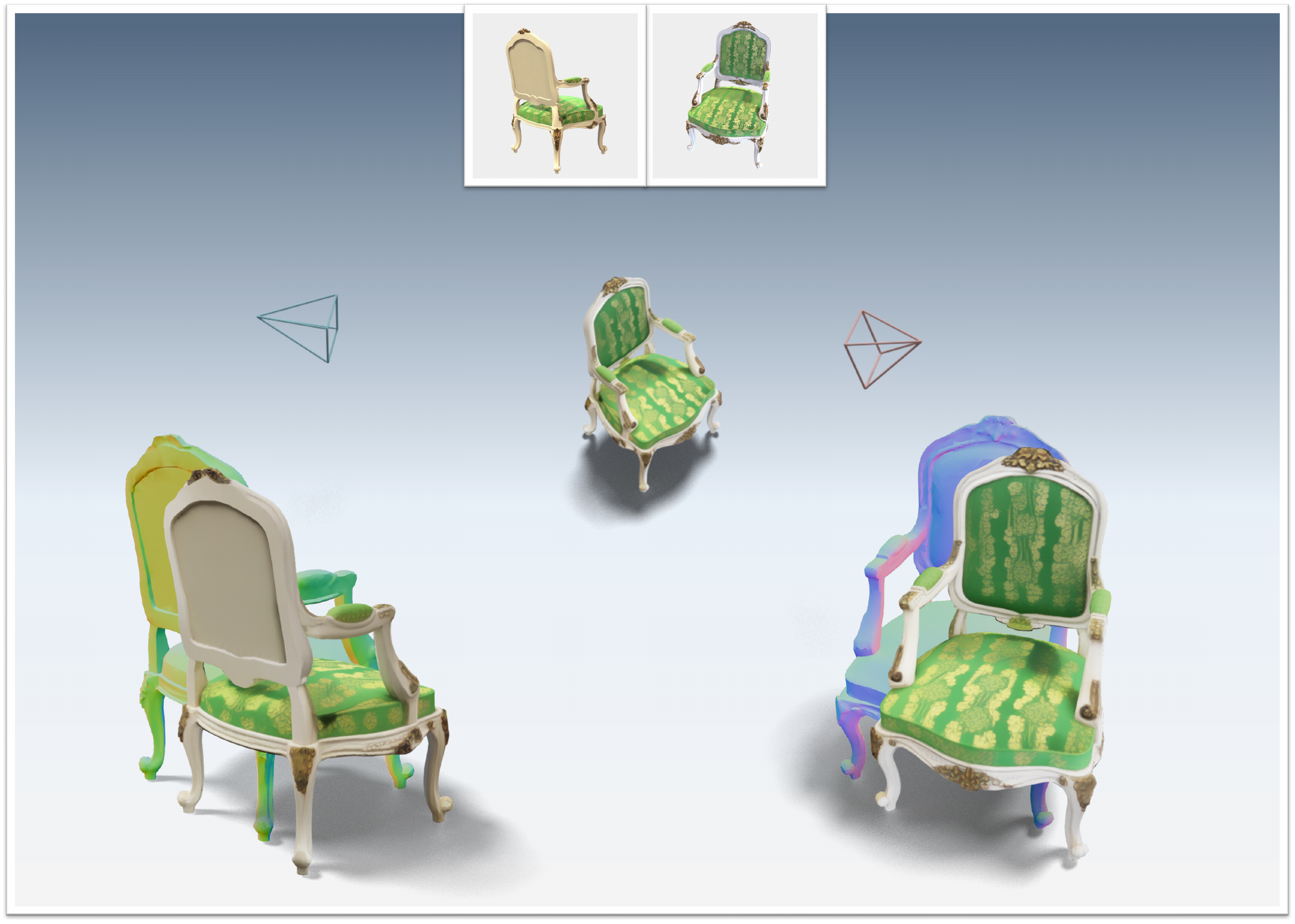}
\end{subfigure} &
\begin{subfigure}[t]{\mvteaserwidthb}
  \includegraphics[width=\columnwidth]{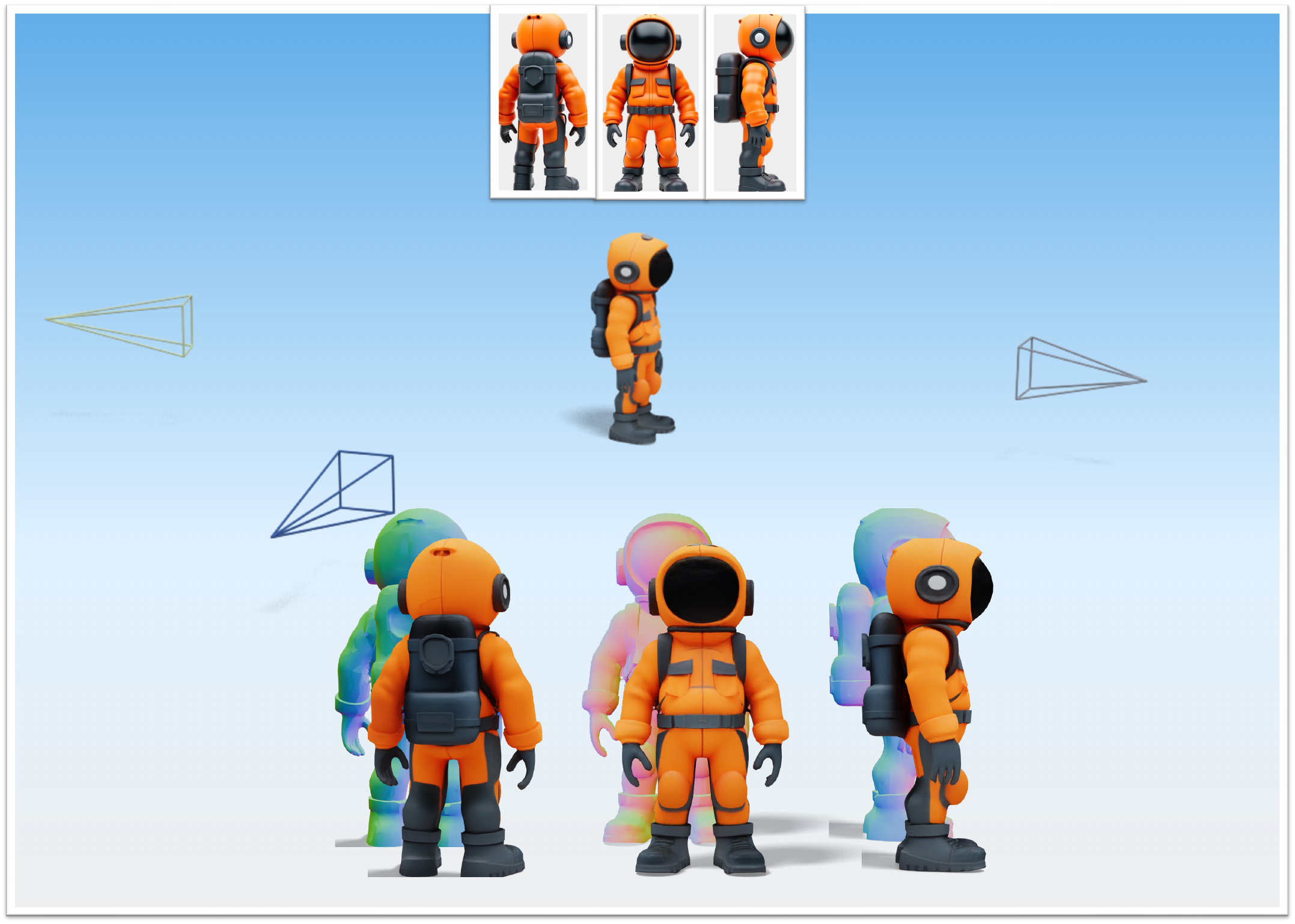}
\end{subfigure} &
\begin{subfigure}[t]{\mvteaserwidthb}
  \includegraphics[width=\columnwidth]{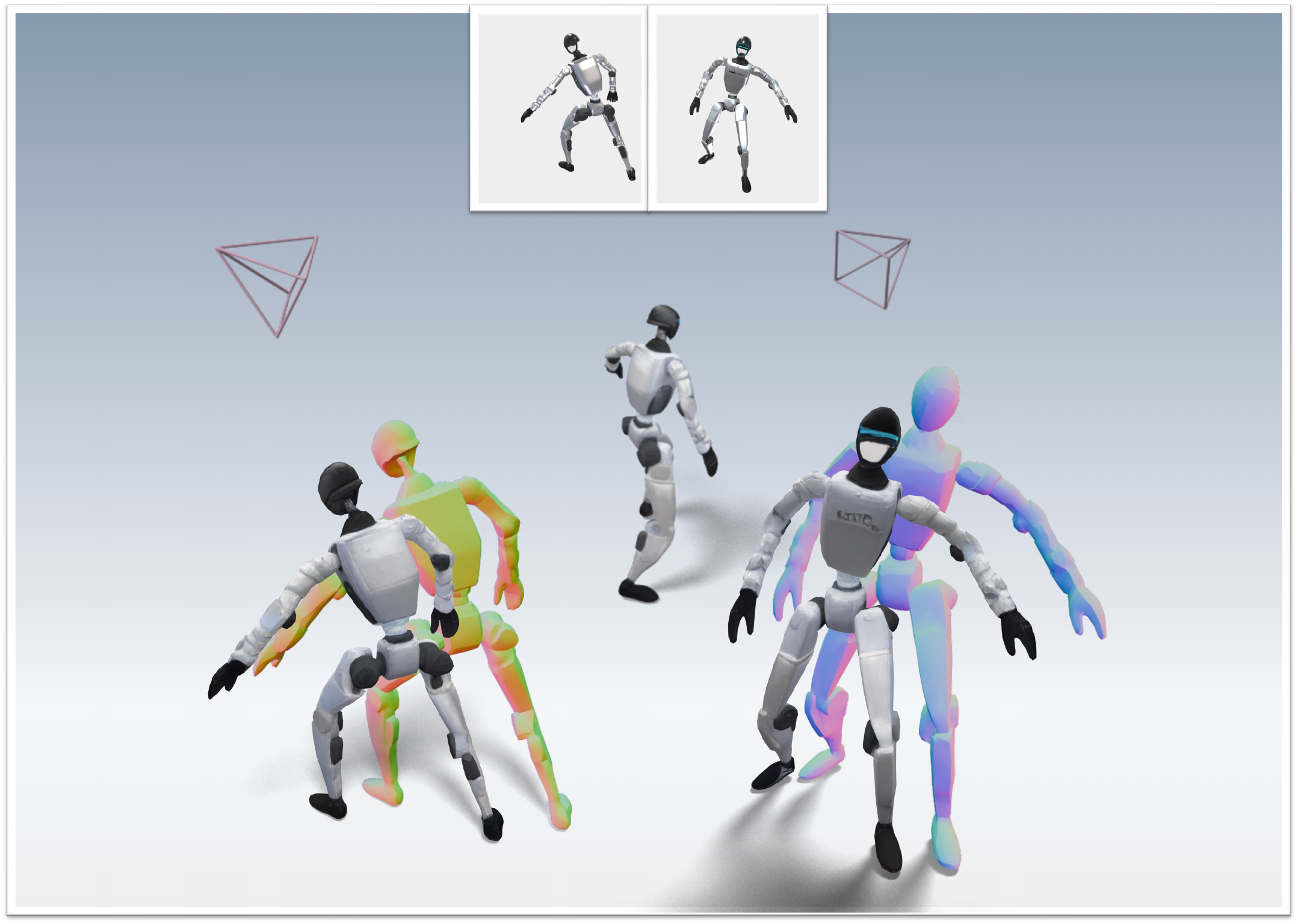}
\end{subfigure} \\ 
\end{tabular}%
}
\caption{
\textbf{Multi-view conditioning.} Our decoupled joint modeling naturally supports multi-view conditioning. When multiple input views are available, we fuse the shared \textit{view-agnostic} object latent across flow paths (similar to MultiDiffusion~\citep{bar2023multidiffusion}), enabling object and cameras refinement across all views. Top: inputs; Middle: reconstructed 3D object and camera poses; Bottom: rendered images and geometry.}
\label{fig:mv_images}
\end{figure*}

Our approach demonstrates clear superiority over view-centric reconstruction methods such as LRM and LaRa. On the GSO dataset, it reduces the average Chamfer Distance (CD) by 50\% relative to LRM. This improvement stems not only from accurate intrinsic modeling via generative priors but also from our precise extrinsic camera parameter prediction. By contrast, although OnePoseGen excels in intrinsic modeling, it falls short on Toys4K and GSO, as these datasets contain geometric symmetries and textureless regions that cause its registration modules to fail during extrinsic pose estimation.
Crucially, our model is also highly competitive with methods that produce only partial geometries (\emph{e.g.}, VGGT, MoGe). This suggests that complex, multi-stage pipelines which use these partial methods as a first step~\citep{yao2025castcomponentaligned3dscene} can be streamlined. Our approach achieves comparable monocular accuracy in a single step, eliminating the need for such intermediate representations.

\paragraph{Input view consistency.} 
We evaluate input view consistency to measure alignment between the 3D reconstruction and the input image. We quantify the difference between the re-rendered and input images, reporting PSNR, SSIM, and LPIPS in \Tabref{tab:appearance}, with qualitative comparisons in \Figref{fig:view_consistency}.
Our method achieves substantial fidelity improvements over existing reconstruction-based methods.
We find LaRa's texture quality is affected by inconsistencies in the generated novel views, while LRM struggles to predict detailed textures. OnePoseGen can produce plausible 3D objects but suffers severely from texture misalignment (e.g., color shifting), often failing on pose estimation. Our method achieves strong pose accuracy while retaining high-quality texture details, as evidenced by the zoomed-in views in \Figref{fig:view_consistency}.

\paragraph{Full 3D evaluation.} 
Evaluating single-image-to-3D is challenging because many distinct 3D shapes can explain a single view. We therefore emphasize (1) qualitative assessments (\Figref{fig:view_consistency}) and (2) quantitative semantic alignment using CLIP similarity~\citep{radford2021learning} on Toys4K (\Tabref{tab:clip}). Our method consistently outperforms all baselines. Moreover, it reconstructs high-fidelity 3D objects and scenes from in-the-wild images (\Figref{fig:teaser}), which prior methods do not achieve.

\subsection{Ablation Studies}
\label{sec:ablation}

We compare pose-aligned conditioning (PAC) variants in the second-stage refinement on Toys4K: (a) the TRELLIS baseline (without PAC); (b) adding DINOv2 positional embeddings to SLat latents; (c) concatenating DINOv2 feature volumes with view-conditioned voxel latents from the frozen SLat encoder; (d) as in (c) but zeroing latents of occluded voxels in the input view; and (e) concatenating additional visual features from a convolutional layer applied to the input image. We report results using both GT coarse geometry and camera pose, and samples from the first stage.

As shown in \Tabref{tab:abl_appearance}, when provided with ground-truth coarse geometry and camera pose, methods (b) and (c) validate the benefit of pose-aligned conditioning in generating geometry and appearance; (d) indicates that performance remains consistent whether occlusion is present or absent; our full method (e) provides the best visual quality by supplementing DINOv2 features with missing low-level cues, yielding better color and detail alignment, as shown in \Figref{fig:abl_appearance}. 
Crucially, these advantages are largely preserved when using sampled geometry and camera poses from the first stage (30.05 PSNR for (e) vs. 27.47 for baseline (a)). This confirms that our pose-aligned conditioning is highly robust to small perturbations from stochastic generation.
\section{Discussion}

We present \Cupid, a \textit{generative reconstruction} framework that unifies 3D generation and reconstruction by jointly modeling canonical 3D object and camera pose that formulate a given observation. By incorporating the missing camera pose prior and pose-aligned conditioning into 3D generators, this simple yet effective design enables \Cupid to significantly improve geometric and photometric consistency with the input image for single-view reconstruction.

Despite strong performance and consistent gains over prior methods, \Cupid has limitations. First, it requires object masks like existing 3D generation methods, and boundary errors in real images can degrade reconstruction quality. Second, lighting can be baked into the appearance, which can be improved with better material–light disentanglement. Third, our synthetic training images are mostly centered, making off-centered objects more challenging in real scenes. Nevertheless, these are not fundamental limitations and can be alleviated with better data and supervision.

Looking forward, this generative reconstruction framework enables promising test-time extensions, particularly for \textit{multi-view} scenarios. From multiple images, our method refines 3D reconstructions to align with all observations by fusing a shared object latent during sampling, similar to Multi-Diffusion~\citep{bar2023multidiffusion}. As shown in \Figref{fig:mv_images}, this yields an SfM-like system, though challenges like misaligned 3D orientations~\citep{lu2025orientation} from diverse views require advanced fusion schemes. Moreover, \Cupid's joint modeling enables bidirectional capabilities: flexible generation with known poses or pose estimation from images of given objects, facilitating mixed reality and embodied AI applications.

{
    \small
    \bibliographystyle{ieeenat_fullname}
    \bibliography{main}
}

\clearpage
\appendix
\section{Appendix}
\definecolor{lightgray}{gray}{0.8} 

\newcommand{\imgvarwidth}{.13\textwidth}
\begin{figure*}[ht]
\captionsetup[subfigure]{labelformat=empty}
\centering
\makebox[\textwidth][c]{
\renewcommand{\arraystretch}{1.2}
\newlength{\oldfboxsep}
\newlength{\oldfboxrule}
\setlength{\oldfboxsep}{\fboxsep}
\setlength{\oldfboxrule}{\fboxrule}
\setlength{\fboxsep}{1pt} 
\setlength{\fboxrule}{1pt} 
\begin{tabular}{@{}c@{\betweencols}c@{\betweencols}c@{\betweencols}c@{\betweencols}c@{\hspace{3pt}\vrule width 0.6pt \hspace{2pt}}c@{\betweencols}c@{}}
\begin{subfigure}[t]{\imgvarwidth}
\fcolorbox{lightgray}{white}{\includegraphics[width=0.95\columnwidth]{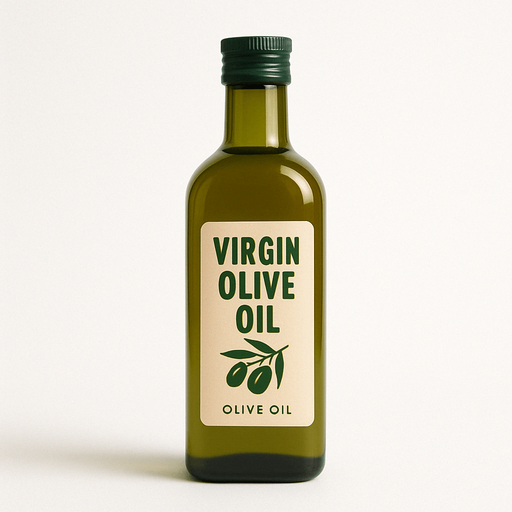}}
\end{subfigure} &
\begin{subfigure}[t]{\imgvarwidth}
\fcolorbox{lightgray}{white}{\includegraphics[width=0.95\columnwidth]{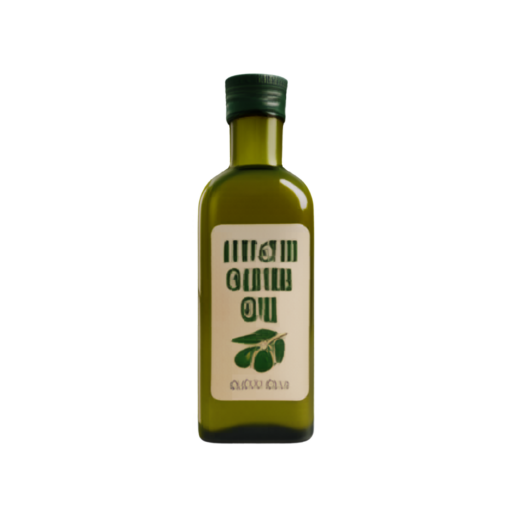}}
\end{subfigure} &
\begin{subfigure}[t]{\imgvarwidth}
\fcolorbox{lightgray}{white}{\includegraphics[width=0.95\columnwidth]{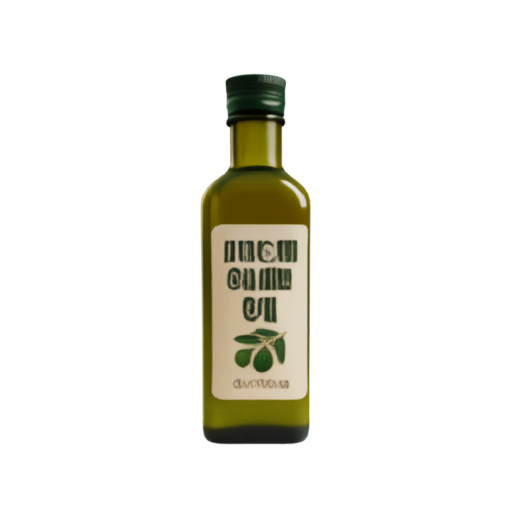}}
\end{subfigure} &
\begin{subfigure}[t]{\imgvarwidth}
\fcolorbox{lightgray}{white}{\includegraphics[width=0.95\columnwidth]{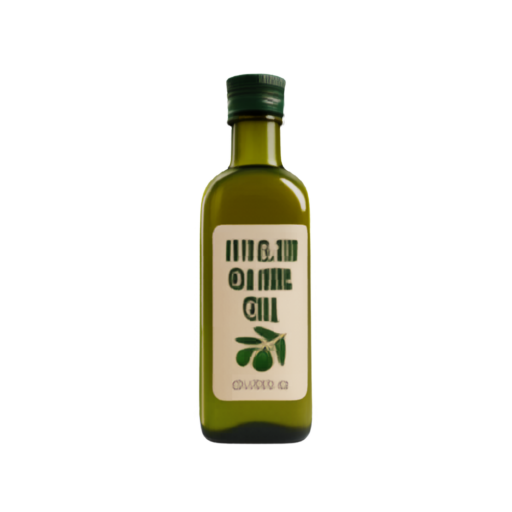}}
\end{subfigure} &
\begin{subfigure}[t]{\imgvarwidth}
\fcolorbox{lightgray}{white}{\includegraphics[width=0.95\columnwidth]{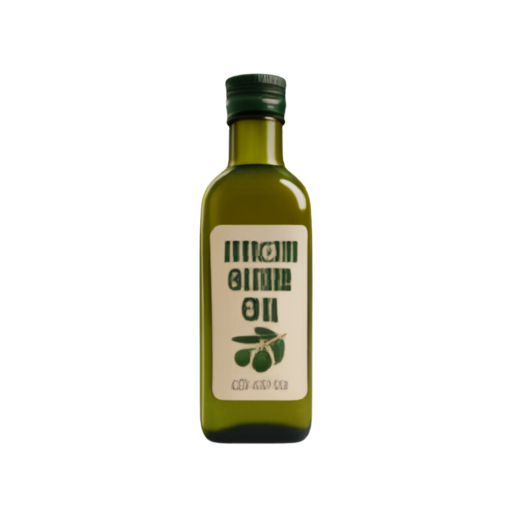}}
\end{subfigure} & 
\begin{subfigure}[t]{\imgvarwidth}
\fcolorbox{lightgray}{white}{\includegraphics[width=0.95\columnwidth]{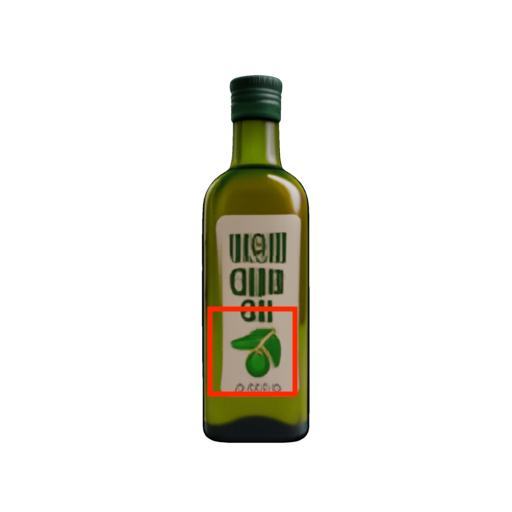}}
\end{subfigure} &
\begin{subfigure}[t]{\imgvarwidth}
\fcolorbox{lightgray}{white}{\includegraphics[width=0.95\columnwidth]{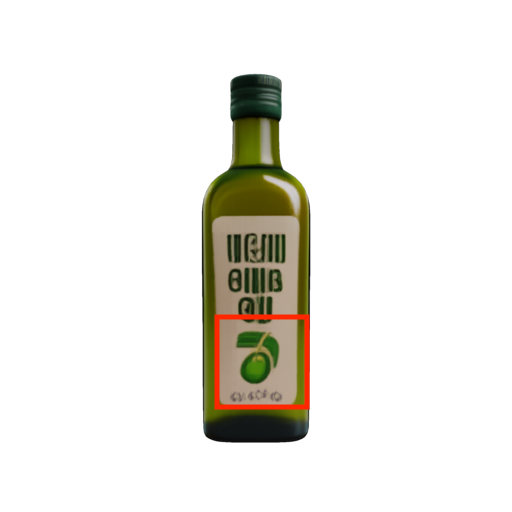}}
\end{subfigure}
\\
\begin{subfigure}[t]{\imgvarwidth}
\end{subfigure} &
\begin{subfigure}[t]{\imgvarwidth}
\fcolorbox{lightgray}{white}{\includegraphics[width=0.95\columnwidth]{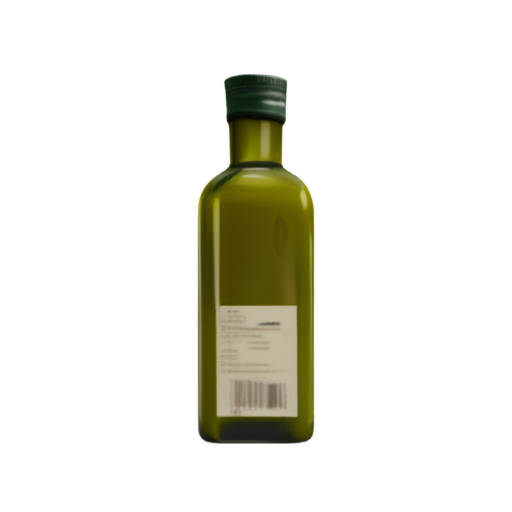}}
\end{subfigure} &
\begin{subfigure}[t]{\imgvarwidth}
\fcolorbox{lightgray}{white}{\includegraphics[width=0.95\columnwidth]{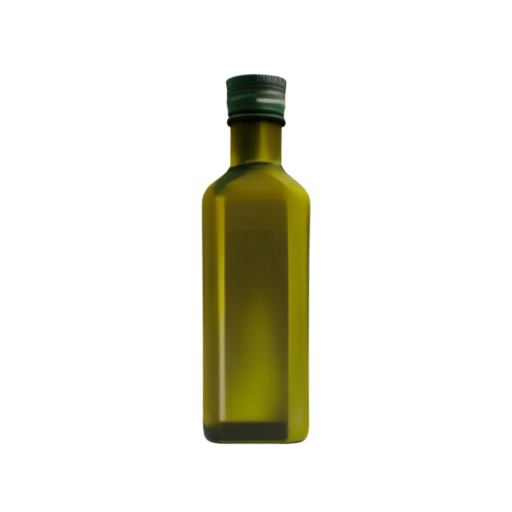}}
\end{subfigure} &
\begin{subfigure}[t]{\imgvarwidth}
\fcolorbox{lightgray}{white}{\includegraphics[width=0.95\columnwidth]{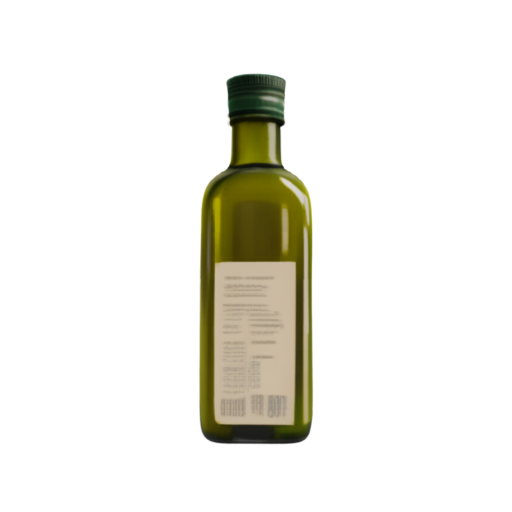}}
\end{subfigure} &
\begin{subfigure}[t]{\imgvarwidth}
\fcolorbox{lightgray}{white}{\includegraphics[width=0.95\columnwidth]{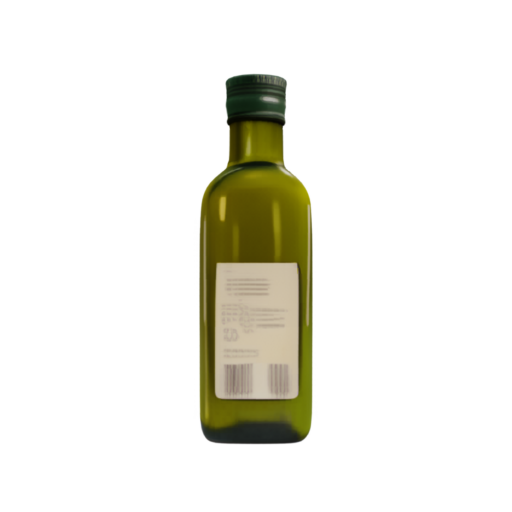}}
\end{subfigure} & 
\begin{subfigure}[t]{\imgvarwidth}
\fcolorbox{lightgray}{white}{\includegraphics[width=0.95\columnwidth]{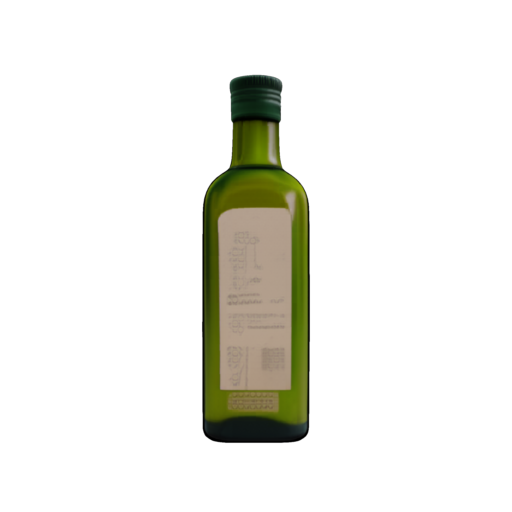}}
\end{subfigure} &
\begin{subfigure}[t]{\imgvarwidth}
\fcolorbox{lightgray}{white}{\includegraphics[width=0.95\columnwidth]{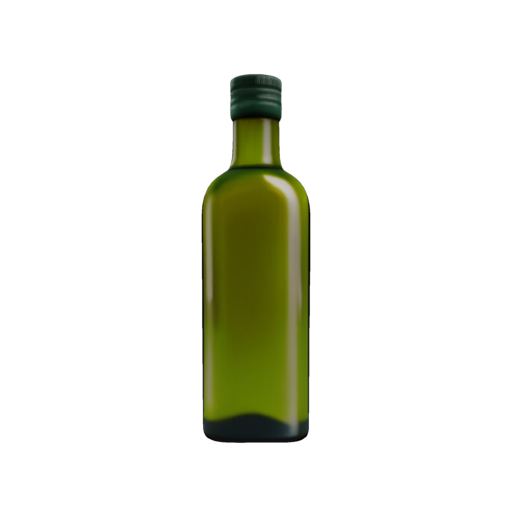}}
\end{subfigure}
\\
\begin{subfigure}[t]{\imgvarwidth}
\fcolorbox{lightgray}{white}{\includegraphics[width=0.95\columnwidth]{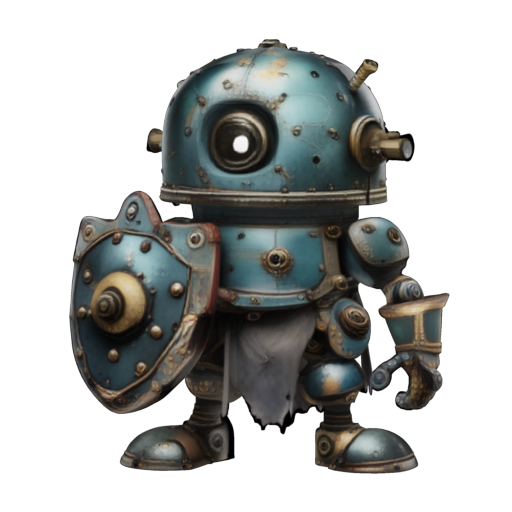}}
\end{subfigure} &
\begin{subfigure}[t]{\imgvarwidth}
\fcolorbox{lightgray}{white}{\includegraphics[width=0.95\columnwidth]{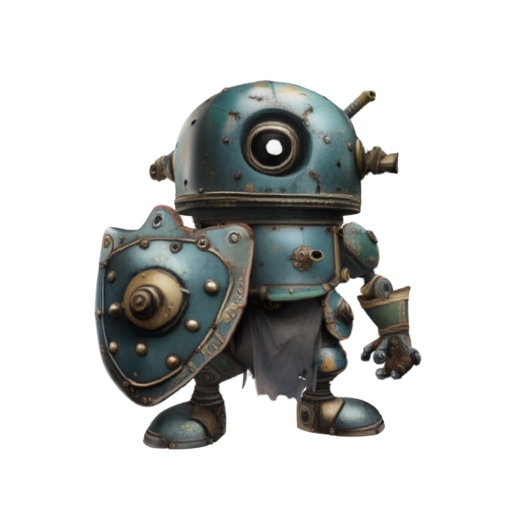}}
\end{subfigure} &
\begin{subfigure}[t]{\imgvarwidth}
\fcolorbox{lightgray}{white}{\includegraphics[width=0.95\columnwidth]{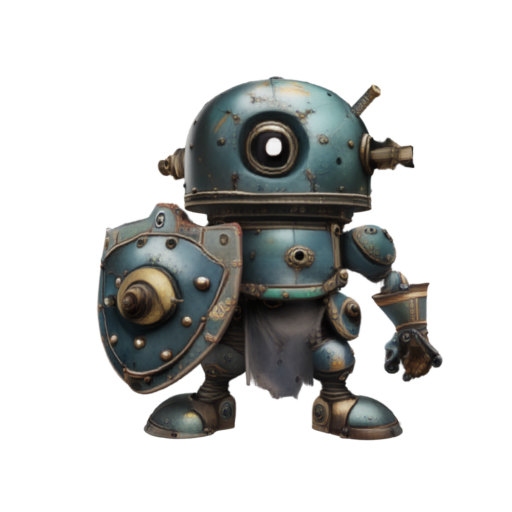}}
\end{subfigure} &
\begin{subfigure}[t]{\imgvarwidth}
\fcolorbox{lightgray}{white}{\includegraphics[width=0.95\columnwidth]{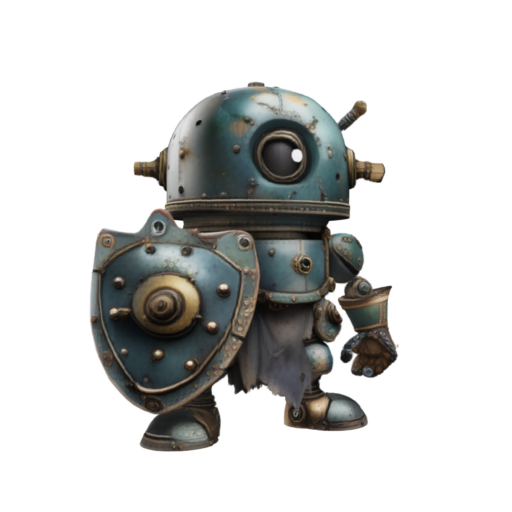}}
\end{subfigure} &
\begin{subfigure}[t]{\imgvarwidth}
\fcolorbox{lightgray}{white}{\includegraphics[width=0.95\columnwidth]{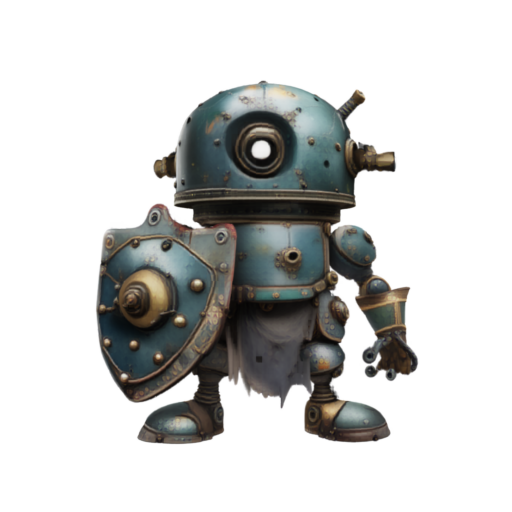}}
\end{subfigure} &
\begin{subfigure}[t]{\imgvarwidth}
\fcolorbox{lightgray}{white}{\includegraphics[width=0.95\columnwidth]{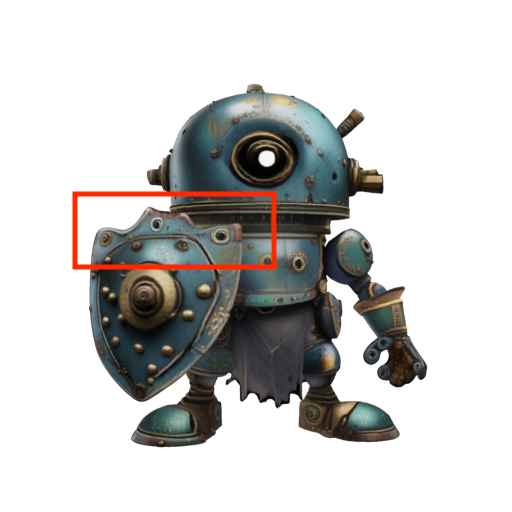}}
\end{subfigure} &
\begin{subfigure}[t]{\imgvarwidth}
\fcolorbox{lightgray}{white}{\includegraphics[width=0.95\columnwidth]{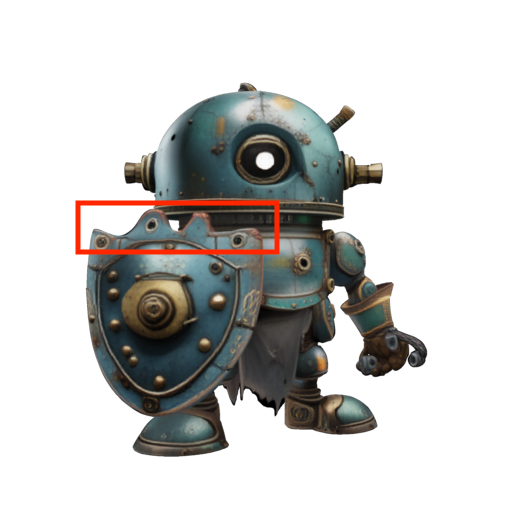}}
\end{subfigure}
\\
\begin{subfigure}[t]{\imgvarwidth}
\caption{Input}
\end{subfigure} &
\begin{subfigure}[t]{\imgvarwidth}
\fcolorbox{lightgray}{white}{\includegraphics[width=0.95\columnwidth]{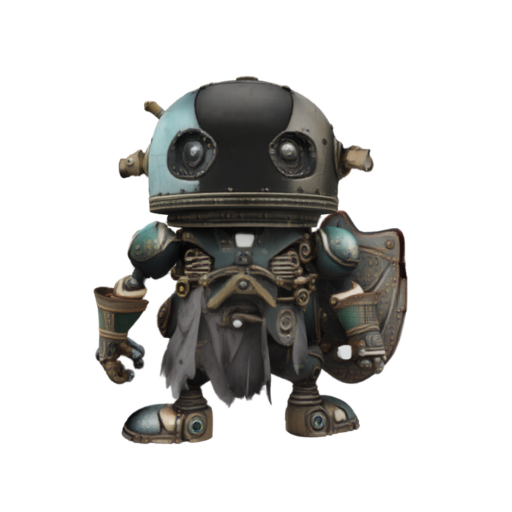}}
\caption{\small Ours (1)}
\end{subfigure} &
\begin{subfigure}[t]{\imgvarwidth}
\fcolorbox{lightgray}{white}{\includegraphics[width=0.95\columnwidth]{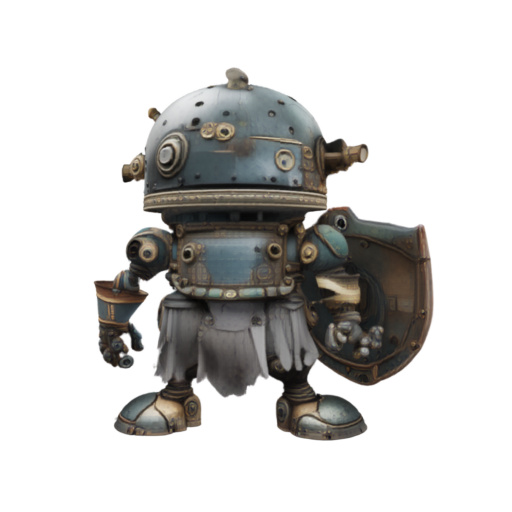}}
\caption{\small Ours (2)}
\end{subfigure} &
\begin{subfigure}[t]{\imgvarwidth}
\fcolorbox{lightgray}{white}{\includegraphics[width=0.95\columnwidth]{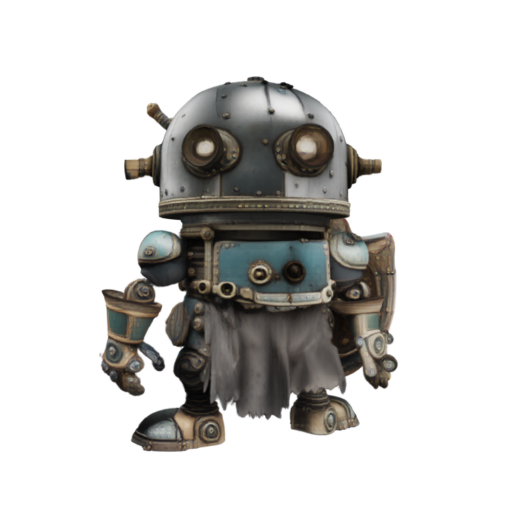}}
\caption{\small Ours (3)}
\end{subfigure} &
\begin{subfigure}[t]{\imgvarwidth}
\fcolorbox{lightgray}{white}{\includegraphics[width=0.95\columnwidth]{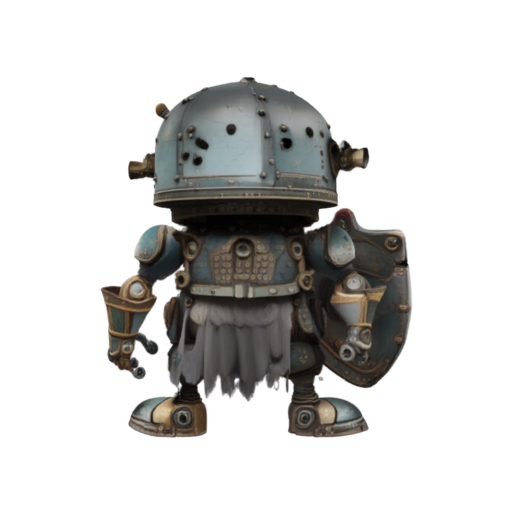}}
\caption{\small Ours (4)}
\end{subfigure} &
\begin{subfigure}[t]{\imgvarwidth}
\fcolorbox{lightgray}{white}{\includegraphics[width=0.95\columnwidth]{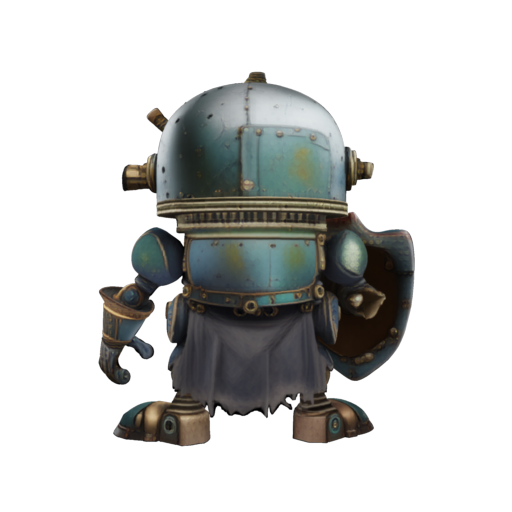}
}
\caption{\small TRELLIS (1)}
\end{subfigure} &
\begin{subfigure}[t]{\imgvarwidth}
\fcolorbox{lightgray}{white}{\includegraphics[width=0.95\columnwidth]{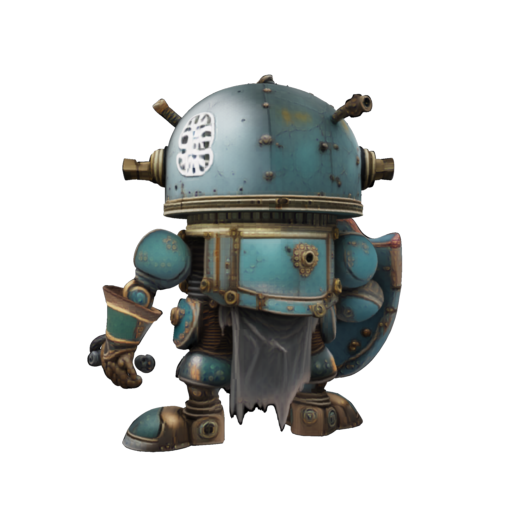}
}
\caption{\small TRELLIS (2)}
\end{subfigure}
\end{tabular}
%
%
\setlength{\fboxsep}{\oldfboxsep}
\setlength{\fboxrule}{\oldfboxrule}
}
\vspace{10pt}
\caption{
\textbf{Diversity of our \textit{Generative Reconstruction}.} We visualize canonical front and back views of generated 3D objects using random seeds (1-4). Given a single input image, our model synthesizes diverse hypotheses for unobserved regions while remaining highly consistent with visible regions. In contrast, the base 3D generator~\citep{xiang2025structured} struggles to conform to the input image, or produce less diverse unobserved regions. 
}
\label{fig:model_var}
\end{figure*}

\subsection{Implementation}
\label{sec:implementation}

\paragraph{Dataset.}
We train our model using several datasets, including ABO~\citep{collins2022abo}, HSSD~\citep{khanna2024habitat}, 3D-FUTURE~\citep{fu20213d}, and a subset of Objaverse-XL~\citep{deitke2023objaverse}, totaling approximately 260K 3D assets. These artist-curated datasets are predominantly aligned to a canonical frame where the ground plane corresponds to \(z = 0\). Following TRELLIS~\citep{xiang2025structured}, we encode each asset into occupancy grids and structured latents that are suitable for training the flow transformer. The structured latents can be decoded into high-quality triangle meshes or Gaussian splats using the SLat decoders. To enable diverse camera generation, we render 24 conditioning images from random viewpoints for each asset, with augmented focal lengths ranging from 24 mm to 200 mm. 

\paragraph{Pose Encoding and Recovery.}
Following the methodology outlined in the main text, we represent the camera pose and coarse scene geometry as a voxelized 3D occupancy grid paired with a UV coordinate volume, denoted as \(\{\point_i, \pixcoord_i(\pose)\}_{i=1}^{\numpts}\). To facilitate efficient learning, we employ a 3D VAE to compress the input UV volume (resolution \(64^3\)) into a compact latent representation (resolution \(16^3\)). These pose latents are concatenated with occupancy embeddings to condition the training of the structural flow model, \(\sflow\). For pose recovery, the latents generated by \(\sflow\) are decoded back into the UV volume. We then estimate the camera extrinsics $[\mathbf{R}|\mathbf{t}]$ and intrinsics $\mathbf{K}$ using the Perspective-n-Point (PnP) algorithm in Algo.~\ref{alg:dlt}. We evaluate the fidelity of this reconstruction in \Tabref{tab:pose_vae}, reporting the mean and 95th percentile for Reprojection Error in normalized pixel coordinates, Relative Translation Error (RTE), Relative Rotation Error (RRE), and Relative Field-of-View (RFov) error in degrees. With the solved camera pose, we finally train the latent flow model \(\lflow\) to synthesize the structured features $\{ \feat_i \}_{i=1}^{\numpts}$ conditioned on the occupancy and the pose-aligned visual features.

\vspace{9pt}
\begin{table}[h]
\centering
\caption{
\textbf{Pose Reconstruction Fidelity.}
We evaluate the accuracy of camera poses recovered from the decoded UV volumes via the PnP algorithm. We report the Reprojection Error in normalized pixels, RRE, RTE, and RFov in degrees.
}
\label{tab:pose_vae}
\renewcommand{\arraystretch}{1.0} 
\resizebox{0.9\linewidth}{!}{
\begin{tabular}{l|cccc}
\toprule
Metric & Reproj. Err. & RRE & RTE & RFov \\
\midrule
Mean & 0.0009 & 0.46 & 0.45 & 0.15 \\
95\% Quantile & 0.0011 & 1.14 & 1.14 & 0.31 \\
\bottomrule
\end{tabular}
}
\end{table}

\begin{algorithm}[t]
\caption{Perspective-n-Point solver via Direct Linear Transformation (DLT)}
\label{alg:dlt}
$\mathcal{X}$: Sets of 3D coordinates $\{\mathbf{x}_i\}_{i=1}^N$ \\
$\mathcal{U}$: 2D coordinates $\{\mathbf{u}_i\}_{i=1}^N$
\smallskip
\hrule 
\begin{algorithmic} 
    \State $N \gets |\mathcal{X}|$ \Comment{Number of correspondences}
    \State $\mathbf{A} \gets \mathbf{0}_{2N \times 12}$ \Comment{Initialize coefficient matrix}
    \For{$i = 1$ to $N$}
        \State $\tilde{\mathbf{x}}_i \gets [\mathbf{x}_i^\top, 1]^\top$ \Comment{Homogeneous coordinates}
        \State $u_i, v_i \gets \mathbf{u}_i$
        \State $\mathbf{A}[2i-1, :] \gets [\tilde{\mathbf{x}}_i^\top, \mathbf{0}^\top, -u_i \tilde{\mathbf{x}}_i^\top]$ 
        \State $\mathbf{A}[2i, :] \quad \gets [\mathbf{0}^\top, \tilde{\mathbf{x}}_i^\top, -v_i \tilde{\mathbf{x}}_i^\top]$ 
    \EndFor
    \State $[\mathbf{U}, \mathbf{\Sigma}, \mathbf{V}^\top] \gets \text{SVD}(\mathbf{A})$
    \State $\mathbf{p} \gets \mathbf{V}[:, -1]$ \Comment{Solution is the last column of $\mathbf{V}$}
    \State $\mathbf{P} \gets \text{reshape}(\mathbf{p}, 3, 4)$ \Comment{Camera matrix $\mathbf{P} \in \mathbb{R}^{3 \times 4}$}
    \If{$\det(\mathbf{P}_{1:3, 1:3}) < 0$} \Comment{Fix scale/sign ambiguity}
        \State $\mathbf{P} \gets -\mathbf{P}$
    \EndIf
    \State $[\mathbf{K}, \mathbf{R}, \mathbf{t}] \gets \text{RQ}(\mathbf{P})$ \Comment{Decompose camera matrix}
    \State \Return $\mathbf{K}, \mathbf{R}, \mathbf{t}$
\end{algorithmic}
\end{algorithm}

\paragraph{Training Details.}

We initialize our models using the pretrained weights of TRELLIS. During training, we apply classifier-free guidance~\citep{ho2022classifier} (CFG) with a drop rate of 0.1. Both \(\sflow\) and \(\lflow\) are trained using AdamW~\citep{loshchilov2017decoupled} at a fixed learning rate of \(1 \times 10^{-4}\) for 500k and 100k steps, respectively. Training completes in approximately one week on 32 GPUs. At inference, we use 25 sampling steps with classifier-free guidance strengths of 7.5 and 3.0 for $\sflow$ and $\lflow$, respectively.

\definecolor{lightgray}{gray}{0.8}
\newcommand{\imgcolorwidth}{.16\textwidth}
\begin{figure*}[htb]
\captionsetup[subfigure]{labelformat=empty}
\centering
\makebox[\textwidth][c]{
\renewcommand{\arraystretch}{1.2}
\setlength{\oldfboxsep}{\fboxsep}
\setlength{\oldfboxrule}{\fboxrule}
\setlength{\fboxsep}{1pt} 
\setlength{\fboxrule}{1pt} 
\begin{tabular}{@{}c@{\betweencols}c@{\betweencols}c@{\betweencols}c@{\betweencols}c@{\betweencols}c@{}}
\begin{subfigure}[t]{\imgcolorwidth}
\fcolorbox{lightgray}{white}{\includegraphics[width=0.95\columnwidth]{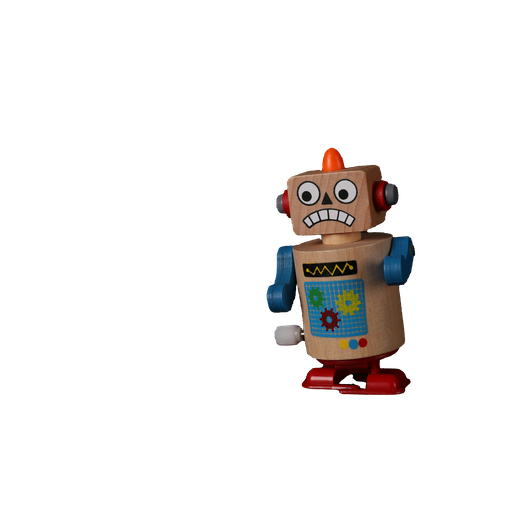}}
\end{subfigure} &
\begin{subfigure}[t]{\imgcolorwidth}
\fcolorbox{lightgray}{white}{\includegraphics[width=0.95\columnwidth]{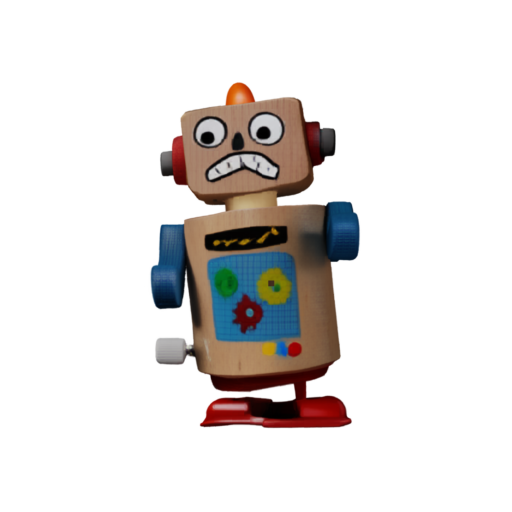}}
\end{subfigure} &
\begin{subfigure}[t]{\imgcolorwidth}
\fcolorbox{lightgray}{white}{\includegraphics[width=0.95\columnwidth]{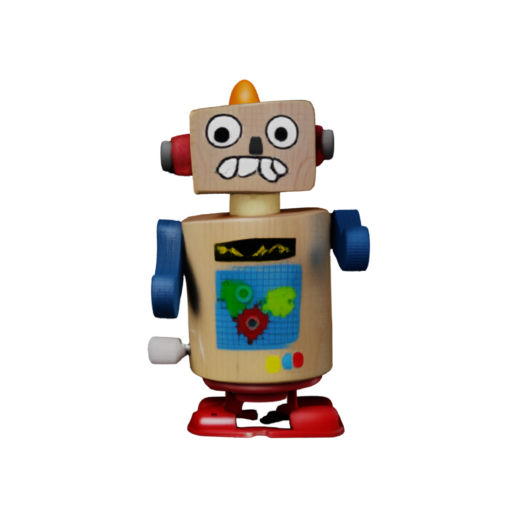}}
\end{subfigure} &
\begin{subfigure}[t]{\imgcolorwidth}
\fcolorbox{lightgray}{white}{\includegraphics[width=0.95\columnwidth]{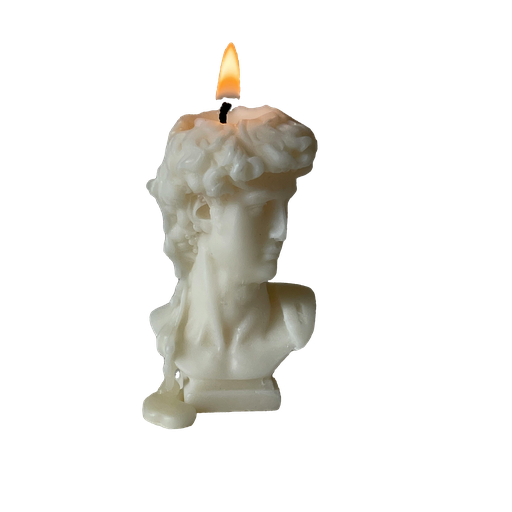}}
\end{subfigure} &
\begin{subfigure}[t]{\imgcolorwidth}
\fcolorbox{lightgray}{white}{\includegraphics[width=0.95\columnwidth]{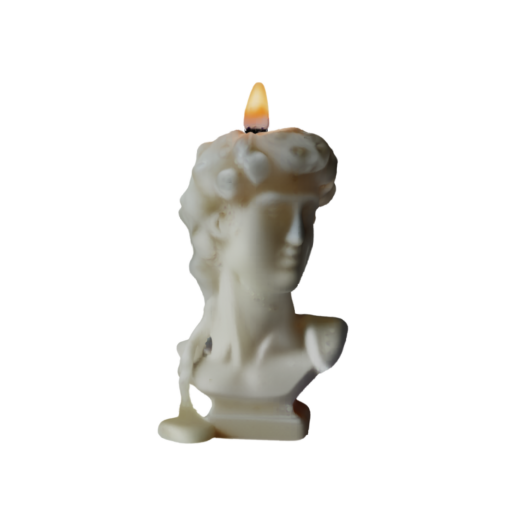}}
\end{subfigure} &
\begin{subfigure}[t]{\imgcolorwidth}
\fcolorbox{lightgray}{white}{\includegraphics[width=0.95\columnwidth]{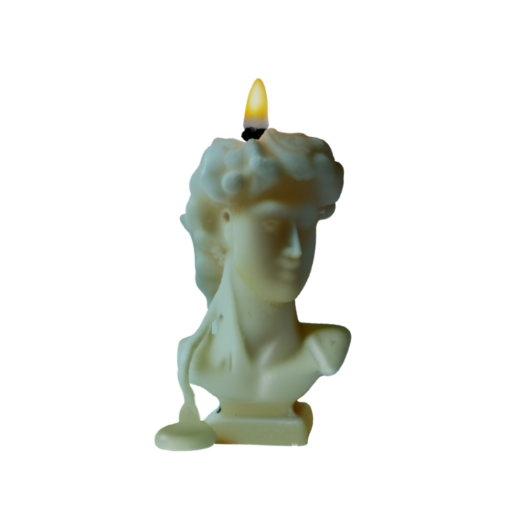}}
\end{subfigure}
\\
\begin{subfigure}[t]{\imgcolorwidth}
\end{subfigure} &
\begin{subfigure}[t]{\imgcolorwidth}
\fcolorbox{lightgray}{white}{\includegraphics[width=0.95\columnwidth]{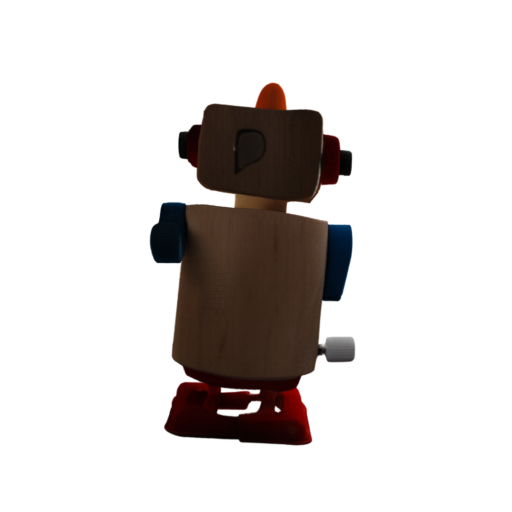}}
\end{subfigure} &
\begin{subfigure}[t]{\imgcolorwidth}
\fcolorbox{lightgray}{white}{\includegraphics[width=0.95\columnwidth]{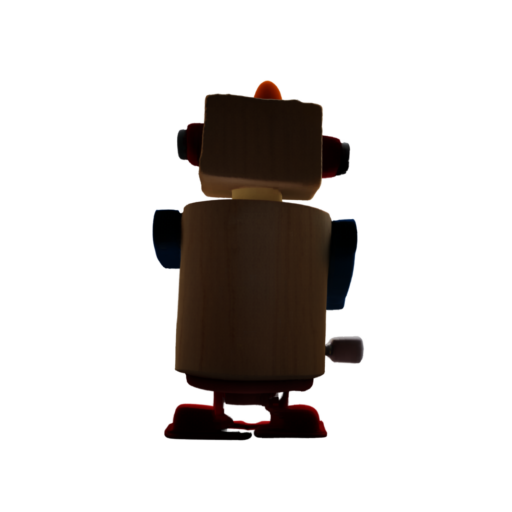}}
\end{subfigure} &
\begin{subfigure}[t]{\imgcolorwidth}
\end{subfigure} &
\begin{subfigure}[t]{\imgcolorwidth}
\fcolorbox{lightgray}{white}{\includegraphics[width=0.95\columnwidth]{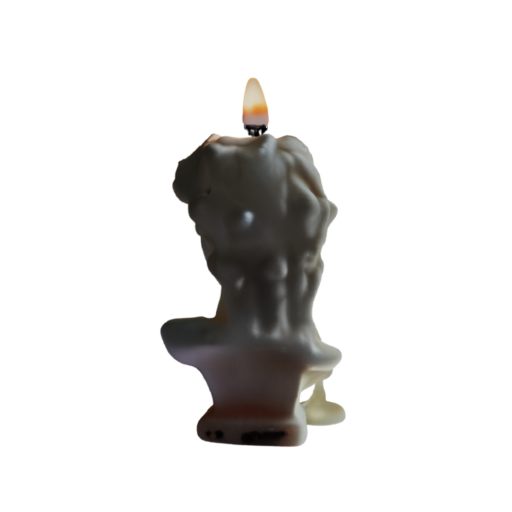}}
\end{subfigure} &
\begin{subfigure}[t]{\imgcolorwidth}
\fcolorbox{lightgray}{white}{\includegraphics[width=0.95\columnwidth]{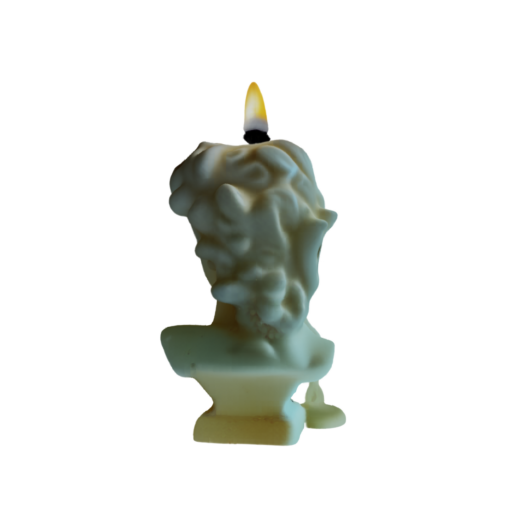}}
\end{subfigure}
\\
\begin{subfigure}[t]{\imgcolorwidth}
\fcolorbox{lightgray}{white}{\includegraphics[width=0.95\columnwidth]{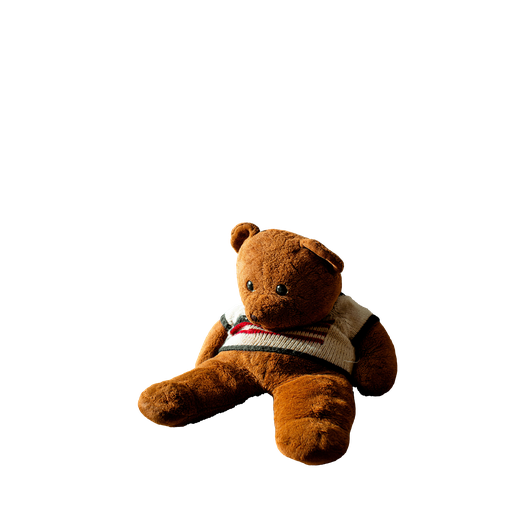}}
\end{subfigure} &
\begin{subfigure}[t]{\imgcolorwidth}
\fcolorbox{lightgray}{white}{\includegraphics[width=0.95\columnwidth]{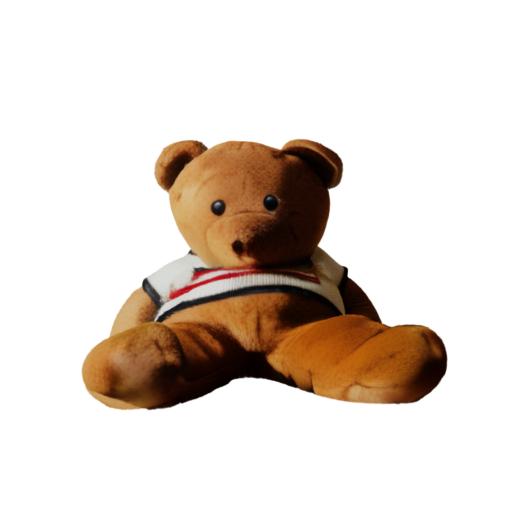}}
\end{subfigure} &
\begin{subfigure}[t]{\imgcolorwidth}
\fcolorbox{lightgray}{white}{\includegraphics[width=0.95\columnwidth]{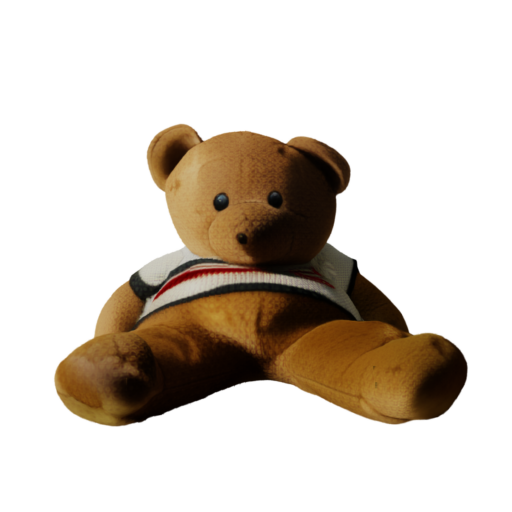}}
\end{subfigure} &
\begin{subfigure}[t]{\imgcolorwidth}
\fcolorbox{lightgray}{white}{\includegraphics[width=0.95\columnwidth]{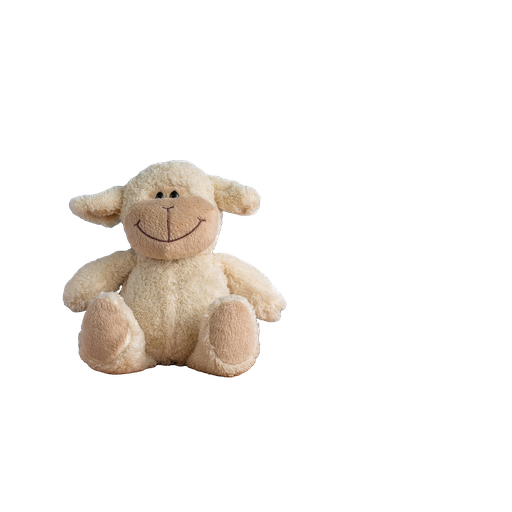}}
\end{subfigure} &
\begin{subfigure}[t]{\imgcolorwidth}
\fcolorbox{lightgray}{white}{\includegraphics[width=0.95\columnwidth]{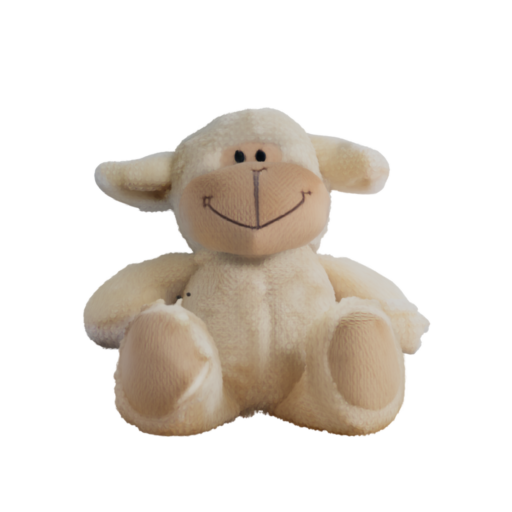}}
\end{subfigure} &
\begin{subfigure}[t]{\imgcolorwidth}
\fcolorbox{lightgray}{white}{\includegraphics[width=0.95\columnwidth]{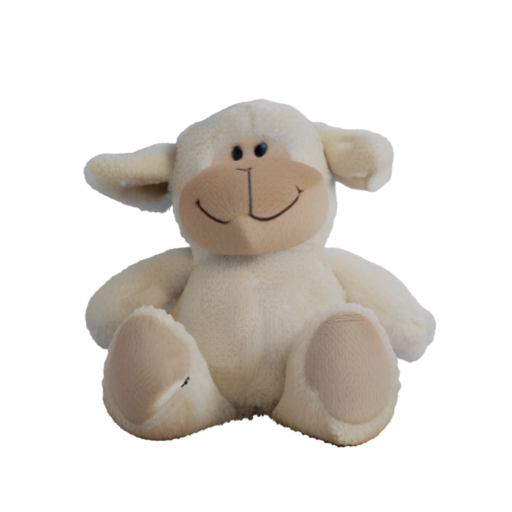}}
\end{subfigure}
\\
\begin{subfigure}[t]{\imgcolorwidth}
\caption{\small Input Image}
\end{subfigure} &
\begin{subfigure}[t]{\imgcolorwidth}
\fcolorbox{lightgray}{white}{\includegraphics[width=0.95\columnwidth]{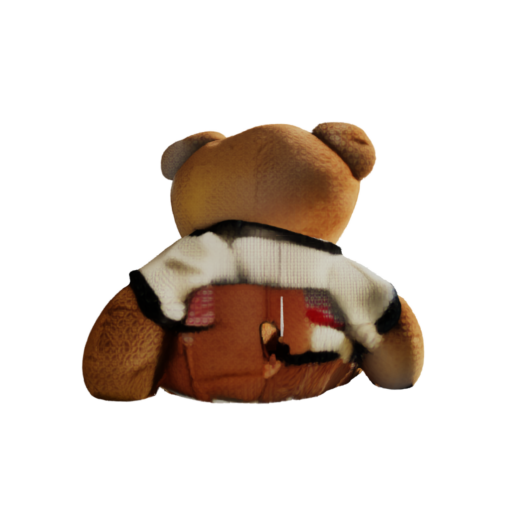}}
\caption{\small Ours}
\end{subfigure} &
\begin{subfigure}[t]{\imgcolorwidth}
\fcolorbox{lightgray}{white}{\includegraphics[width=0.95\columnwidth]{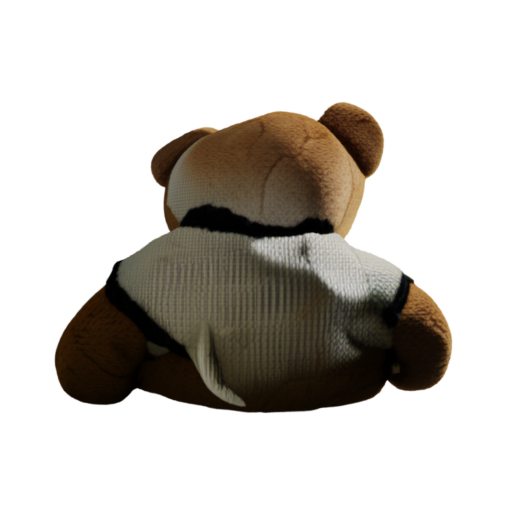}}
\caption{\small TRELLIS}
\end{subfigure} &
\begin{subfigure}[t]{\imgcolorwidth}
\caption{\small Input Image}
\end{subfigure} &
\begin{subfigure}[t]{\imgcolorwidth}
\fcolorbox{lightgray}{white}{\includegraphics[width=0.95\columnwidth]{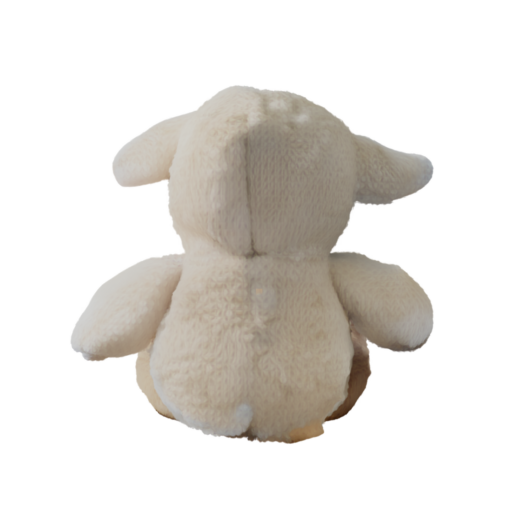}}
\caption{\small Ours}
\end{subfigure} & 
\begin{subfigure}[t]{\imgcolorwidth}
\fcolorbox{lightgray}{white}{\includegraphics[width=0.95\columnwidth]{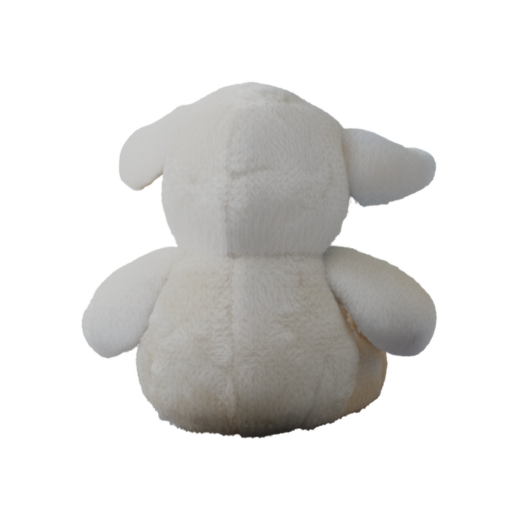}}
\caption{\small TRELLIS}
\end{subfigure}
\end{tabular}
\setlength{\fboxsep}{\oldfboxsep}
\setlength{\fboxrule}{\oldfboxrule}
}
\vspace{10pt}
\caption{
\textbf{Qualitative comparison: \textit{Generative Reconstruction} vs. \textit{3D Generation}.}
Unlike standard 3D generation, which aims to create novel objects from images, our generative reconstruction is specifically designed to accurately recreate a particular object that replicates the visible regions of the input image while maintaining diversity in the invisible regions. This difference in objectives is crucial, as prior 3D generators~\citep{xiang2025structured} are not optimized for this task and often produce artifacts such as \textit{color drift} and \textit{texture inconsistencies}. As demonstrated in the canonical front and back views, our method's pose-aligned image conditioning effectively reduces these issues, resulting in a reconstruction that remains true to the input.  
}
\label{fig:color_drifting}
\end{figure*}

\subsection{Occlusion-aware conditioning}
\label{sec:occlusion}

To handle occlusion in complex scene reconstruction, our model leverages partial 3D object observations as conditions to generate complete objects. Our model takes a 2D occlusion mask \(\mask \in\{0,1\}^{H \times W}\) as input alongside the visible object observation \(\image\).
The mask \(\mask\) identifies pixels that may belong to the object if occluders were removed, with values set to zero when no occlusion is present. Together, \(\mask\) and the alpha channel of \(\image\) identify three pixel classes:
(a) directly observed object pixels, (b) background pixels that must not contain the object, and (c) occluded pixels that may or may not contain the object.

We apply two modifications to both flow transformers to incorporate the mask.
First, inspired by Amodal3R~\citep{wu2025amodal3r}, we modulate the attention weight matrix during global condition injection via cross-attention using the mask. Specifically, the attention weight for each input token is computed by patching the mask to match the DINOv2 tokens and calculating the ratio of unmasked pixels in each patch. The logarithm of the weight values is added to the attention logits before applying the softmax operation. Second, for the geometry and appearance flow model that takes additional pose-aligned features, we concatenate the input image with the mask as an additional channel before feeding it into the convolution layer in the second stage.

During training, we randomly generate occlusion masks \(\mask\) following Amodal3R and zero out the corresponding regions in \(\image\), preventing information leakage on occlusion regions and encouraging the model to reconstruct complete 3D objects from partial observations. At inference time, occlusion masks can be obtained heuristically or manually for scene reconstruction.

\newcommand{\suppwidth}{.3\textwidth}
\begin{figure*}[t]
\captionsetup[subfigure]{labelformat=empty}
\centering
\makebox[\textwidth][c]{
\renewcommand{\arraystretch}{0.4}
\begin{tabular}{@{}c@{\betweencols}c@{\betweencols}c@{\betweencols}c@{}}
\begin{subfigure}[t]{\suppwidth}
\includegraphics[width=\columnwidth]{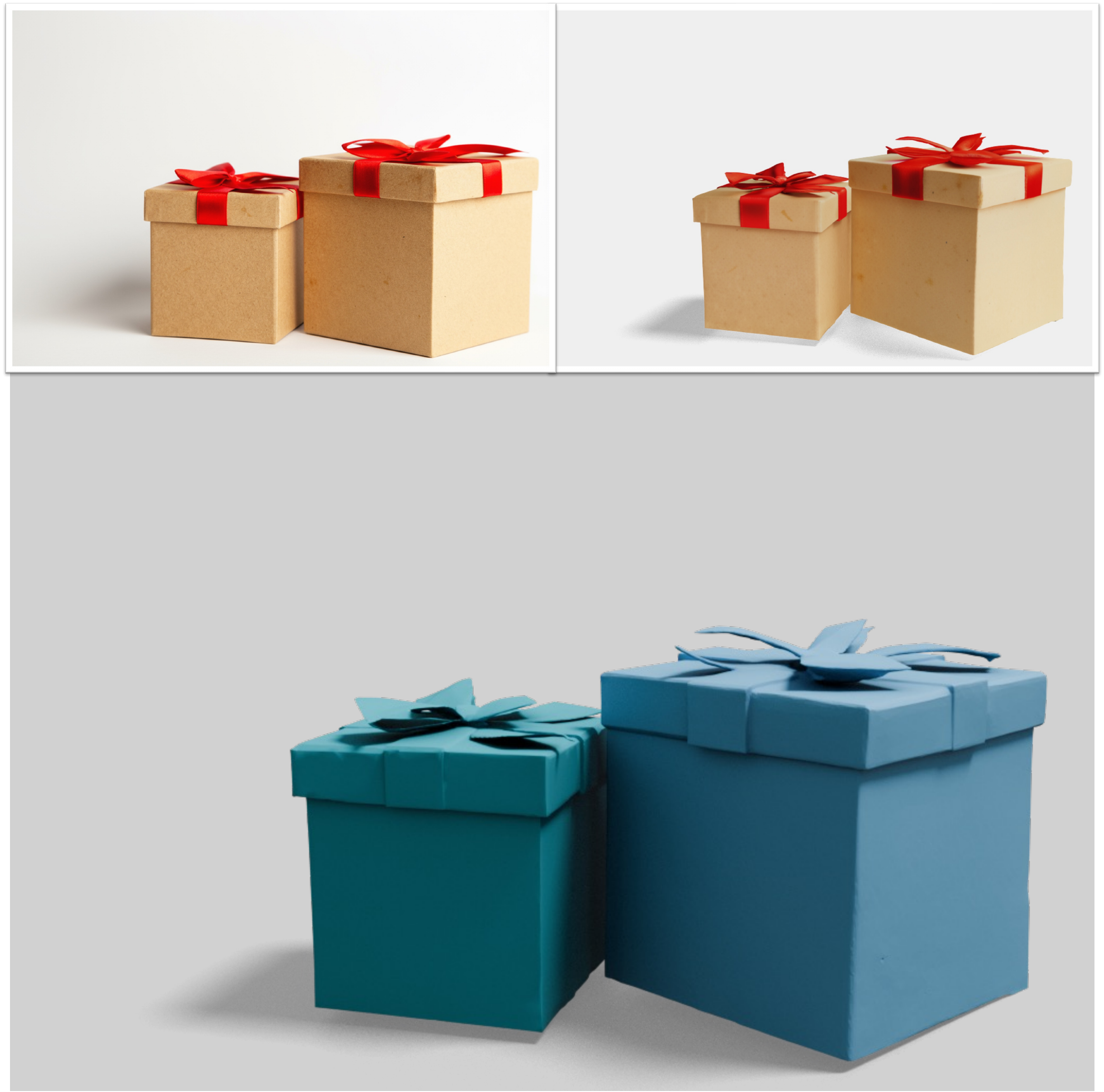}
\end{subfigure} & 
\begin{subfigure}[t]{\suppwidth}
\includegraphics[width=\columnwidth]{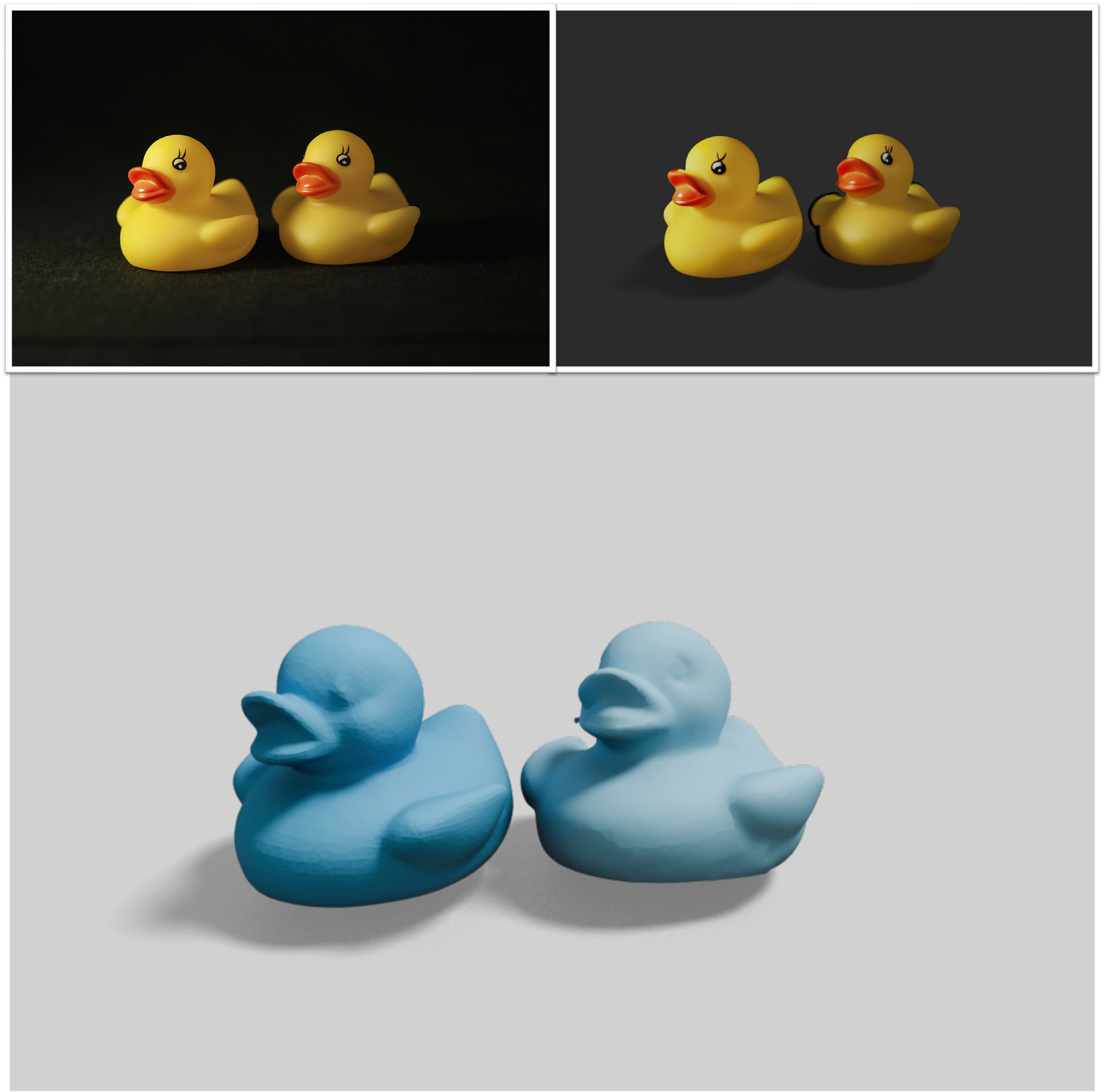}
\end{subfigure} &
\begin{subfigure}[t]{\suppwidth}
\includegraphics[width=\columnwidth]{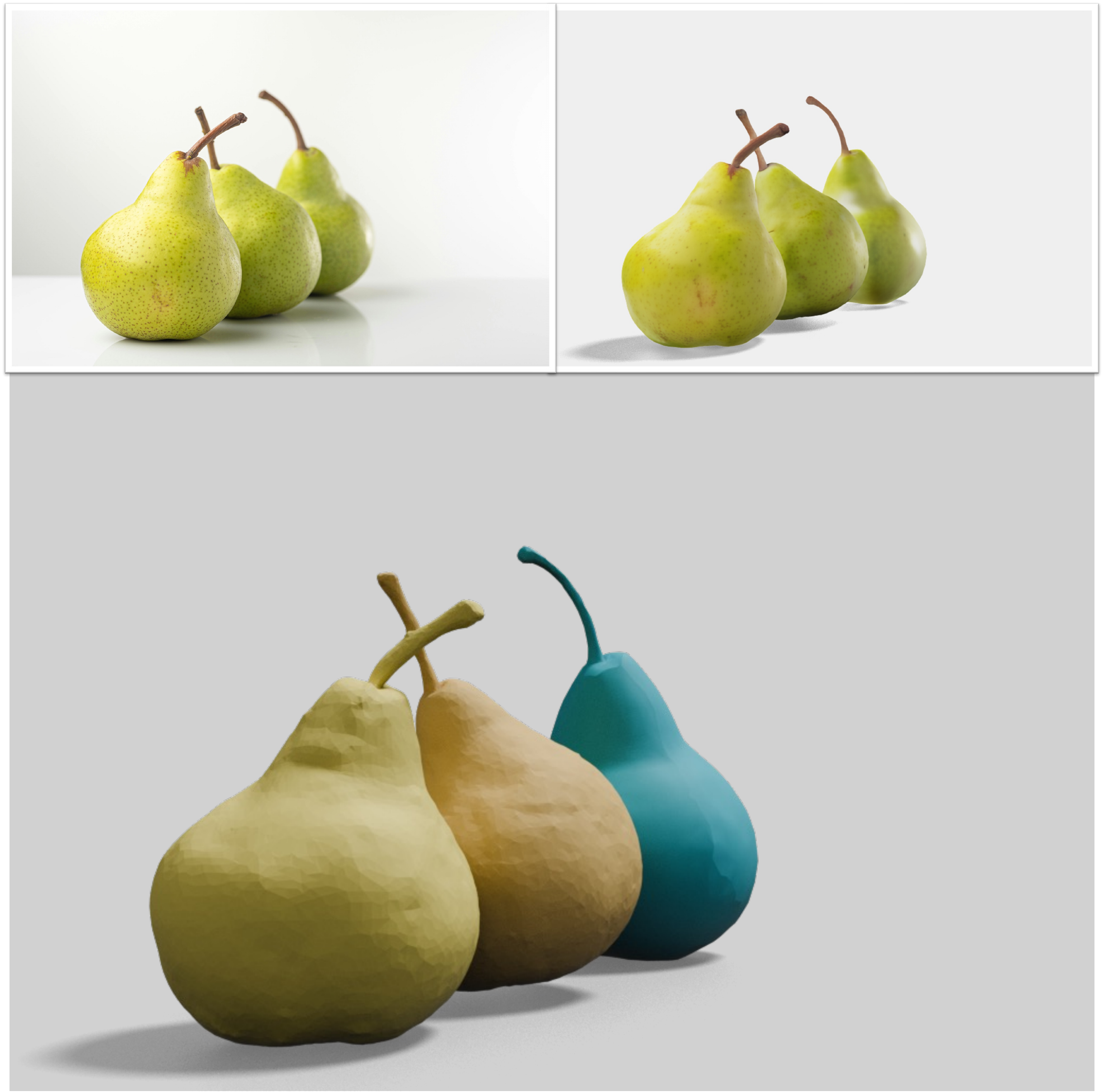}
\end{subfigure} \\
\begin{subfigure}[t]{\suppwidth}
\includegraphics[width=\columnwidth]{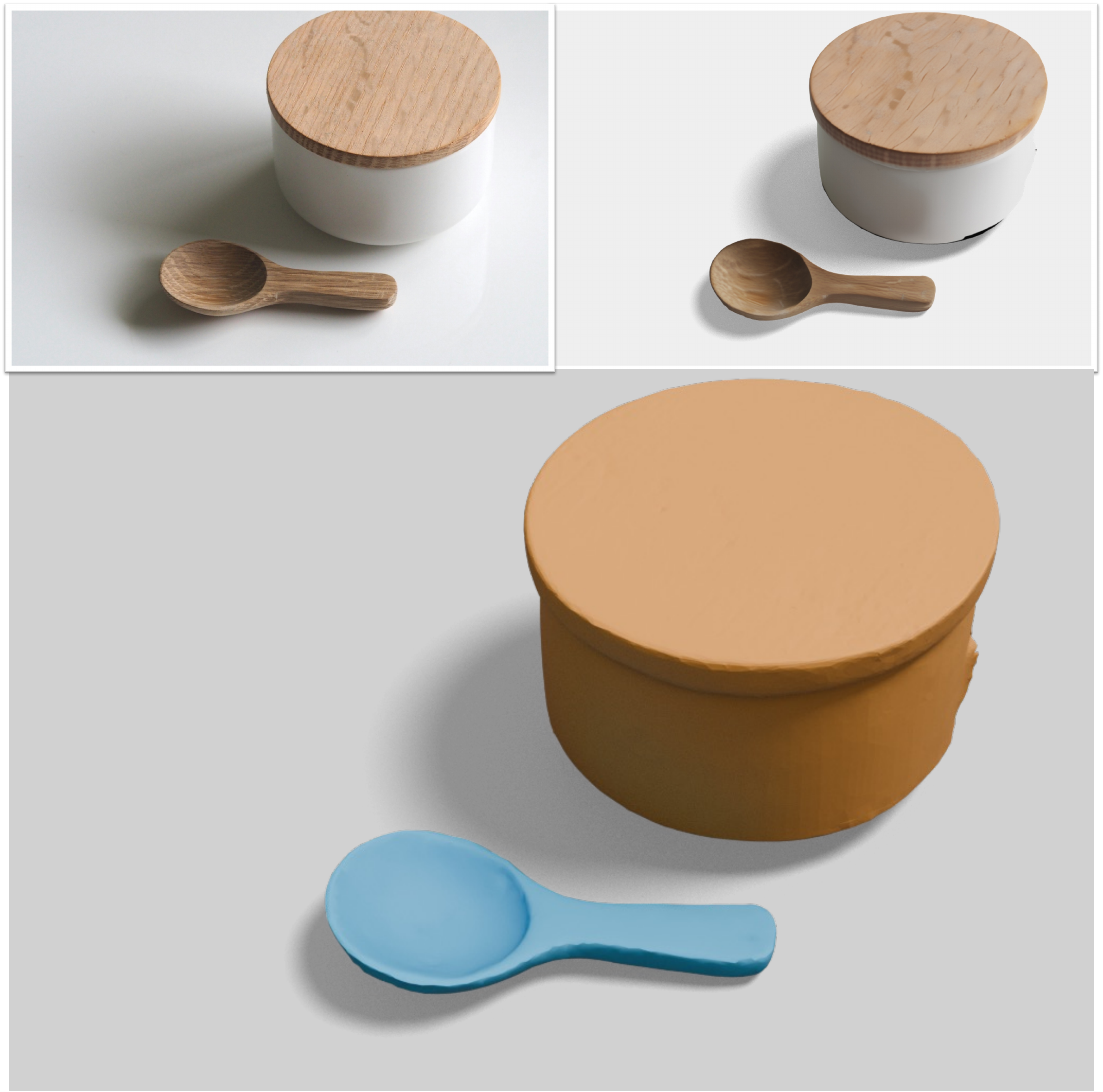}
\end{subfigure} &
\begin{subfigure}[t]{\suppwidth}
\includegraphics[width=\columnwidth]{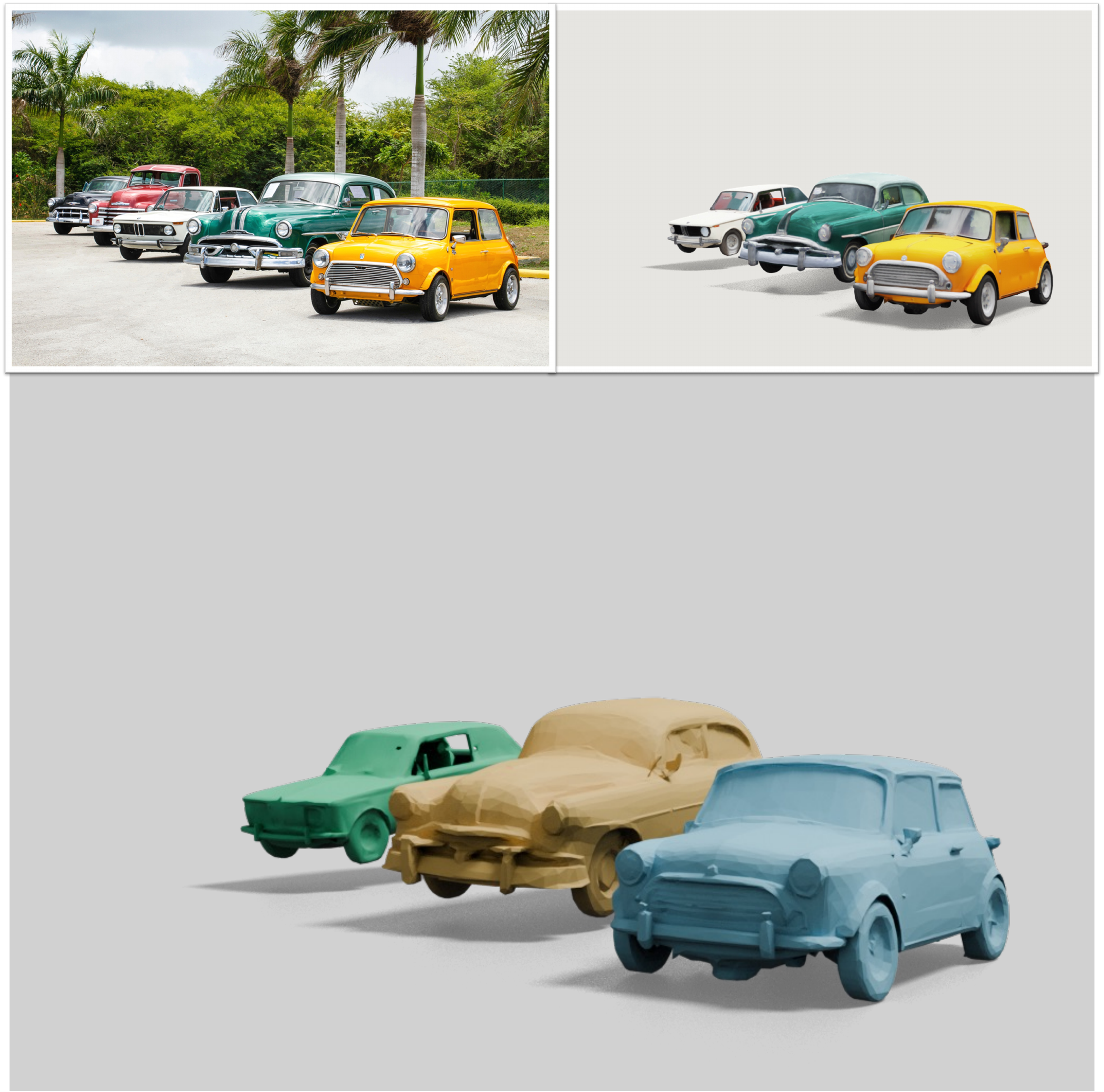}
\end{subfigure} & 
\begin{subfigure}[t]{\suppwidth}
\includegraphics[width=\columnwidth]{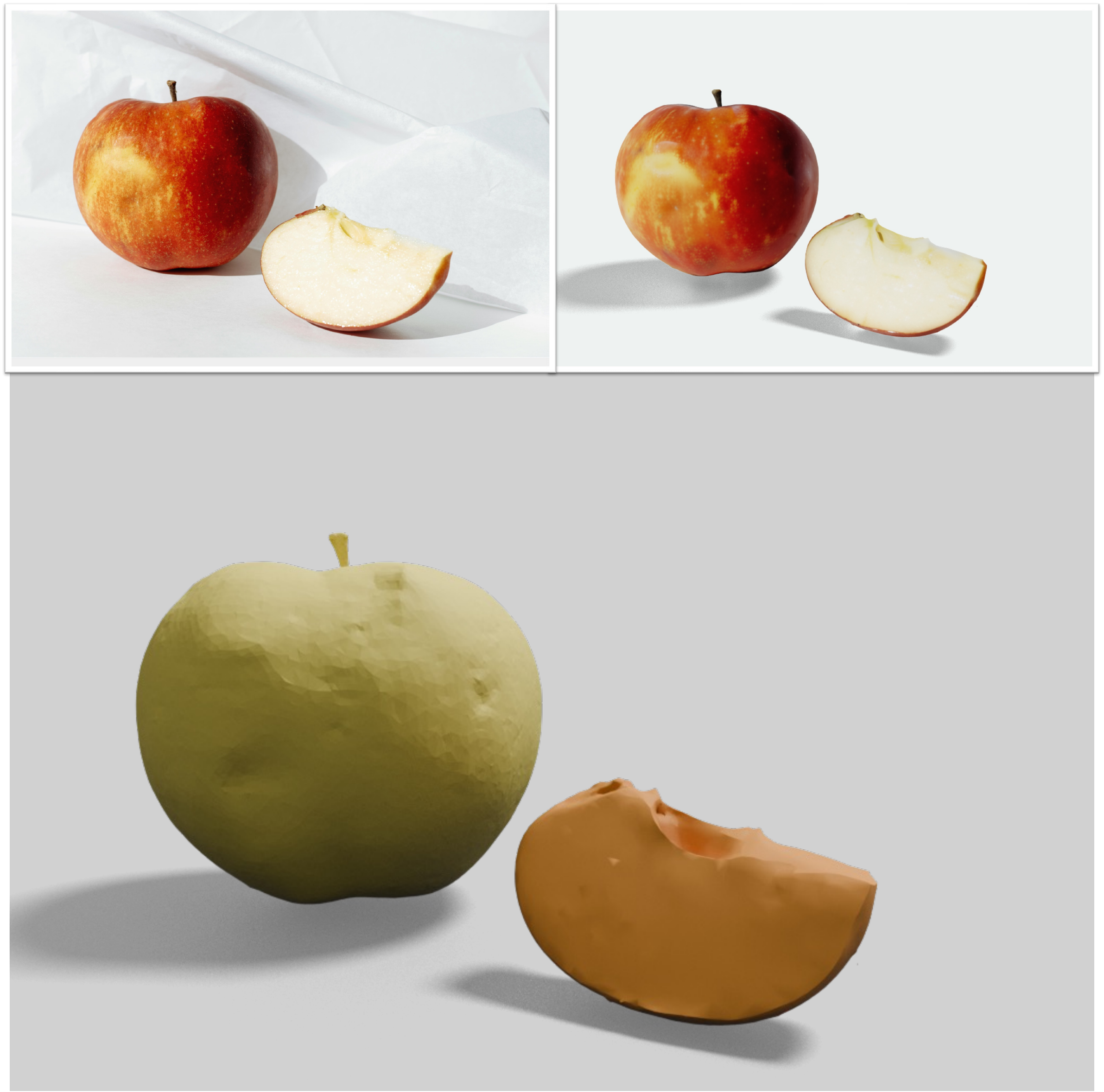}
\end{subfigure} \\
\begin{subfigure}[t]{\suppwidth}
\includegraphics[width=\columnwidth]{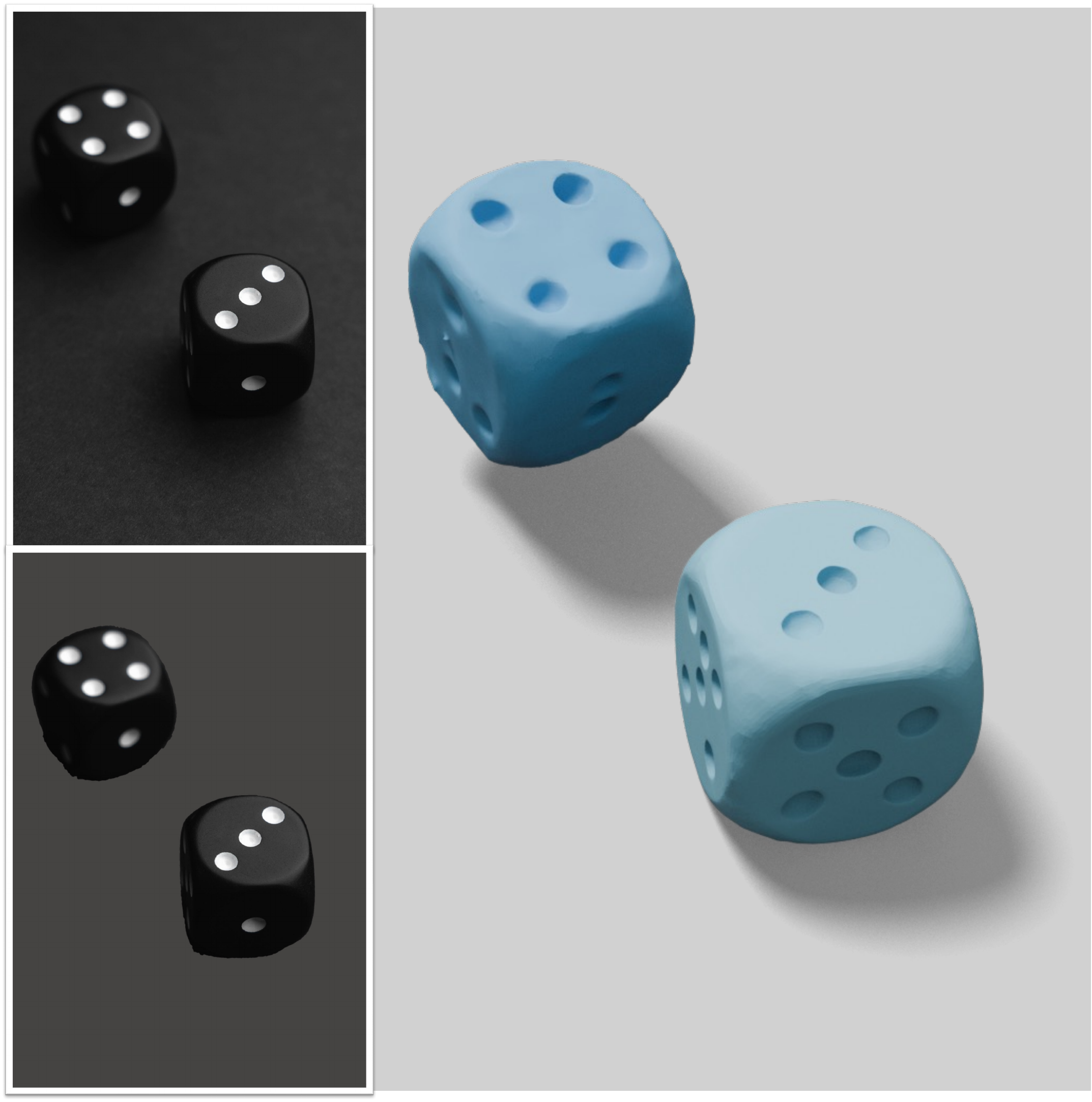}
\end{subfigure} &
\begin{subfigure}[t]{\suppwidth}
\includegraphics[width=\columnwidth]{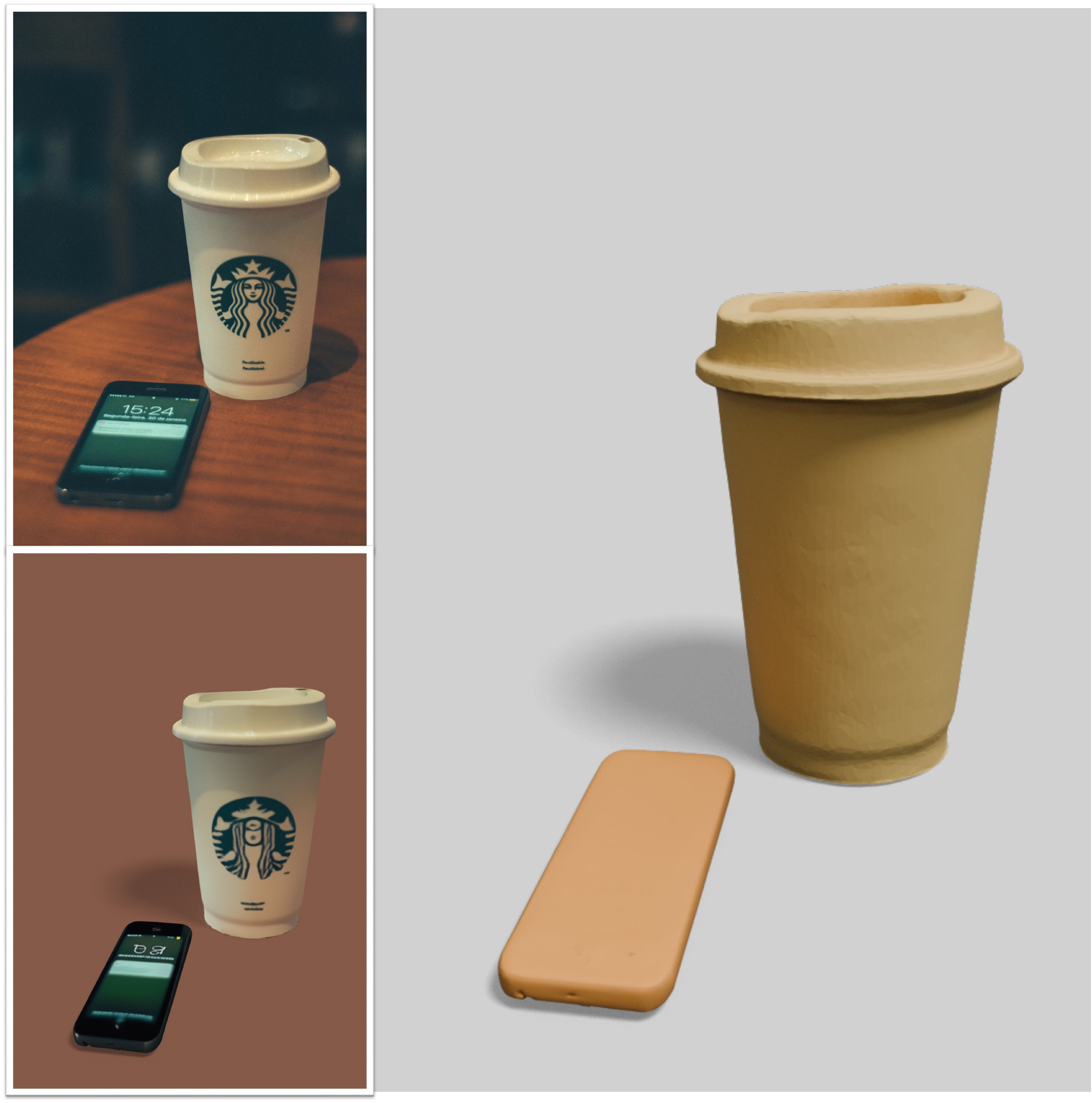}
\end{subfigure} &
\begin{subfigure}[t]{\suppwidth}
\includegraphics[width=\columnwidth]{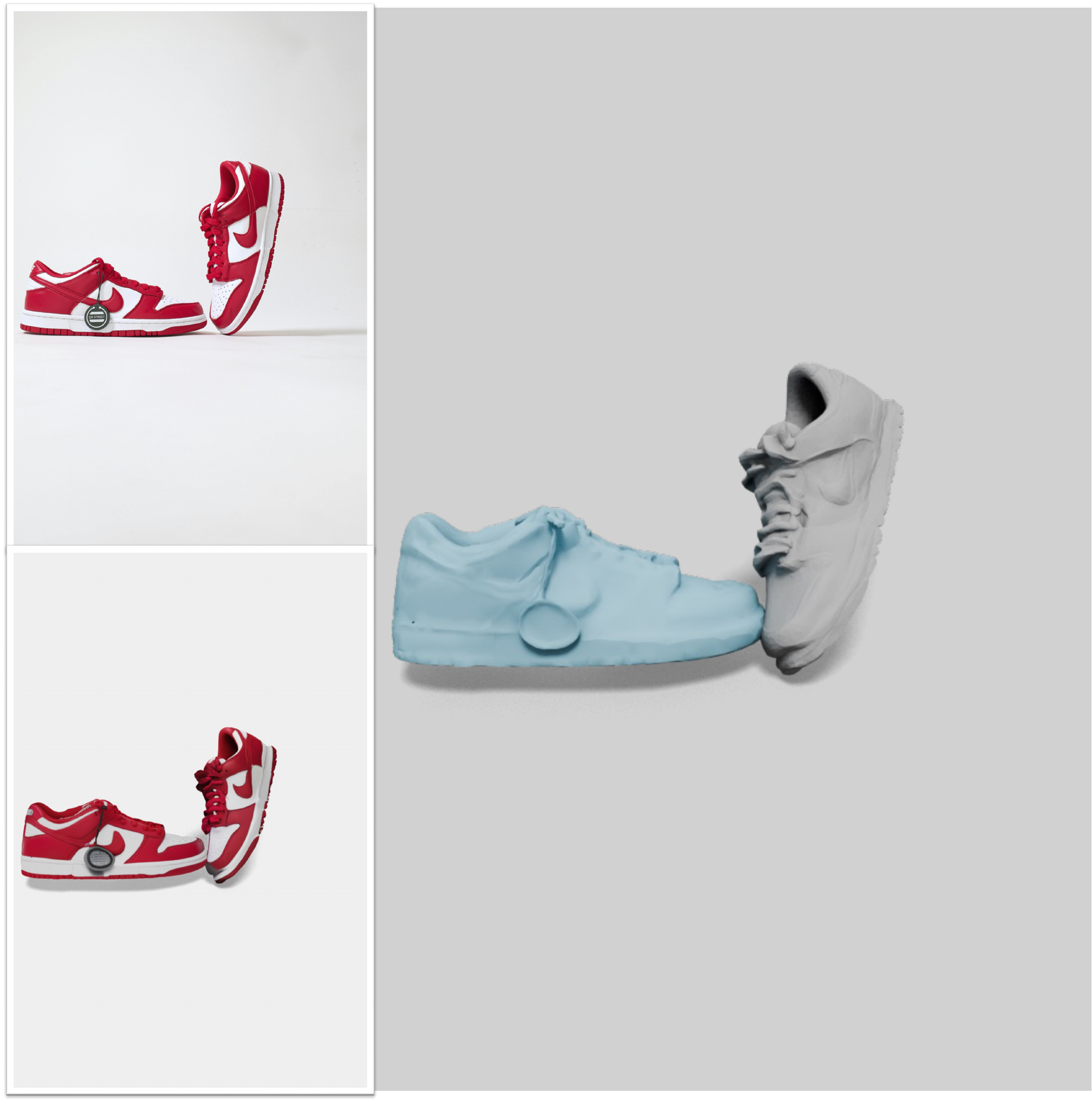}
\end{subfigure} \\ 
\begin{subfigure}[t]{\suppwidth}
\includegraphics[width=\columnwidth]{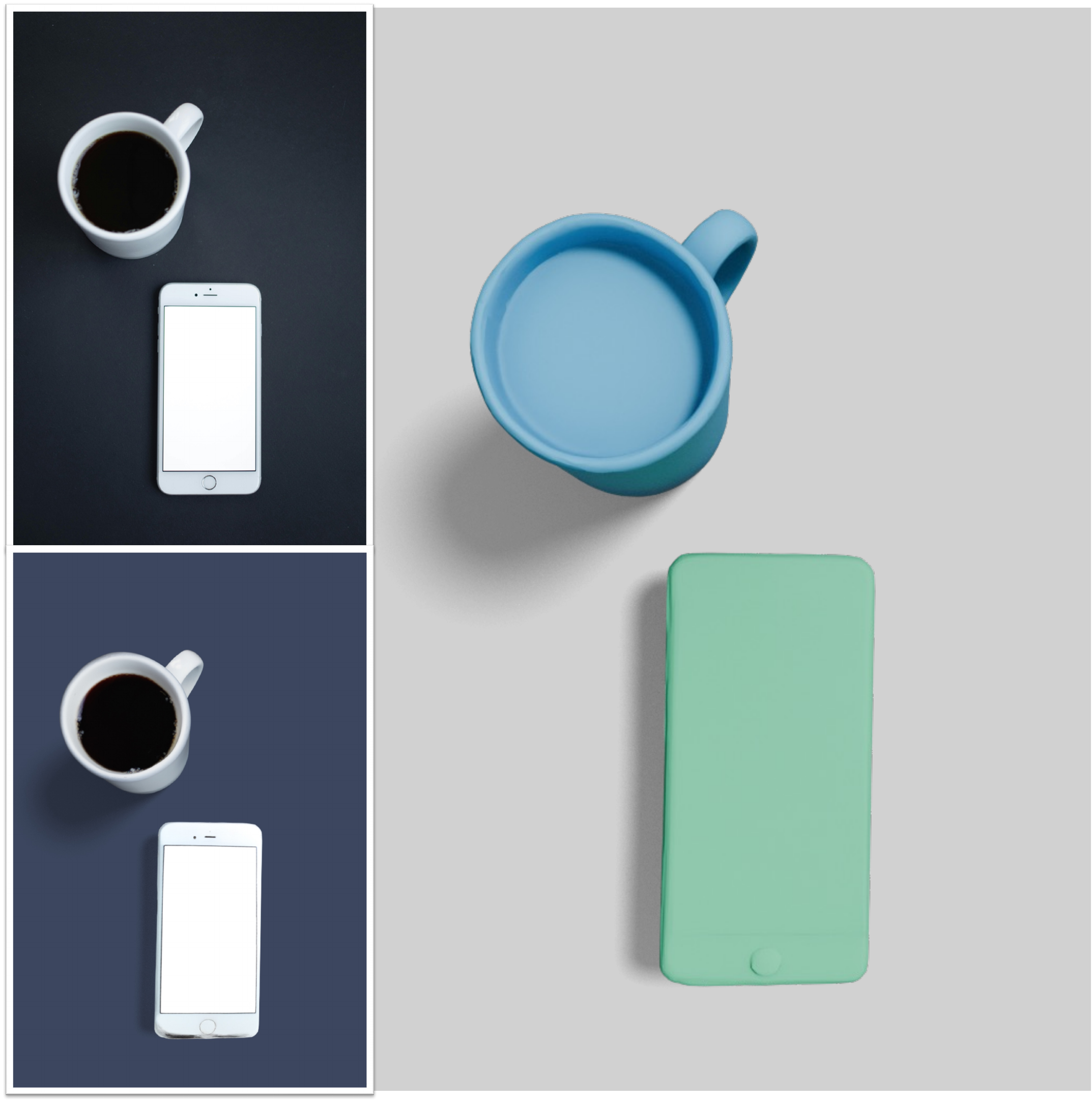}
\end{subfigure} &
\begin{subfigure}[t]{\suppwidth}
\includegraphics[width=\columnwidth]{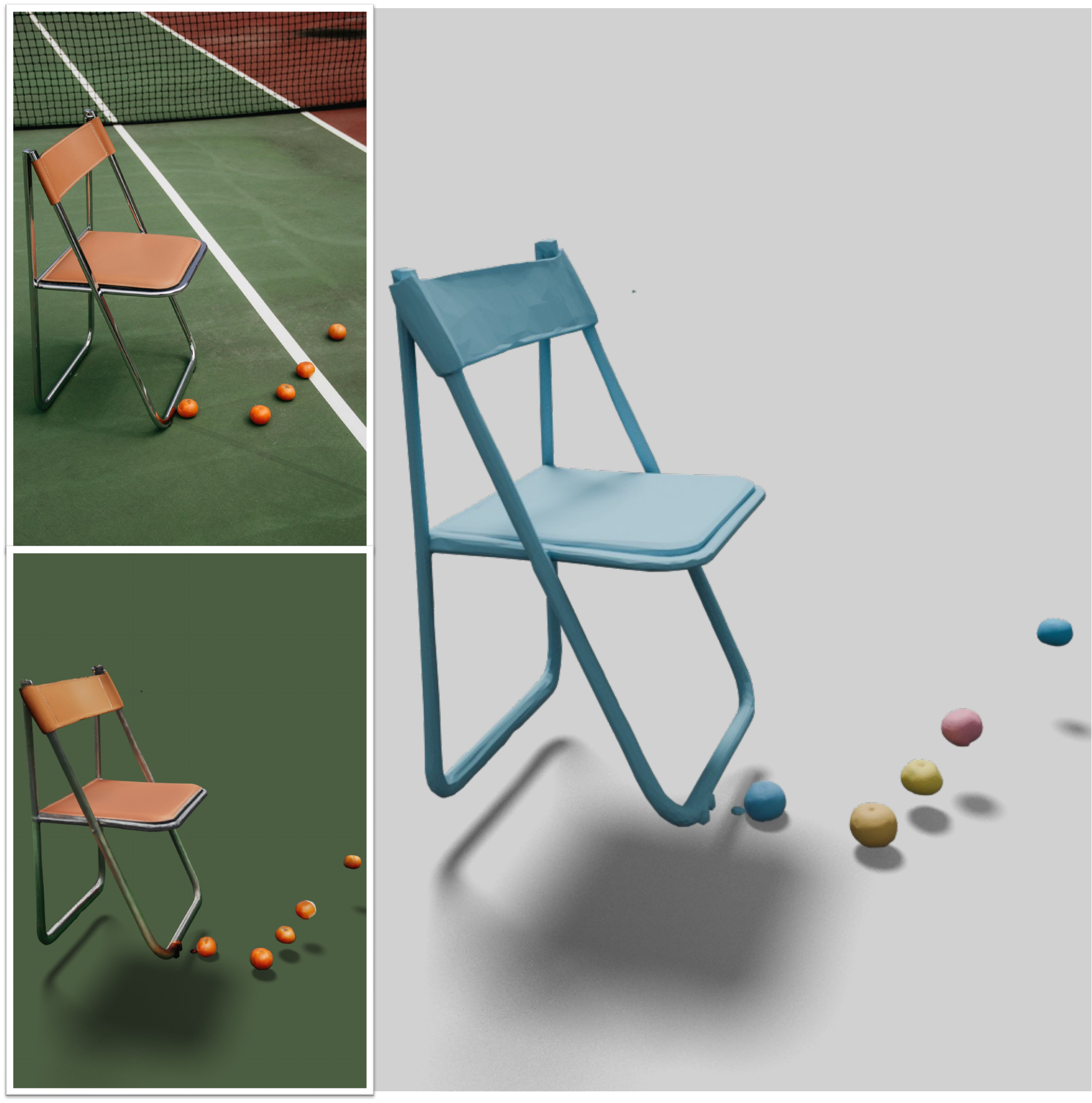}
\end{subfigure} &
\begin{subfigure}[t]{\suppwidth}
\includegraphics[width=\columnwidth]{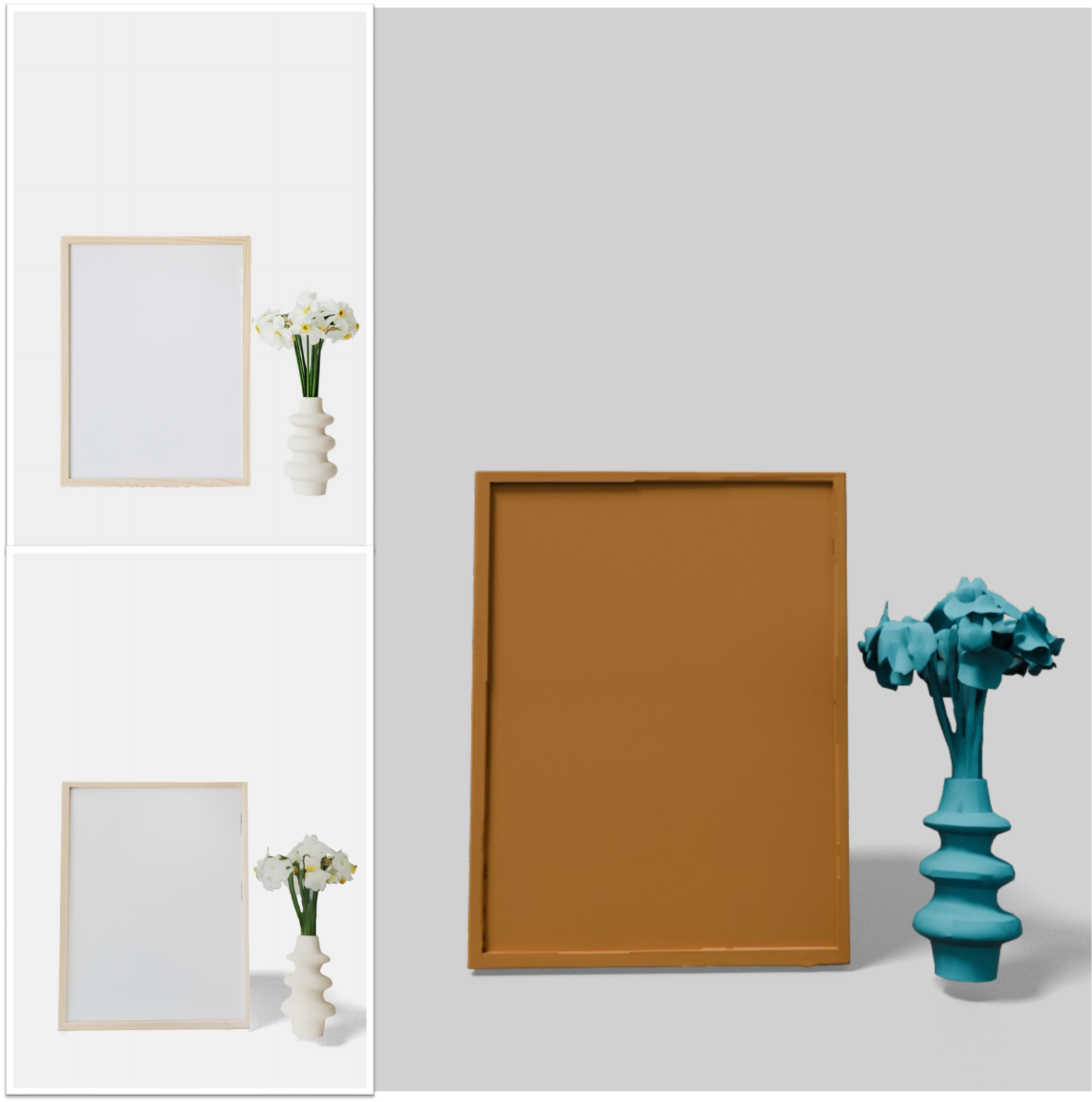}
\end{subfigure}\\
\end{tabular}
}
\caption{
\textbf{Additional examples of component-aligned scene reconstruction.} For each example shown, the panels display: (top left) the input image, (top right or bottom left) the final rendered output, and (bottom) the reconstructed individual components, color-coded for clarity.
}
\label{fig:scene_supp}
\end{figure*}

\subsection{Generative reconstruction diversity.}
\label{ap:scene-diversity} 
Reconstruction methods like LRM~\citep{hong2023lrm} typically generate a single 3D object from one image. However, our approach not only surpasses LRM in terms of reconstruction quality but also produces multiple plausible interpretations—what we refer to as generative reconstruction. Since standard metrics do not effectively capture this diversity, we offer qualitative comparisons in Figure~\ref{fig:model_var}. The results demonstrate that our framework creates diverse 3D models with visible regions that consistently align with the input image. In contrast, conventional 3D generation methods~\citep{xiang2025structured} often struggle to maintain consistency in the visible regions of the objects they create. Furthermore, while directly processing an image with multiple objects may still yield a scene, the generation quality is often degraded as such inputs are out-of-distribution. In contrast, our method allows for component-aligned compositional reconstruction, as demonstrated by the additional examples in Figure~\ref{fig:scene_supp}.

\subsection{Qualitative comparison with 3D generators}
\label{sec:comparison_3dgen}

Since TRELLIS~\citep{xiang2025structured} does not provide an explicit object-centric camera pose for the input image and post-hoc alignment is unreliable, we cannot conduct fair quantitative evaluation of reprojection consistency. We therefore focus on qualitative comparisons in Figure~\ref{fig:color_drifting}. For both methods, we render canonical object from front and back views. As shown, TRELLIS struggles to maintain texture consistency with the input image, exhibiting notable color drifting on the candle statue, teddy bear, and the lamb. In contrast, our method closely preserves the appearance of input. Importantly, although we incorporate pixel-level cues via back-projection, the model does not simply copy pixels: the generated back views remain coherent with the front, indicating learned view-consistent geometry and appearance. These results demonstrate that our mechanism effectively mitigates color drifting and texture inconsistency. We hypothesize that this limitation is common among 3D generators that lack localized, pixel-attended conditioning, including follow-up works of TRELLIS. We hope these findings can inspire future designs of 3D generators.


\end{document}